\documentclass[dvipsnames]{article}

\PassOptionsToPackage{accepted}{template/arxiv}
\usepackage{template/arxiv}

\usepackage{amsmath}
\usepackage{amsfonts}
\usepackage{bm}
\usepackage{amssymb}
\usepackage{mathtools}
\usepackage{amsthm}
\usepackage{dsfont}
\usepackage{bbm} 
\usepackage{stmaryrd} 
\usepackage{enumitem}
\usepackage{todonotes}

\usepackage{xcolor}  
\definecolor{darkcerulean}{rgb}{0.03, 0.27, 0.49} 
\definecolor{iris}{rgb}{0.35, 0.31, 0.81} 
\definecolor{carmine}{rgb}{0.59, 0.0, 0.09} 
\definecolor{lavenderblue}{rgb}{0.8, 0.8, 1.0}
\definecolor{blue(pigment)}{rgb}{0.2, 0.2, 0.6}
\definecolor{blue-violet}{rgb}{0.54, 0.17, 0.89}
\usepackage{hyperref}
   \hypersetup{
     final=true,
     plainpages=false,
     pdfstartview=FitV,
     pdftoolbar=true,
     pdfmenubar=true,
     bookmarksopen=true,
     bookmarksnumbered=true,
     breaklinks=true,
     linktocpage=true,
     colorlinks=true,
     linkcolor=blue-violet, 
     urlcolor=iris, 
     citecolor=darkcerulean, 
     anchorcolor=black
}

\usepackage{url}            
\usepackage{nicefrac}       
\usepackage{microtype}      
\usepackage{subfigure}

\usepackage{algorithm2e}
\RestyleAlgo{ruled}

\usepackage{booktabs}
\usepackage{multirow}
\usepackage{graphicx}
\usepackage{caption} 
\def\thick{0.8} 

\usepackage[capitalize]{cleveref}

\usepackage{amsthm}
\newtheorem{thm}{Theorem}[section]

\newtheorem{rmk}[thm]{Remark}

\def\propcolor{lavenderblue!25}
\usepackage{mdframed}
\newmdtheoremenv[topline=false, bottomline=false, leftline=false, rightline=false, backgroundcolor=\propcolor,%
innertopmargin=\topskip, splittopskip=\topskip, skipbelow=\baselineskip, skipabove=\baselineskip]{boxthm}{Theorem}[section]

\newmdtheoremenv[topline=false, bottomline=false, leftline=false, rightline=false, backgroundcolor=\propcolor,%
innertopmargin=\topskip, splittopskip=\topskip, skipbelow=\baselineskip, skipabove=\baselineskip]{boxprop}[boxthm]{Proposition}

\newmdtheoremenv[topline=false, bottomline=false, leftline=false, rightline=false, backgroundcolor=\propcolor,%
innertopmargin=\topskip, splittopskip=\topskip, skipbelow=\baselineskip, skipabove=\baselineskip]{boxexample}[boxthm]{Example}
\newmdtheoremenv[topline=false, bottomline=false, leftline=false, rightline=false, backgroundcolor=\propcolor,%
innertopmargin=\topskip, splittopskip=\topskip, skipbelow=\baselineskip, skipabove=\baselineskip]{boxcor}[boxthm]{Corollary}
\newmdtheoremenv[topline=false, bottomline=false, leftline=false, rightline=false, backgroundcolor=\propcolor,%
innertopmargin=\topskip, splittopskip=\topskip, skipbelow=\baselineskip, skipabove=\baselineskip]{boxlem}[boxthm]{Lemma}
\newmdtheoremenv[topline=false, bottomline=false, leftline=false, rightline=false, backgroundcolor=\propcolor,%
innertopmargin=\topskip, splittopskip=\topskip, skipbelow=\baselineskip, skipabove=\baselineskip]{boxdef}[boxthm]{Definition}

\usepackage{tikz}
\usepackage{varwidth}

\usepackage{stmaryrd}

\usepackage{dirtytalk}
\MakeRobust{\say} 

\usepackage{makecell}

\usepackage{wrapfig}

\usepackage{enumitem}

\usepackage{nicematrix}

\usepackage{mathrsfs}

\def\thick{0.8}

\usepackage{overpic}

\usepackage{url}

\newcommand{\tick}{\pmb{\textcolor{ForestGreen}{$\bm{\checkmark}$}}~}
\newcommand{\crossi}{\pmb{\textcolor{BrickRed}{$\times$}}~}

\newcommand{\EE}{\mathbb{E}}
\newcommand{\NN}{\mathbb{N}}
\newcommand{\PP}{\mathbb{P}}
\newcommand{\QQ}{\mathbb{Q}}
\newcommand{\RR}{\mathbb{R}}

\newcommand{\II}{\mathbb{I}}

\newcommand{\mbf}[1]{\mathbf{#1}}
\newcommand{\mcal}[1]{\mathcal{#1}}
\newcommand{\mrm}[1]{\mathrm{#1}}
\newcommand{\bbm}[1]{\mathbbm{#1}}

\newcommand{\ie}{i.e.}

\newcommand{\RNum}[1]{\uppercase\expandafter{\romannumeral #1\relax}}

\usepackage{mleftright} 

\newcommand{\relu}[1]{\operatorname{\mathrm{ReLU}}\mleft(#1\mright)}
\newcommand{\softmax}[1]{\operatorname{\mathrm{softmax}}\mleft(#1\mright)}

\newcommand{\TV}[2]{d_\mrm{TV}\mleft(#1, #2\mright)}
\newcommand{\Helling}[2]{H\mleft(#1, #2\mright)}
\newcommand{\KL}[2]{d_\mrm{KL}\mleft(#1||#2\mright)}
\newcommand{\Exp}[2]{\EE_{#1}\mleft[#2\mright]}
\newcommand{\abs}[1]{\lvert#1\rvert}

\newcommand{\prediv}[2]{\mcal{K}(#1, #2)}

\DeclareMathOperator*{\argmin}{arg\,min}

\newcommand{\Pb}[2]{\PP\mleft(#1\mid#2\mright)}
\newcommand{\Pbl}[2]{\PP_{\mathcal{L}}\mleft(#1\mid#2\mright)}
\newcommand{\Pbemp}[3]{\PP_{#1}\mleft(#2\mid#3\mright)}
\newcommand{\risk}[1]{\mcal{R}\mleft(#1\mright)}
\newcommand{\smallrisk}[1]{\widehat{\mcal{R}}(#1)}
\newcommand{\MCseq}[2]{\mleft(\mbf{#1}_1, \ldots, \mbf{#1}_{#2} \mright)}
\newcommand{\MC}[1]{\mleft(\mbf{#1}_1, \mbf{#1}_2, \ldots \mright)}

\newcommand{\tmix}[1]{t_{\mrm{mix}}\mleft(#1\mright)}
\newcommand{\polish}{\Omega}

\newcommand{\params}{\bm{\Theta}}
\newcommand{\mcparams}{\bm{\vartheta}}
\newcommand{\paramspace}{\mathcal{W}}
\newcommand{\mcparamspace}{\mathcal{W}_{\mrm{mc}}}
\newcommand{\funcspace}{\mathcal{F}}

\newcommand{\mcdata}{\mrm{mc}}
\newcommand{\measurable}{\mleft(\Omega, \mcal{F}\mright)}
\newcommand{\cst}{\bar{B}}
\newcommand{\vocabspace}{\mcal{V}}
\newcommand{\vocabsize}{T}
\newcommand{\class}{\mcal{C}}
\newcommand{\embdim}{r}
\newcommand{\ICL}{\mrm{icl}}
\newcommand{\riskicl}[1]{\mcal{R}_{\ICL}\mleft(#1\mright)}
\newcommand{\smallriskicl}[1]{\widehat{\mcal{R}}_{\ICL}\mleft(#1\mright)}
\newcommand{\train}{\mrm{train}}
\newcommand{\pretrain}{\mrm{pre}}
\newcommand{\risktrain}[1]{\mcal{R}_{\pretrain}\mleft(#1\mright)}
\newcommand{\smallrisktrain}[1]{\widehat{\mcal{R}}_{\pretrain}\mleft(#1\mright)}
\newcommand{\transient}{\mathscr{T}}
\newcommand{\transrec}{\mathscr{TR}}
\newcommand{\recurrent}{\mathscr{R}}
\newcommand{\inputsize}{n}
\newcommand{\hiddendim}{m}
\newcommand{\mcsize}{d}
\newcommand{\cxtsize}{K}
\newcommand{\model}{f}
\newcommand{\lmin}{\ell_{\min}}
\newcommand{\mixmat}{\bm{\Gamma}}
\newcommand{\ubtok}{B_{\mrm{tok}}}
\newcommand{\ubmod}{B_{\params}}
\newcommand{\pmin}{p_{\mrm{min}}}
\newcommand{\subspace}{(i_1 \leq i_2 \leq \ldots \leq i_{\mcsize}) \in [T]^{\mcsize}}
\newcommand{\transmat}{\mbf{Q}_f}
\newcommand{\langmat}{\mbf{Q}^*}

\newcommand{\incompatible}{\mcal{I}}
\newcommand{\last}{j_0}

\newcommand{\btau}{\boldsymbol{\tau}}
\newcommand{\mc}{\boldsymbol{M}}
\newcommand{\trn}{^\intercal}
\newcommand{\eps}{\varepsilon}
\DeclareMathOperator{\diag}{diag}

\icmltitlerunning{Large Language Models as Markov Chains}

\begin{document}

\addtocontents{toc}{\protect\setcounter{tocdepth}{0}}

\twocolumn[
\icmltitle{Large Language Models as Markov Chains}

\icmlsetsymbol{equal}{*}

\begin{icmlauthorlist}
\icmlauthor{Oussama Zekri}{equal,a} \quad
\icmlauthor{Ambroise Odonnat}{equal,b,c}\\ \quad \\ 
\icmlauthor{Abdelhakim Benechehab}{b,e} \quad
\icmlauthor{Linus Bleistein}{c} \quad
\icmlauthor{Nicolas Boull\'e}{d} \quad
\icmlauthor{Ievgen Redko}{b} 
\end{icmlauthorlist}

\icmlaffiliation{a}{ENS Paris-Saclay}
\icmlaffiliation{b}{Huawei Noah's Ark Lab}
\icmlaffiliation{c}{Inria}
\icmlaffiliation{d}{Imperial College London}
\icmlaffiliation{e}{Eurecom}

\icmlcorrespondingauthor{OZ}{\url{oussama.zekri@ens-paris-saclay.fr}}
\icmlcorrespondingauthor{AO}{\url{ambroiseodonnattechnologie@gmail.com}}

\icmlkeywords{Machine Learning, ICML}

\vskip 0.3in
]

\printAffiliationsAndNotice{\icmlEqualContribution} 

\begin{abstract}
    Large language models (LLMs) are remarkably efficient across a wide range of natural language processing tasks and well beyond them. However, a comprehensive theoretical analysis of the LLMs' generalization capabilities remains elusive. In our paper, we approach this task by drawing an equivalence between autoregressive transformer-based language models and Markov chains defined on a finite state space. This allows us to study the multi-step inference mechanism of LLMs from first principles. We relate the obtained results to the pathological behavior observed with LLMs such as repetitions and incoherent replies with high temperature. Finally, we leverage the proposed formalization to derive pre-training and in-context learning generalization bounds for LLMs under realistic data and model assumptions. 
Experiments with the most recent Llama and Gemma herds of models show that our theory correctly captures their behavior in practice.

\end{abstract}

\allowdisplaybreaks
\section{Introduction}
\label{sec:intro}

\begin{figure}[!t]
    \centering
    \includegraphics[width=\linewidth]{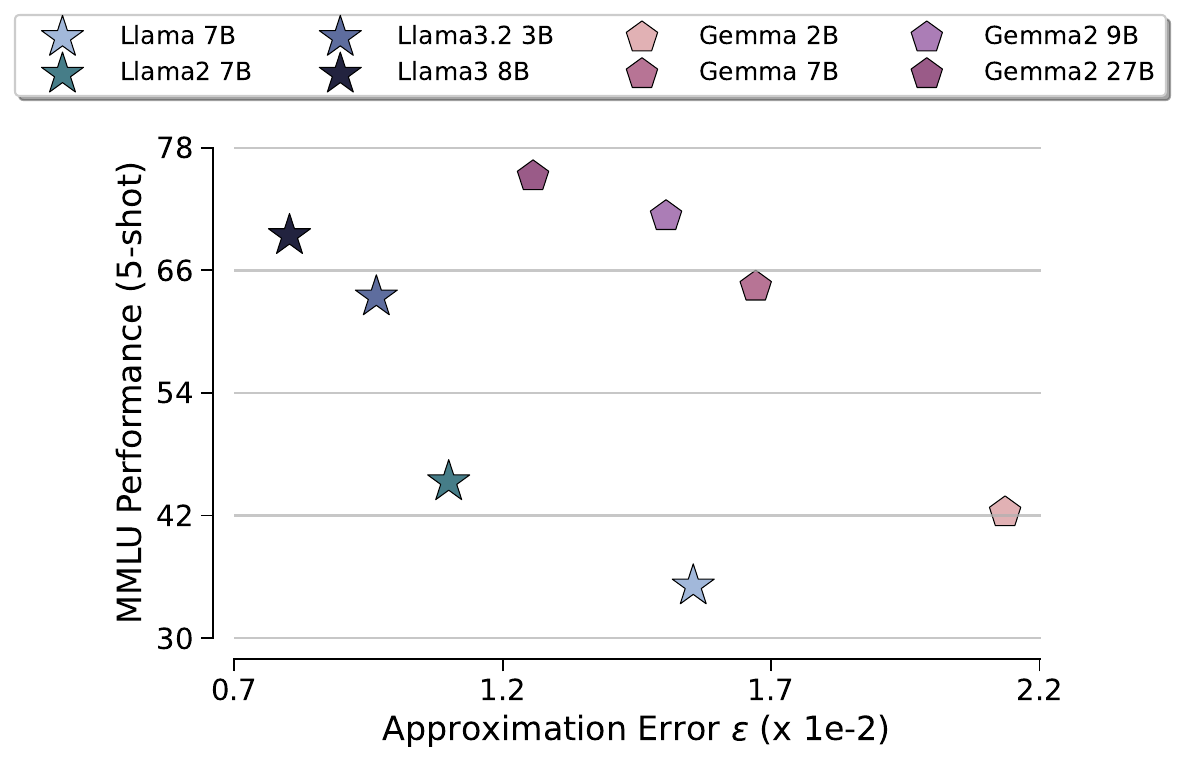}
    \caption{\emph{LLMs' Sample Complexity}. We plot the Massive Multitask Language Understanding (MMLU) \citep{hendrycks2021measuring} performance with respect to the approximation error $\epsilon$ predicted by \cref{cor:sample_complexity}. We set $N^*$ equal to the real number of pre-training tokens. Each point represents a model from the Llama or Gemma families~\citep{gemmateam2024gemma2improvingopen, dubey2024llama3}. The approximation error $\epsilon$ predicted by our theory correlates with the real performance, with different trends between the models' families.}
    \label{fig:mmlu_epsilon_separate}
\end{figure}

Large language models (LLMs)~\citep{brown2020LLMfewshot,touvron2023llama} have reshaped the landscape of machine learning and artificial intelligence. Trained on vast amounts of data, they have reached unprecedented performance in tasks such as machine translation~\citep{brown2020LLMfewshot}, text generation, question answering~\citep{roberts-etal-2020-much}, and sentiment analysis~\citep{zhang2023sentiment}, to name a few. Pre-trained LLMs also exhibit an intriguing capability of performing learning from the context, also called in-context learning (ICL), without explicitly updating their parameters. 

Despite all these remarkable achievements, the theoretical justification for LLMs' impressive performance remains elusive. The two main reasons for this are 1) the complexity of analyzing modern deep transfer-based architectures used to train LLMs and 2) the non-iid nature of their pre-training data. Prior works usually bypass these challenges by imposing simplifying assumptions. Such assumptions include considering a shallow single-head transformer \citep{makkuva2024attention}, only self-attention mechanism \citep{ildiz2024self} or attention-only transformers \citep{jeon24a,edelman2024evolutionstatisticalinductionheads} when studying LLMs. The others are to study the auto-regressive mechanism of the LLMs without taking into account their architecture explicitly \citep{xie2022iclbayesian,lotfi2023non,lotfi2024unlocking}. When the architecture is studied in full generality \citep{zhang2023whathowicl,li2023transformers}, it is also common to relax the non-iid assumption by a certain predefined Markovian process. 
 
\begin{figure*}[!t]
    \centering
    \includegraphics[width=.9\textwidth]{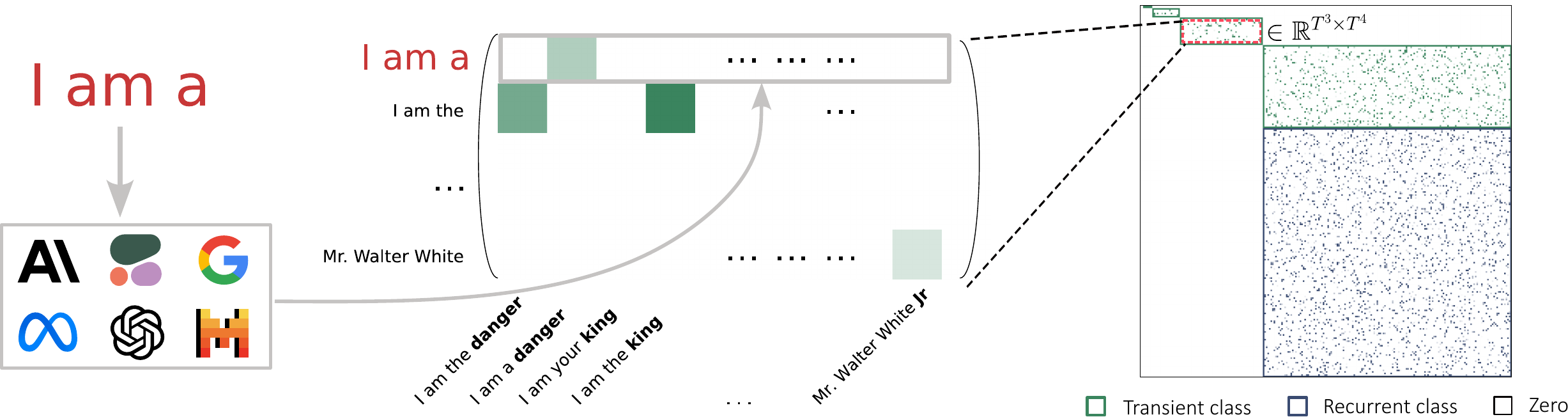}
    \caption{\emph{LLM as a Markov chain}. A large language model with vocabulary size $T$ and context window $\cxtsize$ is equivalent to a Markov chain with a sparse and block-structured transition matrix of size $\smash{\sum_{i\leq \cxtsize} T^i \sim \mcal{O}(T^{\cxtsize})}$. The latter captures all possible outputs of a given LLM for all possible input sequences allowed by its vocabulary and context window.}
    \label{fig:exp_matrix}
\end{figure*}

This work provides a unified theoretical analysis of LLM's inference, pre-training, and ICL under realistic assumptions on model architecture and pre-training data. Our core idea is to note that despite the seemingly infinite generation capacity of LLMs, their vocabulary and context window are finite, making all possible input and output sequences countable. This leads to an intuitive, yet overlooked, approach that interprets LLMs as Markov chains whose transition matrix operates on a finite state space of such sequences (see \cref{fig:exp_matrix}). The generalization of LLMs is then their capacity to learn the transition probabilities of language itself. 

\noindent\textbf{Main contributions.} Our contributions can be summarized as follows. 
\begin{enumerate}[leftmargin=*, label={\arabic*)}]
    \item We characterize the inference mechanism of any LLM as a \emph{finite-state} Markov chain. This provides a deeper understanding of LLMs' generation behavior and sheds new light on some of their pathological behaviors. 

    \item We prove sample complexity bounds for deep transformer-based LLMs pre-trained on \emph{non-iid} data. They are more general than existing results (see \cref{tab:theory_recap} for a full comparison with the literature) and are verified in practice on the Llama and Gemma herd of models from $2$B to $27$B (see \cref{fig:mmlu_epsilon_separate} and \cref{sec:theoretical_analysis}).
    
    \item We derive ICL generalization bounds when LLMs are prompted with Markov chains. We validate our results by showing that the most recent open-source LLMs obey the in-context \emph{scaling laws} predicted by our theory. 
\end{enumerate}

\renewcommand{\arraystretch}{1.05}
\begin{table}[!t]
    \centering
    \resizebox{\linewidth}{!}{
    \begin{tabular}{lccccc}
    \toprule[\thick pt]
         \textbf{Method} & \textbf{Pre-Train.} & \textbf{ICL} & \textbf{Input}&  \textbf{Model-Dep.} & \textbf{Exp. val.}\\
        \midrule[\thick pt]
        \cite{xie2022iclbayesian} & \crossi & \tick & HMM
        & \crossi & \tick\\ 
        \cite{zhang2023whathowicl} & \tick & \tick & MC
        & \tick & \crossi \\
        \cite{li2023transformers} & \tick & \tick & MC
        & \tick & \tick\\ 
        \cite{lotfi2024unlocking} & \tick & \crossi &  non-iid
        & \crossi & \tick \\
        \midrule
        \textbf{Ours}  & \tick & \tick & non-iid
        & \tick & \tick\\
        \bottomrule[\thick pt]
    \end{tabular}
    }
    \caption{LLMs generalization bounds proposed in the literature. \textbf{Pre-train.} stands for pre-training; \textbf{Input} refers to the assumptions on the data generating process (Markov Chains (MC); Hidden Markov Model (HMM); \textbf{Model-dep.} means explicit dependence on model's architecture; \textbf{Exp.val.} stands for the experimental validation of the theory.}
    \label{tab:theory_recap}
\end{table}
\noindent\textbf{Organization of the paper.\quad} \cref{sec:background} provides background material on transformer-based autoregressive LLMs and Markov chains. In \cref{sec:mc_formalization}, we formalize the multi-step inference mechanism of a pre-trained LLM as a Markov chain. In \cref{sec:theoretical_analysis}, we study how a deep transformer-based LLM learns this inference mechanism through pre-training on non-iid data using the formalism of Marton couplings and also extend the obtained results to the ICL setting. We verify them through a set of experiments in \cref{sec:experiment}.

\section{Background Knowledge}
\label{sec:background}
We recall some facts about Markov chains~\citep{robert2004MC, paulin2015ineqMC} and LLMs. More notations and background materials are available in \cref{app:notations,app:background_transformer,app:background_mc}.

\noindent\textbf{Markov chains.\quad}
Let $\polish$ be a discrete finite set of size $\abs{\polish}$. A discrete-time, time-homogeneous Markov chain $\text{MC}(\polish, \mathbf{Q})$ defined on a state space $\polish = \{x_i\}_{i=1}^{\lvert \polish \rvert}$ with transition matrix $\smash{\mathbf{Q} \in \RR^{\lvert \polish \rvert \times \lvert \polish \rvert}}$ with entries $\mbf{Q}_{ij} = \mbf{Q}\mleft(x_i, x_j\mright) \in [0, 1]$ is a sequence of random variables $\MC{X}$ taking values in $\polish$ such that for any $n \in \NN$ and $(x_1, \ldots, x_{n+1}) \in \polish^{n+1}$, we have
\begin{equation*}
\begin{split}
    &\Pb{\mbf{X}_{n+1} = x_{n+1}}{\mbf{X}_n = x_n, \ldots, \mbf{X}_1 = x_1} \\
    &\qquad = \Pb{\mbf{X}_{n+1} = x_{n+1}}{\mbf{X}_n = x_n} \\
    &\qquad \coloneqq \mathbf{Q}(x_n,x_{n+1}).
\end{split}
\end{equation*}
A distribution $\pi$ on $\polish$ is called stationary if $\pi \mathbf{Q}= \pi$. 
For any $x \in \polish$, $\text{MC}(\polish, \mathbf{Q})$ converges to $\pi$ if $\lim_{n \to \infty} \TV{\mathbf{Q}^n\mleft(x, \cdot\mright)}{\pi} = 0$, where $\mathbf{Q}^n\mleft(x, \cdot\mright)$ denotes the probability of $\mbf{X}_n$ conditioned on $\mbf{X}_1 = x$ and the total variation between two distributions $\PP$ and $\QQ$, defined on $\mleft( \Omega, \mcal{F} \mright)$, is
    $\TV{\PP}{\QQ} \coloneqq \sup_{A \in \mcal{F}} \lvert \PP\mleft(A\mright) - \QQ\mleft(A\mright) \rvert.$
We recall that the mixing time $t_\textnormal{mix}(\varepsilon)$ of a Markov chain is the minimal time needed to be $\varepsilon$-close to its stationary distribution (see \cref{def:mixing_time}). Intuitively, a Markov chain mixes slowly when it remains close to the initial state after a given number of steps and doesn't explore its state space. A Markov chain with rapid mixing quickly forgets its initial state and transitions more easily to a wider set of states. 

\noindent\textbf{Large language models.\quad} Let $\mcal{V}$ denote a dictionary of size $\vocabsize$, referred to as the \emph{vocabulary size}, used to encode an arbitrary sequence into a sequence of predefined tokens belonging to $\mcal{V}$. We assume that our model admits a maximum of $\cxtsize$ tokens as input, referred to as the \emph{context window} of the model. The domain of the transformer-based autoregressive LLM is the set of all sequences consisting of at most $\cxtsize$ elements of $\vocabspace$. We denote this space by $\vocabspace_\cxtsize^*$, i.e., $\smash{\vocabspace_\cxtsize^* := \{ v \in \vocabspace^*,\, \lvert v\rvert\leq \cxtsize\}}$ with $\lvert v\rvert$ the length of $v$. We define an LLM with trainable parameters $\params$ as a function $\smash{\model_{\params}^{\vocabsize,\cxtsize}: \mcal{V}_\cxtsize^* \rightarrow \Delta(\mcal{V})}$, where $\Delta(\mcal{V})$ is the probability simplex over $\mcal{V}$. Given a sequence of tokens $v$, $\smash{\model_{\params}^{\vocabsize,\cxtsize}}$ outputs a probability distribution over the whole state space indicating the likelihood for each of its elements to appear after $v$ (see \cref{app:background_transformer} for more details). Noting that an LLM is defined for a given vocabulary size $\vocabsize$ and context window $\cxtsize$, we will drop the exponents and simply write $\model_{\params}$ to ease the notations. We consider a setting where the learner’s objective is to approximate the probability distribution of sequences over an input vocabulary $\PP_{\mcal{L}}: \mcal{P}(\vocabspace_\cxtsize^*) \rightarrow [0,1]$ where $\mcal{P}(\vocabspace_\cxtsize^*)$ denotes the powerset of $\vocabspace_\cxtsize^*$. We also assume the existence of a constant $c_0 > 0$ such that for any $n \in [N]$ and $( x_1, \ldots, x_{n+1}) \in \polish^{n+1}$,
\begin{equation}
    \label{eq:ambiguity_language}
    \Pbl{\mbf{X}_{n+1}=x_{n+1}}{\mbf{X}_n = x_n, \ldots, \mbf{X}_1=x_1} \geq c_0.
\end{equation}
This is a common assumption used in \citep{zhang2023whathowicl,xie2022iclbayesian, wies2024learnability}. 

\noindent\textbf{Transformer model.\quad}
Without loss of generality, $\model_{\params}$ is assumed to be a transformer model with $L$ layers and $H$ heads, consisting of alternating multi-head attention (MHA) and feed-forward blocks (more details in ~\cref{app:background_transformer}). The first layer receives an input $\smash{\mbf{S}^{(0)} = \mbf{S}}$ embedded in an $\embdim$-dimensional space. To obtain a probability distribution on the vocabulary $\vocabspace$, the output $\mbf{S}^{(L)} \in \RR^{\embdim \times T}$ of the final layer is projected back to the vocabulary size by an \say{unembedding layer} $\mbf{W}_U \in \RR^{\vocabsize \times \embdim}$ and averaged over the columns to obtain a vector in $\RR^{\vocabsize}$. A softmax layer is finally applied to obtain the probability distribution of the next token $\Pbemp{\params}{\cdot}{\mbf{S}}\coloneqq
     \mrm{softmax}\mleft( \frac{1}{\inputsize \tau}\mbf{W}_U \mbf{S}^{(L)} \bbm{1}_\inputsize\mright) \in \Delta_{\vocabsize}$, where $\params$ denotes the parameters of the entire network and $\tau$ is the softmax temperature~\citep{hinton2015distilling}. Unless otherwise specified, we assume that the unembedding layer is bounded. The classes of parameters and neural networks it generates respectively write $\paramspace = \{ \params \text{ s.t. } \lVert \mbf{W}_U^\top \rVert_{2, 1} \leq B_U\}$ and $\funcspace = \{ \model_{\params} \text{ s.t. } \params \in \paramspace \}$.
     
\section{Large Language Models as Markov Chains}
\label{sec:method}
To proceed, we formally define a Markov chain that explicitly captures the inference of a given LLM $\model_{\params}$. We build upon a high-level idea that associates a tokenized input sequence with a state $v_i$, from which we transition to a new state $v_j=[v_i,v]$ by concatenating the token $v$ predicted by an LLM to it. We then provide a theoretical characterization of this Markov chain. Our analysis holds for any model with finite fixed vocabulary, and context window. This setup includes deep transformer-based LLMs while excluding models such as RNNs and LSTMs. 

\subsection{Markov Chain Formalization}
\label{sec:mc_formalization}
We first introduce the notion of \say{incompatible sequences} in the sense of next-token prediction. 
\begin{boxdef}[Incompatible Sequences]
    Let $u, v \in \mcal{V}_\cxtsize^*$ be two sequences of at most $\cxtsize$ tokens. We say that $v$ is incompatible with $u$ if $v$ cannot be a plausible completion of $u$, that is
    \begin{equation}
    \label{eq:incompatible}
    \exists l\in \{1, \ldots, |u|-1\}, \text{s.t. }(u)_{l+1} \neq (v)_l.
    \end{equation}
We denote by $\incompatible$ the set of incompatible sequences, i.e., $\incompatible \coloneqq \{ (u, v) \text{ s.t. } v \text{ is incompatible with } u\}$.
\end{boxdef}
The order matters since a necessary condition for $v$ to be compatible with $u$ is that $v$ has as many or one more token 
than $u$. We now proceed with the definition of the transition matrix associated with an LLM $\smash{\model_{\params}}$.
\begin{boxprop}\label{prop:LLM_formal_def}
    Any large language model $\model_{\params}$ can be equivalently represented by a Markov chain $\text{MC}(\mcal{V}_\cxtsize^*, \transmat)$, with a sparse transition matrix $\transmat \in \RR^{|\mcal{V}_\cxtsize^*|\times |\mcal{V}_\cxtsize^*|}$ that writes for any $v_i,v_j \in \mcal{V}_\cxtsize^*$:
    \[
         \transmat(v_i, v_j) =
        \begin{cases}
            0,                                                 & \text{if } (v_i, v_j) \in \incompatible \\
            \{\model_{\params}(v_i)\}_{\last}, & \text{otherwise},                                                                   \\
        \end{cases}\]
        where $\last$ denotes the index in $\vocabspace$ of the last token of $v_j$. We have $|\mcal{V}_\cxtsize^*| = \vocabsize(\vocabsize^{\cxtsize} - 1)/(\vocabsize-1)$ and the proportion of non-zero elements in $\transmat$ is $(\vocabsize - 1)/(\vocabsize^{\cxtsize}-1)$. 
\end{boxprop}
\cref{prop:LLM_formal_def} states that if $v_j$ is incompatible with $v_i$, then the probability to go from the state $v_i$ to $v_j$ is zero as LLM only transitions between compatible sequences. For such compatible sequences, the probability of going from $v_i$ to $v_j$ is naturally the next-token probability associated with the last token of $v_j$, i.e., $\{\model_{\params}(v_i)\}_{\last}$. 

\begin{figure}[!h]
\centering
\includegraphics[width=0.9\linewidth]{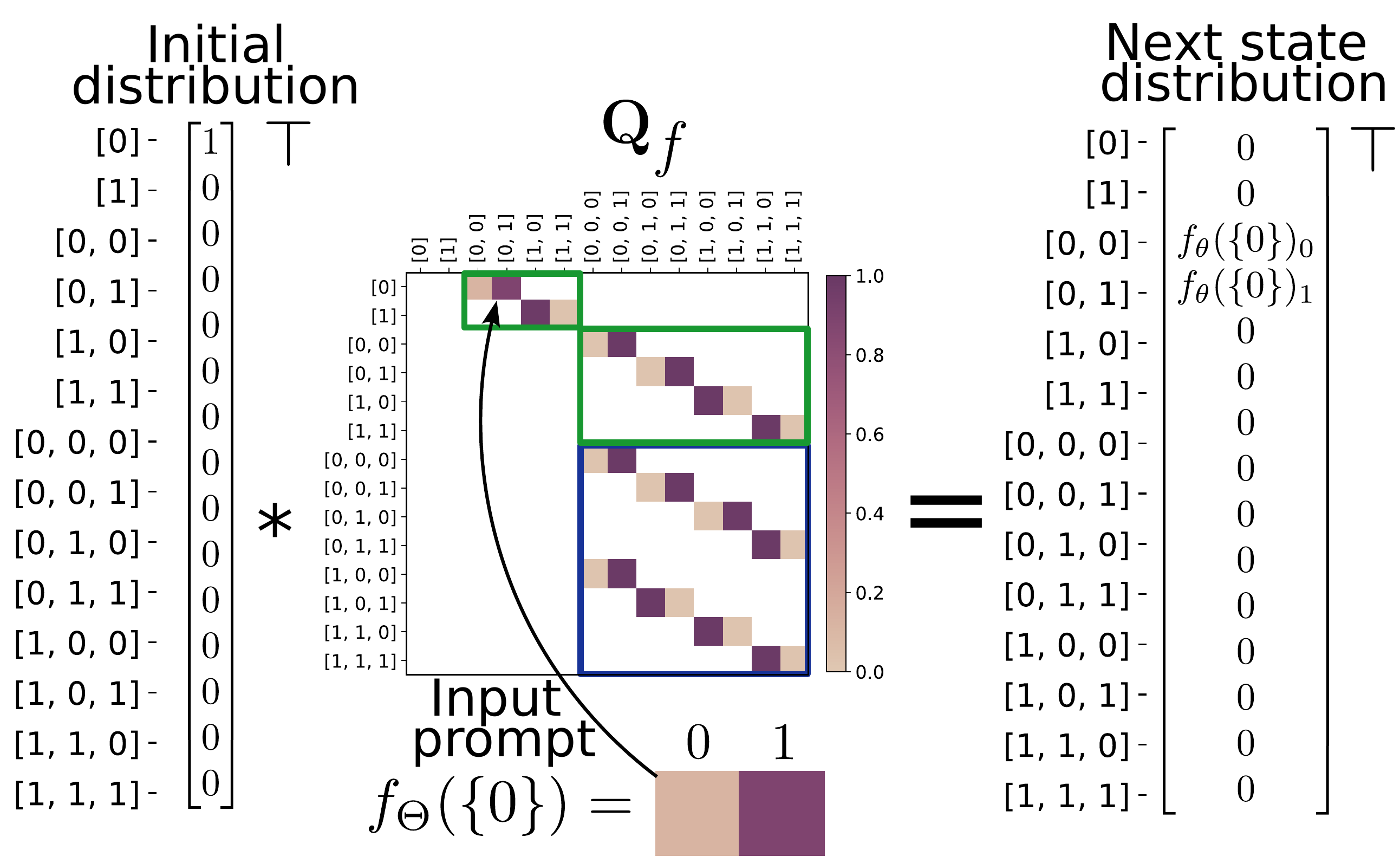}
\caption{\emph{\cref{prop:LLM_formal_def} with $\vocabsize=2$ and $\cxtsize=3$}.}
\label{wrapfig_mult_Q}
\end{figure}

\noindent\textbf{Example: binary sequences.\quad} To help unpack \cref{prop:LLM_formal_def}, we illustrate it with $\vocabsize=2$ and $\cxtsize=3$ in \cref{wrapfig_mult_Q} (see the general case in \cref{fig:exp_matrix}). 
Here, we start with an input prompt consisting of one token. We then transition to sequences of increased size, while attributing a 0 probability to incompatible sequences. Each generation step for different values of $k< \cxtsize$ defines green rectangular blocks of size $\vocabsize^k \times \vocabsize^{k+1}$ in the transition matrix $\transmat$. 

Reaching the blue \textit{square} block in the transition matrix means that the input sequence is of the maximum size $K$. The model can no longer append tokens to it and has to delete the first token to proceed. This blue block is of size $T^\cxtsize \times T^{\cxtsize}$: it captures transitions between sequences of the maximum admissible length. We define similarly the reference transition matrix $\langmat$ of the language where the probability of transitions $\smash{\{\model_{\params}(v_i)\}_j}$ are replaced by ground-truth probabilities $\Pbl{v_j}{v_i}$. To use $\transmat$ as $\model_{\params}$, it is sufficient to define an input distribution $\delta_0$ of the Markov chain based on input prompt $v$. It is a one-hot encoding vector of size $|\mcal{V}_\cxtsize^*|$ with 1 at the position of the state corresponding to $v$. Then, the transition to the next state writes $\delta_1 = \delta_0\transmat$. The output of $\model_{\params}(v)$ for individual tokens in $\vocabspace$ would then correspond to probabilities in $\delta_1$ for states that are concatenations of $v$ with $T$ tokens from $\vocabspace$. This process is illustrated in \cref{wrapfig_mult_Q}.
\begin{figure*}[!t]
    \centering
    \begin{overpic}[width=\textwidth]{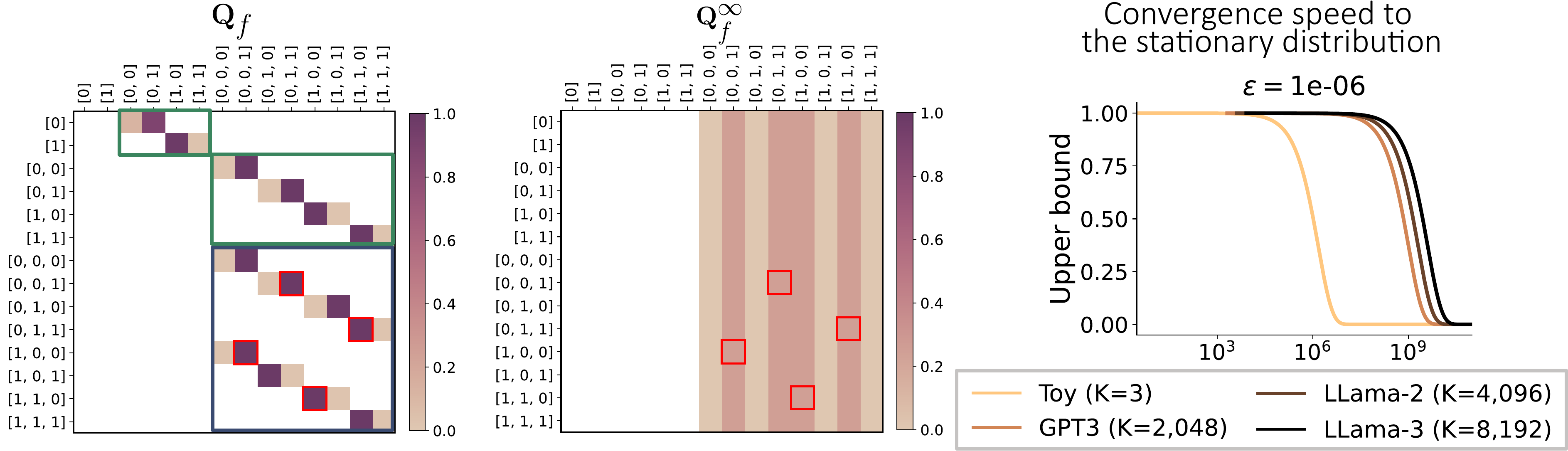}
        \put(0,27){(a)}
        \put(30,27){(b)}
        \put(60,27){(c)}
    \end{overpic}
    \caption{\emph{Markov chain with a small GPT-like model}. (a) Transition matrix $\transmat$ of the model where $\textcolor{red}{\square}$ denotes the examples from the training set. 
        (b) The stationary distribution of the trained model assigns almost uniform probabilities to the states seen during training. (c) Convergence rate to the stationary distribution for the considered toy model along with three LLMs, highlighting the dependence on $\cxtsize$. The y-axis is the upper bound in \cref{prop:stationary_distrib}.}
    \label{fig:ex_sec3}
\end{figure*}
We study the properties of $\text{MC}(\mcal{V}_\cxtsize^*, \transmat)$ below.
\begin{boxprop}\label{prop:ergodic_unichains}
    Let $\text{MC}(\mcal{V}_\cxtsize^*, \transmat)$ be a Markov chain defined in \cref{prop:LLM_formal_def}. Then $\text{MC}(\mcal{V}_\cxtsize^*, \transmat)$ is ergodic and admits a unique stationary distribution.
\end{boxprop}
The proof builds on showing that such a Markov chain has at most one recurrent class (blue block in \cref{fig:exp_matrix}) plus some additional transition states (green blocks in \cref{fig:exp_matrix}).
We now characterize how many times one should apply $\transmat$ to the input to reach the stationary distribution.
\begin{boxprop}
\label{prop:stationary_distrib}
    Let $\text{MC}(\mcal{V}_\cxtsize^*, \transmat)$ be as in \cref{prop:ergodic_unichains}  and $e = (1, 1, \hdots, 1)^\top$. Then we have that $\lim_{n \to \infty} \transmat^n = e\bm{\pi}$, where $\bm{\pi}$ is the stationary distribution of the recurrent class $\recurrent$ of states, expanded by $0$’s for each transient state of the unichain. Moreover, for all $n \geq \cxtsize,$ $$\displaystyle\lvert (\transmat^n)_{i,j} - (e\bm{\pi})_{i,j} \rvert \leq (1-2\varepsilon)^{\lfloor\frac{n}{\cxtsize}\rfloor - 1},$$
    where $\varepsilon = \underset{i,j \in \recurrent^2}{\min}\{(\transmat^\cxtsize)_{i,j}\}>0$.
\end{boxprop}
The stationary distribution is the long-term equilibrium of the multi-step inference of an LLM. While its true value is intractable because of the size of $\mcal{V}_\cxtsize^*$, we discuss below the implications of \cref{prop:stationary_distrib} for LLMs. 
\begin{enumerate}
    \item \textbf{Looping.\quad} The stationary distribution is independent of the initial state (\ie, input prompt), and reaching it should send the LLM into a deterministic loop of repetitions. This behavior is well known and occurs frequently in practice with many existing LLMs \citep{ivgi2024loops}, requiring adding a repetition penalty.
    \item \textbf{Temperature and coherence.\quad} Increasing temperature of the model leads to a decreasing coherence of its outputs as measured by perplexity \citep{peeperkorn2024temperaturecreativityparameterlarge}. Temperature, in turn, impacts  $\varepsilon$ (the smallest element of the $\cxtsize$\textsuperscript{th} power of the transition matrix), as it modifies the probabilities of the predicted next tokens. Consequently, increasing the temperature increases the speed of convergence to the stationary distribution making the output less coherent. We illustrate this on a toy example below. 
\end{enumerate}

\subsection{Illustration on a Toy Model}
We illustrate the results of \cref{sec:method} on a toy model trained on a sequence of $0$s and $1$s. Here, each subsequent token is $0$ if the sum of three previous tokens is even and $1$ otherwise. Therefore, $T=2$ and $\cxtsize=3$. We generate a sequence of $40$ digits, resulting in $37$ distinct supervised examples, and train a small ``GPT-like'' model~\citep{karpathy_minGPT} on it. We extract the logits from the model by prompting it with all possible combinations of $0$s and $1$s of length less than three to obtain the transition matrix $\transmat \in \RR^{14\times 14}$ depicted in \cref{fig:ex_sec3}(a). The transition matrix's structure (e.g., presence of transient and recurrent classes) matches the one presented in \cref{fig:exp_matrix}. \cref{fig:ex_sec3}(b) displays the stationary distribution of the trained model obtained by raising $\transmat$ to power $10^5$. We note that it has a strong bias toward seen training samples in accordance with our intuition behind the stationary distribution presented earlier. Finally, \cref{fig:ex_sec3}(c) illustrates the convergence rate of the toy model, predicted by \cref{prop:stationary_distrib}, and compares it to models with larger dictionary size $T$ and context window $\cxtsize$. In \cref{fig:ex_sec3}(c), we set $\varepsilon=10^{-6}$ and note that this parameter reflects the ability of the LLM to explore the state space. 

\begin{figure*}[t]
    \centering
    \begin{overpic}[width=\textwidth]{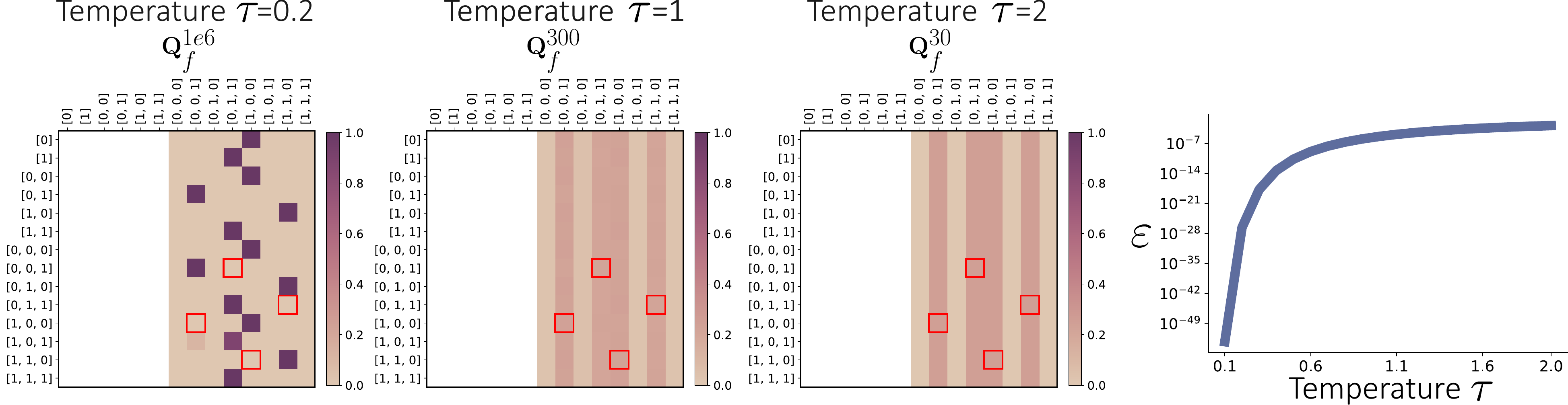}
        \put(0,22){(a)}
        \put(24,22){(b)}
        \put(48,22){(c)}
        \put(72,22){(d)}
    \end{overpic}
    \caption{\emph{Dependence of $\varepsilon$ on the temperature of the model}. (a) For low temperatures, $\varepsilon$ becomes too small to achieve convergence to the stationary distribution. (b)-(c) Increasing the temperature from $1$ to $2$ leads to a $\times 10$ faster convergence. (d) $\varepsilon$ (log-scale) increase for temperature values in $[0.1, 2]$.}
    \label{fig:ex_temps_epsilon}
\end{figure*}

\noindent\textbf{Role of the temperature.\quad} To better illustrate the role of $\varepsilon$, we now plot the transition matrix of the studied Markov chain obtained when applying different temperature scaling to the logits returned by the trained model. As the temperature is commonly linked to the ability of LLMs to transition more freely to a large set of states~\citep{creativity}, we expect that lower temperatures should negatively impact the convergence speed to the stationary distribution. In \cref{fig:ex_temps_epsilon}(a), we show that for a low temperature ($0.2$), the Markov chain mixes slowly and is unable to reach its stationary distribution (same line in the transition matrix as in \cref{fig:ex_sec3}(c)) even after $10^6$ steps. In the case of a more commonly used temperature equal to $1$ (\cref{fig:ex_temps_epsilon}(b)), the model requires only $300$ steps to converge. Finally, setting the model's temperature to $2$ (\cref{fig:ex_temps_epsilon}(c)) makes the convergence extremely fast, reaching the stationary distribution after only $30$ steps. The interplay between $\varepsilon$ and the model's temperature is displayed in \cref{fig:ex_temps_epsilon}(d), increasing the temperature leads to a drastic improvement in the convergence speed.

We have demonstrated that any autoregressive transformer-based LLM admits an equivalent Markov chain formulation. This equivalence connects nicely to some of the pathological behaviors of LLMs observed in practice. We will now use this formalization to provide a fine-grained analysis of LLMs' generalization capabilities that may be of interest to both theorists and practical users.
\section{Sample Complexity and Generalization}
\label{sec:theoretical_analysis}
We now study the generalization of an LLM $\model_{\params}$ as its capacity to infer correctly all the elements of $\mbf{Q}_f$, while approximating the true reference matrix of transition probabilities $\mbf{Q}^*$. The hardness of this task is highlighted by the fact that an LLM observes a negligible amount of $\mbf{Q}^*$'s elements during its pre-training. Indeed, for GPT-3~\citep{brown2020LLMfewshot}, this represents $5\times 10^{11}$ training tokens, which pales in comparison with the number of non-zero elements in $\mbf{Q}_f$, given by $T^{K+1} \approx 10^{9632}$. In this section, we determine the number of pre-training tokens required to be $\epsilon$-close to perfect generalization. This result stems from a novel pre-training generalization bound on \emph{non-iid} random variables in \cref{sec:pretraining}. We extend this bound to the in-context learning in \cref{sec:icl}.

\subsection{Main Result: Pre-training Sample Complexity}
\label{sec:sample_complexity}
\paragraph{Non-iid data.} We denote by $X = \MCseq{X}{N_{\train}}$ the $N_{\train}$ tokens in $\vocabspace$ that $\model_{\params}$ observes during pre-training. The training sequences of tokens can be written as $\mbf{S}_n = \MCseq{X}{n}$ if $n \leq \cxtsize$ and  $\mbf{S}_n = \mleft(\mbf{X}_{n-K+1}, \ldots, \mbf{X}_n \mright)$ otherwise due to the deletion process (see~\cref{def:deletion_process}). In particular, the $\mbf{S}_n$ are elements of $\vocabspace^*_{\cxtsize}$. We assume that the pre-training data $S = \MCseq{S}{N_{\train}}$ is a sequence of \emph{dependent} random variables with a mild coupling structure, namely that a Marton coupling with mixing matrix $\mixmat$ exists for $S = \MCseq{S}{N_{\train}}$ (more details on Marton coupling can be found in~\cref{app:background_marton}). This ensures that our setting remains very broad as it subsumes the case of independent variables, $m$-dependent variables, language bigrams~\citep{bietti2023birth}, and the Markov chain setting considered in state-of-the-art ICL analysis of LLMs~\citep{zhang2023whathowicl, hu2024statscot}. 

We state our main result on the pre-training sample complexity below. The proof is deferred to \cref{app:sample_complexity}. 
\begin{boxprop}[Sample complexity]
\label{cor:sample_complexity}
   Let $\delta \in [0, 1]$ and $\epsilon > 0$. Assuming a perfect pre-training of $\model_{\params}$ and $N_{\train} $ pre-training tokens with $N_{\train} \geq N^* \coloneqq \lceil\frac{4\bar{B}^2}{\epsilon^2}\log{\mleft( \frac{2}{\delta}\mright)}\rceil$, we have with probability at least $1-\delta$,
    \begin{equation*} 
        \EE_{\mbf{S} \sim \PP_{\mcal{L}}}\lVert \langmat\mleft(\mbf{S}, \cdot \mright) - \transmat\mleft(\mbf{S}, \cdot \mright)\rVert_1
        \leq \epsilon,
    \end{equation*}
    where $$\bar{B}=2\lVert \mixmat \rVert\max\{\log{(\vocabsize)} + 2B_U/\tau, \log{(1/c_0)}\}^{1/2}$$ is a model- and data-dependent constant.
\end{boxprop}
\cref{cor:sample_complexity} shows that the number of pre-training tokens needed to achieve a good approximation mostly depends on the problem's parameters captured by $\bar{B}$. This constant includes two main ingredients. On the one hand, it accounts for the model's architecture: the context window size $\cxtsize$, the temperature scaling of the output logits $\tau$, and the norm of the unembedding layer $B_U$. On the other hand, it captures -- via the mixing matrix $\mixmat$ -- how interdependent the tokens are, including the special cases of \emph{iid} data ($\lVert \mixmat \rVert = 1$) and \emph{strongly dependent} sequences of tokens. Thus, the bound is both model and data-dependent, contrary to the previously proposed non-vacuous bounds of~\citet{lotfi2024unlocking} where the model's architecture was not fully taken into account.
  
\noindent\textbf{Practical insights.\quad} \cref{cor:sample_complexity} can be used to predict how well an LLM understands language given a fixed number of pre-training tokens. Letting $N_{\train} = N^*$ in \cref{cor:sample_complexity} leads to an approximation error $\epsilon = \frac{2\bar{B}}{\sqrt{N_{\train}}} \sqrt{\log{\mleft( \frac{2}{\delta}\mright)}}$. Since $\bar{B}$ can be approximated as $\bar{B} \sim 2(\log{(\vocabsize)} + 2\vocabsize \sqrt{\embdim}/\tau)^{1/2}$, we can verify the correlation between $\epsilon$ and the reported performance of the Llama~\citep{touvron2023llama, touvron2023llama2openfoundation, dubey2024llama3} and Gemma models~\citep{gemmateam2024gemmaopenmodelsbased, gemmateam2024gemma2improvingopen} (see \cref{app:sample_complexity_exp} for the experimental details). We use the values of $\vocabsize$, $\embdim$ and $N_{\train}$ given in the technical reports of the open-source LLMs and plot the MMLU performance with respect to the approximation error $\epsilon$ in \cref{fig:mmlu_epsilon_separate}. This shows that our theory is model-specific as it results in distinctly different trends for the Llama and Gemma models. The difference is due to the larger $T$ of the Gemmas that compensates for the generally smaller embedding dimension $\embdim$ and leads to higher values of $\bar{B}$. The predicted approximation error correlates with the performance for each family of models: the larger the predicted error, the worse the models perform. Finally, we note that in this work we didn't estimate the norm of the Marton coupling as it requires access to the pre-training data. Studying this quantity and its approximation is of independent interest and we leave it for future work. 

\subsection{Pre-Training Generalization Bound}
\label{sec:pretraining}
Our main result on sample complexity is derived from the generalization bound that we present below.
\begin{figure*}[!t]
\label{fig:scaling_laws}
    \centering
    \begin{overpic}[width=0.32\textwidth]{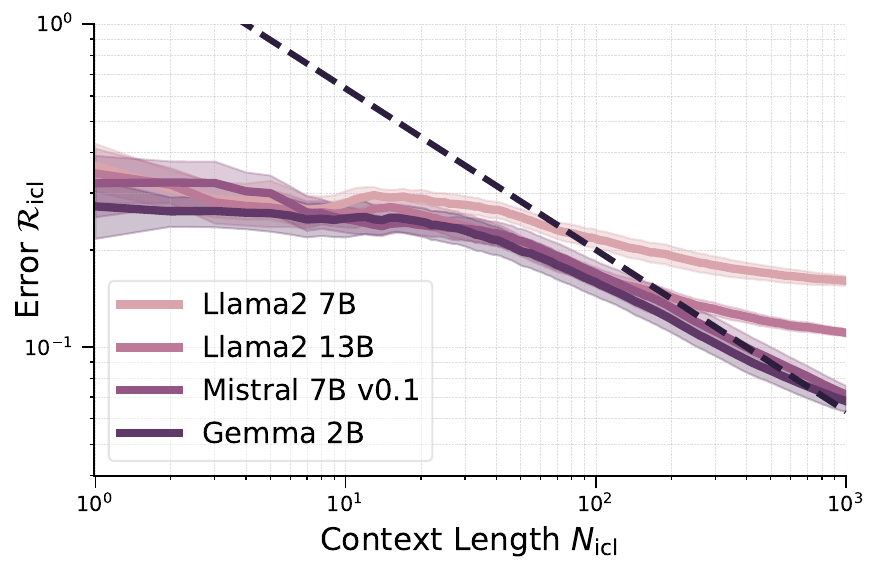}
    \put(42,57){\rotatebox{-33}{\tiny $\mcal{O}(N_{\ICL}^{-1/2})$}}
    \put(0,55){(a)}
    \end{overpic}
     \begin{overpic}[width=0.32\textwidth]{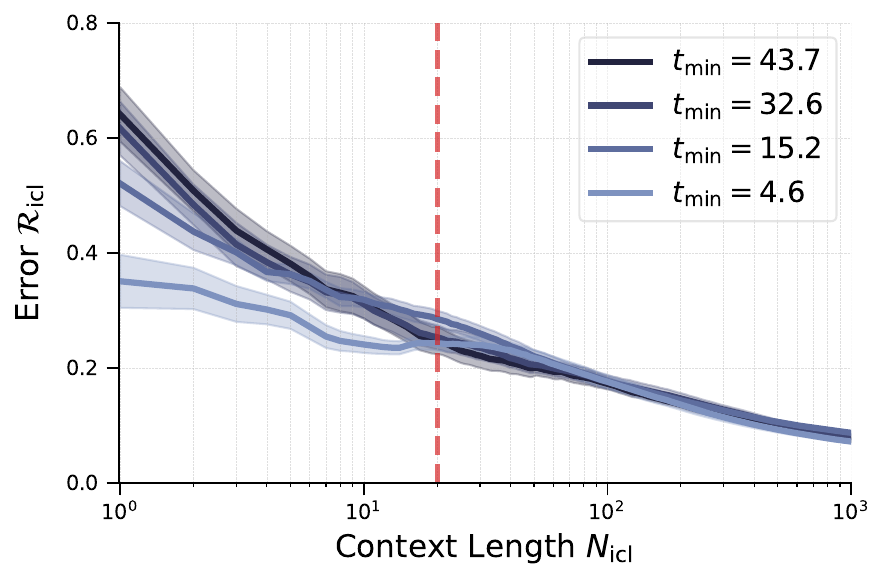}
     \put(19,14){\scriptsize Small $N_{\ICL}$}
     \put(57,14){\scriptsize Scaling law}
    \put(0,55){(b)}
     \end{overpic}
     \begin{overpic}[width=0.32\textwidth]{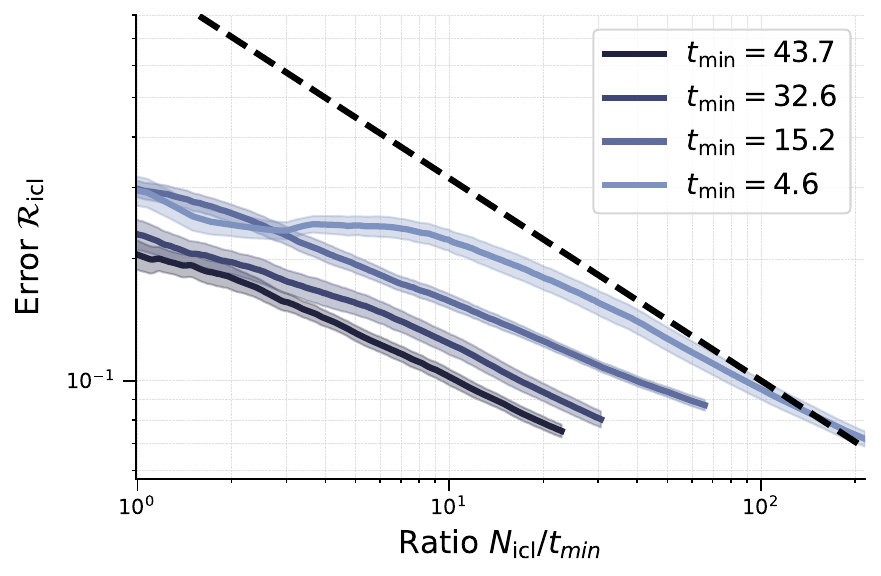}
     \put(30,62){\rotatebox{-33}{\tiny $\mathcal{O}(\sqrt{t_{\mathrm{min}}/{N_{\ICL}}})$}}
    \put(0,55){(c)}
     \end{overpic}
    \caption{\textbf{In-context scaling laws.} The risk $\mcal{R}_{\ICL}$ as functions of $N_{\ICL}$, with $95\%$ confidence intervals. (a) Risks for different LLMs along with the scaling law of \cref{thm:risk_bound_llm}. (b)-(c) Risks with \texttt{Mistral 7B v0.1} for random $3$-state transition matrices and different $t_{\mrm{min}}$ as functions of $N_{\ICL}$ and $N_{\ICL}/t_{\mathrm{min}}$.}
    \label{fig:icl_scaling_laws}
\end{figure*}

\noindent\textbf{Risk definition.\quad} We recall that for any $n \geq 1$, the true probability of next token $\mbf{X}_{n+1}$ given a past sequence $\mbf{S}_n$ is $\Pbl{\cdot}{\mbf{S}_n} \in \Delta_{\vocabsize}$ and the probability estimated by the model is denoted by $\Pbemp{\params}{\cdot}{\mbf{S}_n}$. Following the Markov chain formalization introduced in \cref{sec:mc_formalization}, we define the theoretical and empirical risks for any $\params \in \paramspace$ as
\begin{equation}
    \label{eq:estimation_error_def}
    \begin{split}
    &\risk{\params} \coloneqq \EE_{\mbf{S} \sim \PP_{\mcal{L}}} \mleft[ \TV{\langmat\mleft(\mbf{S}, \cdot\mright)}{\transmat\mleft(\mbf{S}, \cdot\mright)}\mright] \\
    &\smallrisk{\params} \coloneqq \frac{1}{N}\sum_{n=1}^N \TV{\Pbl{\cdot}{\mbf{S}_n}}{\Pbemp{\params}{\cdot}{\mbf{S}_n}}.
    \end{split}
\end{equation}
Formally, the expected risk can be written as 
\begin{align*}
\risk{\params} &= \EE_{\mbf{S} \sim \PP_{\mcal{L}}} \mleft[ \TV{\langmat\mleft(\mbf{S}, \cdot\mright)}{\transmat\mleft(\mbf{S}, \cdot\mright)}\mright] \\
&= \EE_{\mbf{S} \sim \PP_{\mcal{L}}} \mleft[ \TV{\Pbl{\cdot}{\mbf{S}}}{\Pbemp{\params}{\cdot}{\mbf{S}}}\mright] \\
&=\EE[\smallrisk{\params}].
\end{align*}
The generalization problem consists of bounding the difference $\smash{\risk{\params} - \smallrisk{\params}}$. 

\begin{rmk}[Choice of risk]
\label{rmk:choice_risk}
    Our risk definition departs from empirical risk minimization commonly used in statistical learning theory ~\citep{redko2019da, vapnik99overview, bach2024learning, marion2023generalization}. To assess how well the model estimates the probability \emph{distribution} of the next token, we follow~\citep{zhang2023whathowicl, hu2024statscot} and study the TV distance used in learning and identity testing of Markov chains literature~\citep{wolfer2023learning, wolfer2019minimax}. \cref{app:kl_thm} provides extended results with the KL divergence.
\end{rmk}

\noindent\textbf{Generalization bound.\quad} We denote the risks by $\risktrain{\params}$ and $\smallrisktrain{\params}$ to indicate that we take $N=N_{\train}$ in~\cref{eq:estimation_error_def}. Below, we state a generalization bound on the estimation risk of pre-training (see \cref{app:pre_training_risk_bound_llm} for the proof).
\begin{boxthm}[Pre-training generalization bound]
    \label{thm:pre_training_risk_bound_llm}
    Consider an LLM $\model_{\params} \in \funcspace$. We denote by $\mixmat$ the mixing matrix of the pre-training sequences of tokens $\MCseq{S}{N_{\train}}$. Let $0<\delta<1$, then with probability at least $1-\delta$,
    \begin{equation*}
        \risktrain{\params} \leq \smallrisktrain{\params} +  \frac{\bar{B}}{\sqrt{N_{\train}}}\sqrt{\log{\mleft(\frac{2}{\delta}\mright)}},
    \end{equation*}
    where $\bar{B}$ is the constant of \cref{cor:sample_complexity} and writes
    \[\bar{B}=2\lVert \mixmat \rVert\max\{\log{(\vocabsize)} + 2B_U/\tau, \log{(1/c_0)}\}^{1/2}.\]
\end{boxthm}
The obtained result has a desirable dependency on the amount of the pre-training data $\mathcal{O}(N_{\train}^{-1/2})$. Additionally, it showcases an interplay between the model architecture and the unknown reference constant $c_0$. If $B_U \approx \mcal{O}(T\sqrt{r})$ (due to the normalization of the unembedding layer), then the model's hidden dimension $r$ and vocabulary size $T$ should be large enough to ensure $\log(T) + 2B_U/\tau \geq \log(1/c_0)$. Below this threshold, the architecture of $\model_{\params}$ is not expressive enough to have any tangible impact on its generalization, although it may affect the training error $\smash{\smallrisktrain{\params}}$.
  
\subsection{In-Context Learning of Markov Chains}
\label{sec:icl}
Although insightful, the analysis presented above is related to the pre-training of LLMs -- a process that is hard and extremely costly to reproduce in practice. Similarly, we do not have access to the ground-truth matrix $\langmat$ to reason about LLM's ability to infer it in practice. To provide theoretical results that can be confirmed experimentally, we now turn our attention to ICL on Markov chains: a setup where one feeds a pre-trained LLM with an input sequence formed by a Markov chain of size $N_{\ICL}$ defined over a state space $\Omega$ of size $d$\footnote{This is different from another variation of ICL where supervised ($x$,$y$) pairs are provided in-context. Rather, the supervision is provided by observing transitions between states ($x_i$, $x_{i+1}=f(x_i)$) as discussed in \citep[Fig.1]{li2023transformers}.}. The risks $\riskicl{\params}$ and $\smallriskicl{\params}$ are then defined similarly to \cref{eq:estimation_error_def} where transition kernel $\PP$ of the Markov chain replaces the language reference distribution $\PP_{\mcal{L}}$ and we have $N = N_{\ICL}$ (see \cref{app:risk_bound_llm} for more details). To relate the generalization error to the pre-training error, we quantify the discrepancy between an LLM pre-trained mostly on textual data, and a hypothetical LLM with parameters in $\paramspace_{\mcdata}$ that is pre-trained on a dataset of Markov chains with the same data distribution as the Markov chain used as an input during in-context inference. We define the divergence between two models with weights $\params_1, \params_2$ and estimated probability distributions $\PP_{\params_1}, \PP_{\params_2}$ as
\begin{equation*}
    \prediv{\params_1}{\params_2}\! \coloneqq\! \frac{1}{N}\! \sum_{n=1}^N \Exp{\mbf{S}_n}{\TV{\Pbemp{\params_1}{\cdot}{\mbf{S}_n}}{\Pbemp{\params_2}{\cdot}{\mbf{S}_n}}}.
\end{equation*}
The operator $\mcal{K}$ is akin to a distance (the separation property is only verified almost surely, see \cref{sec:details_K} for more details). The next result, whose proof is deferred to \cref{app:risk_bound_llm}, provides a generalization bound on the in-context learning phase.
\begin{boxthm}[In-Context Learning generalization bound]
    \label{thm:risk_bound_llm}
    Consider an LLM $\model_{\params} \in \funcspace$. We provide as input of $\model_{\params}$ a $\mcsize-$state Markov chain $X = \MCseq{X}{N_{\ICL}}$. The sequence of subsequences of the first $n$ terms is denoted by $S = \MCseq{S}{n}$. $S$ is also a Markov chain, and we denote by $t_{\mrm{mix}}(\varepsilon)$ its mixing time. Let $\smash{t_\mrm{min} \coloneqq \inf_{0 \leq \varepsilon < 1} \tmix{\frac{\varepsilon}{2}} ( \frac{2 - \varepsilon}{1-\varepsilon})^2}$. Let $\delta > 0$. Then, with probability at least $1-\delta$,
    \begin{equation} 
    \label{eq_bound_ICL}
    \begin{split}
        \riskicl{\params} &\leq \inf_{\mcparams \in \mcparamspace} \{ \smallriskicl{\mcparams} + \mcal{K}(\mcparams, \params)\} \\
        &\qquad+ \bar{B}\sqrt{\frac{t_{\mrm{min}}}{N_{\ICL}}}\sqrt{\log{\mleft(\frac{2}{\delta}\mright)}},
    \end{split}
    \end{equation}
    with a constant depending on the problem's parameters 
    \[\bar{B}= 2\max\{\log{(\mcsize)} + 2B_U/\tau, \log{(1/\pmin)}\}^{1/2}.
    \]
\end{boxthm}
We note that instead of $||\mixmat||$ seen before, we now have an explicit dependency on $t_{\mrm{min}}$, which is related to the mixing time of the input Markov chain. This, together with the availability of the ground-truth transition matrix, allows us to use \cref{thm:risk_bound_llm} to verify experimentally the in-context learning scaling laws for popular LLMs. \cref{thm:risk_bound_llm} also suggests that an LLM pre-trained on diverse data sequences different from Markov chains should exhibit a certain degree of invariance to correctly infer the transition probabilities of the latter. This is reminiscent of the domain adaptation bounds~\citep{redko2019da} that commonly involve a distribution shift (\ie, a distance or a divergence) term that vanishes if the model is invariant to classes of transformations linking the distribution of the input data with that of test data. A recent success of LLMs in time series analysis~\citep{gruver2024large} suggests that this term may be small for certain types of data not used during pre-training. 

Finally, we note that \cref{thm:risk_bound_llm} implies that the LLM's ability to learn Markov chains exceeds the frequentist method\footnote{To the best of our knowledge, this is the only existing approach of Markov chains learning with theoretical guarantees.}~\citep{wolfer2019minimax} that consists of counting the occurrences of different states to fill in the matrix $\transmat$. This is experimentally confirmed in \cref{fig:LLM_vs_freq} of \cref{sec:experiment} (more details in \cref{app:sample_complexity_mc}).

\section{Numerical Experiments} \label{sec:experiment}
We now evaluate the ability of recent LLMs, namely \texttt{Mistral 7Bv0.1}~\citep{jiang2023mistral}, \texttt{Llama2 7B \& 13B}~\citep{touvron2023llama2openfoundation}, and \texttt{Gemma 2B}~\citep{gemmateam2024gemmaopenmodelsbased} to infer transition probabilities of Markov chains in-context based on \cref{thm:risk_bound_llm}. We associate each state in the $\mcsize$-state Markov chain with a token from the set $\{0,\ldots,d-1\}$, concatenated to obtain a prompt of length $N_{\ICL}$. We add comas whenever necessary to ensure that each state is tokenized separately. Details on the experimental setup and experiments with more Markov chains and \texttt{Llama3.2}~\citep{dubey2024llama3} are in \cref{app:tokenization}.

\noindent\textbf{Dependence on $N_{\ICL}$.\quad} We first analyze the effect of $N_{\ICL}$ on the risk calculated for a randomly generated $3$-state Markov transition matrix. From the results presented in \cref{fig:icl_scaling_laws} (left), we note that \texttt{Llama2} models deviate from our $\smash{\mcal{O}(N_{\ICL}^{-1/2})}$ theoretical scaling law, while most recent models (\texttt{Mistral} and \texttt{Gemma}) stay much closer to \cref{thm:risk_bound_llm}, similarly to what was observed by~\citet{cabannes2023scaling}. Being randomly generated, the Markov chains provided to the models have not been seen during training, and older (weaker) models naturally struggle to generalize.

\noindent\textbf{Dependence on $t_\mrm{min}$.\quad} \cref{thm:risk_bound_llm} states that Markov chains with slow mixing (higher $t_\mrm{min}$) are slower to learn. We now plot the true risk for a single model with different values of $t_\mrm{min}$ highlighting in \cref{fig:icl_scaling_laws} (right) a two-stage regime of ICL. In a first stage, the bound in \cref{eq_bound_ICL} is dominated by $\smash{\sqrt{t_{\mrm{min}}/N_{\ICL}}}$ for small $N_{\ICL}$, and depends strongly on $t_{\mrm{min}}$, while the scaling law $\smash{\mcal{O}(N_{\ICL}^{-1/2})}$ dominates as $N_{\ICL}$ increases beyond $N_{\ICL} \approx 20$.

\begin{figure}[htbp]
    \centering
    \begin{minipage}{.49\linewidth}
    \begin{overpic}[width=\linewidth]{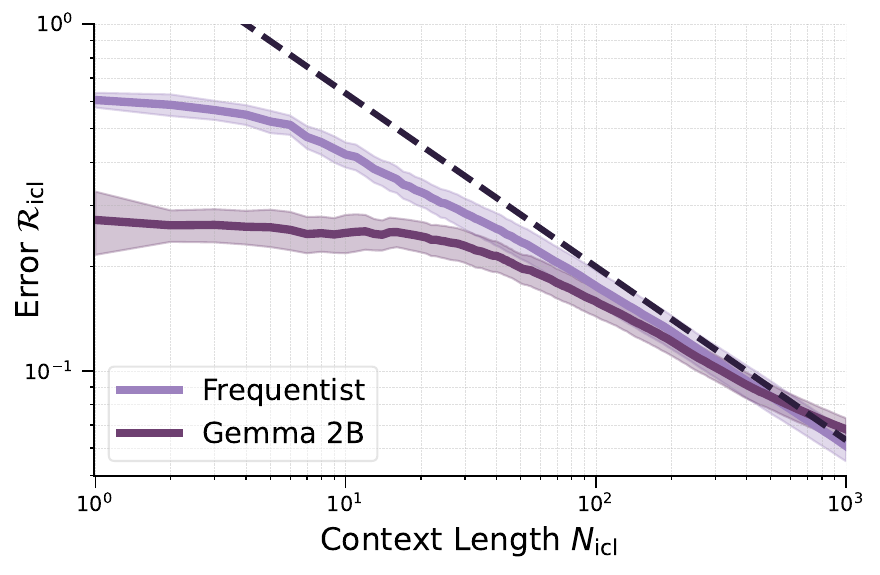}
        \put(72,35){\rotatebox{-33}{\small $\mcal{O}(N_{\ICL}^{-1/2})$}}
    \end{overpic}
    \end{minipage}
    \begin{minipage}{.49\linewidth}
    \begin{overpic}[width=\linewidth]{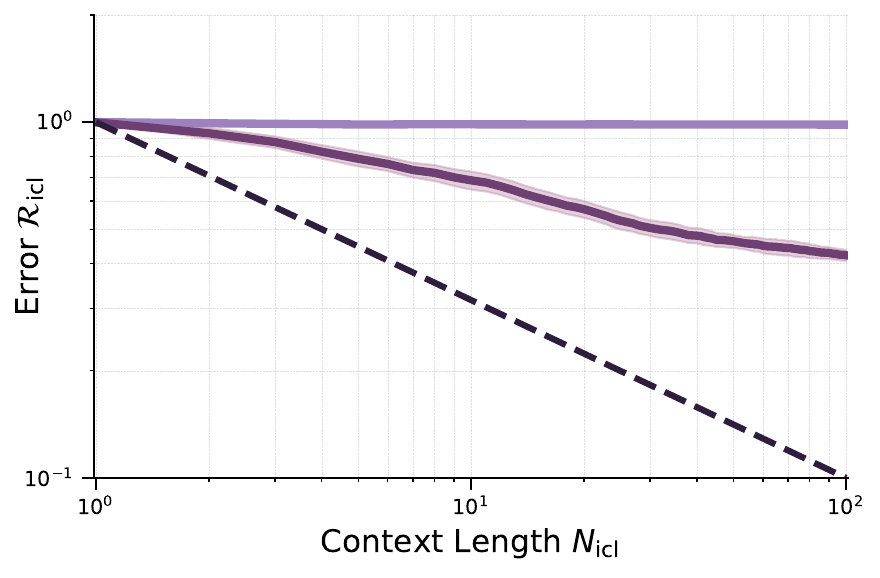}
            \put(68,28){\rotatebox{-25}{\small $\mcal{O}(N_{\ICL}^{-1/2})$}}
    \end{overpic}
    \end{minipage}
    \caption{\textbf{Impact of the number of states.} We plot the risks $\mcal{R}_{\ICL}$ as functions of $N_{\ICL}$ for \texttt{Gemma 2B} and the frequentist approach~\citep{wolfer2019minimax} with $95\%$ confidence intervals. \textbf{Left.} The input sequence is a random $3$-state Markov chain. \textbf{Right.} The input sequence is a Brownian motion discretized as a $700$-state Markov chain, similarly to~\citet{liu2024llmslearngoverningprinciples}.}
    \label{fig:LLM_vs_freq}
\end{figure}

\noindent\textbf{Dependence on $d$.\quad} We now verify \cref{thm:risk_bound_llm} for Markov chains with a different state space size (previously $d=3$). We also consider a baseline given by the frequentist method mentioned before. For the latter, its dependence on $d$ behaves like $\smash{\mcal{O}({\sqrt{d/N_{\ICL}}})}$ \citep{wolfer2019minimax}, while \cref{thm:risk_bound_llm} gives $\smash{\mcal{O}({\sqrt{\log(d)/N_{\ICL}}})}$. For Markov chains with a small number of states $d$, there is no clear difference between the frequentist estimator and a LLM. However, as $d$ grows the frequentist estimator struggles to estimate the transition matrix due to the $\mcal{O}(\sqrt{d})$ scaling factor. This is verified experimentally in \cref{fig:LLM_vs_freq}, where we vary the parameter $d$ from $3$ (left) to $700$ (right). We observe that the LLM follows the theoretical neural scaling law $\smash{\mcal{O}(N_\ICL^{-1/2})}$ and outperforms the frequentist method for $d=700$, while being close to it for $d=3$. We conclude that our analysis gives theoretical insights on the ICL neural scaling law observed empirically in~\citep{liu2024llmslearngoverningprinciples}. The additional experiments conducted in \cref{app:dynamical_systems} show that our bounds remain valid for large values of $d$. 

\section{Conclusion}
\label{sec:conclusion}
This paper proposed an explicit characterization of the inference mechanism in large language models through an equivalent finite-state Markov chain. We provided an insightful theoretical analysis based on the established characterization and the ability of the LLM to infer the transition kernel approximating the true transition probabilities of language. Experiments on commonly used LLMs validate our findings. We adapted our results to in-context learning where experiments confirm our theoretical insights. In the future, we hope that the proposed equivalence will have far-reaching implications on our understanding of LLMs and allow for a more fine-grained understanding of their expressiveness.
\section*{Impact Statement}
This paper presents work that aims to advance the field of Machine Learning. There are many potential societal consequences of our work, none of which must be specifically highlighted here.

\section*{Acknowledgements}
The authors would like to thank Alexander H\"agele for insightful feedback on the experiments with the large language models as well as Alain Durmus, Vasilii Feofanov, Giuseppe Paolo, Albert Thomas, Malik Tiomoko and Aladin Virmaux for fruitful discussions on early versions of this work. Nicolas Boull\'e was supported by the Office of Naval Research (ONR), under grant N00014-23-1-2729.
    
\bibliography{references}
\bibliographystyle{template/icml2025}

\clearpage
\onecolumn
\appendix
\textbf{\LARGE Appendix}

\paragraph{Roadmap.} In~\cref{app:notations}, we first recall our notations. We provide additional details on large language models and transformers in~\cref{app:background_transformer}. Important notions and definitions related to Markov chains and Marton couplings are given in \cref{app:background_mc}. The detailed proofs of our theoretical results are given in \cref{app:proofs}. Finally, we provide additional experiments in \cref{app:add_exp}. 

\addtocontents{toc}{\protect\setcounter{tocdepth}{2}}

\renewcommand*\contentsname{\Large Table of Contents}

\tableofcontents

\clearpage

\section{Notations}
\label{app:notations}
We denote $\{1, \cdots, N\}$ as $[N]$. We represent scalar values with regular letters (e.g., parameter $\lambda$), vectors with bold lowercase letters (e.g., vector $\mbf{x}$), and matrices with bold capital letters (e.g., matrix $\mbf{A}$). The $i$-th row of the matrix $\mbf{A}$ is denoted by $\mbf{A}_i$, its $j$-th
column is denoted by $\mbf{A}_{·,j}$ and its transpose is denoted by by $\mbf{A}^\top$. The identity
matrix of size $n$ is denoted by $\mbf{I}_n \in \RR^{n \times n}$. The vector of size $n$ with each entry equal to $1$ is denoted by $\bbm{1}_n$. We denote by $\lVert \mbf{A} \rVert _{p, q}$ the $L_{p, q}$ matrix norm where the $p$-norm is over columns and the $q$-norm is over rows. We denote by $\lVert \mbf{A} \rVert$ the operator norm of $\mbf{A}$ induced by the $\ell_2$ norm and by $\lVert \mbf{A} \rVert_{\infty} = \max_{ij} \lvert \mbf{A}_{ij}\rvert$ the operator norm induced by the $\ell_{\infty}$-norm. Similarly, $\mbf{x}^\top$ is the transpose of the vector $\mbf{x}$ and $\lVert \mbf{x} \rVert_p$ is its $\ell_p$-norm. The total variation between two probability distributions $\PP, \QQ$ is denoted by $\TV{\PP}{\QQ}$. The term \say{almost surely} is denoted by the notation \say{a.s.} while the term \emph{random variable} is denoted by the notation \say{r.v.}. $\Delta_n \coloneqq \{ \mbf{p} \in [0, 1]^n | \sum_{i=1}^n \mbf{p}_i = 1\}$ is the probability simplex of $\RR^n$. 

\section{Background on Large Language Models}
\label{app:background_transformer}
We first recall important notions regarding large language models before focusing on the most widely used ones, namely the transformer-based LLMs. We describe the components of the vanilla transformer architecture before describing the whole network at the heart of such a model and formally defining the class of parameters and neural networks considered in our work.

\subsection{Token Generation and Deletion Process}
In this section, we recall how the sequences of tokens are processed by the large language model notably regarding the next token generation and the deletion process.

\begin{boxdef}[Generation process]\label{def:generation_process}
    Given an input $s \in \mcal{V}_\cxtsize^*$ of size $p$, an large language model outputs a probability mass function $\model_{\params}^{\vocabsize,\cxtsize}(s)$ over the discrete vocabulary space. A next token $x$ is then sampled from $\model_{\params}^{\vocabsize,\cxtsize}(s)$, to construct a new sequence $(s,x)$ of size $p+1$.
\end{boxdef}

Generation can be repeated by considering $(s,x)$ as new input sequence and iterating this process. Since these models are designed to handle only sequences of size at most $\cxtsize$, a deletion process is required.

\begin{boxdef}[Deletion process]\label{def:deletion_process}
    Given an input $s$ of size $p > \cxtsize$, an large language model outputs a probability mass function $\model_{\params}^{\vocabsize,\cxtsize}(s_\cxtsize)$ where $s_\cxtsize$ is a truncation of $\cxtsize$ tokens of the sequence $s$. Large language models implement \textbf{front truncation}, which is done by setting $s_\cxtsize$ as the last $\cxtsize$ tokens of $s$.
\end{boxdef}

\begin{figure}[htbp]
    \centering
    \newsavebox{\boxA}
    \newsavebox{\boxB}
    \sbox{\boxA}{\includegraphics{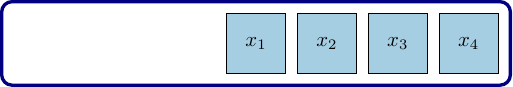}}
    \sbox{\boxB}{\includegraphics{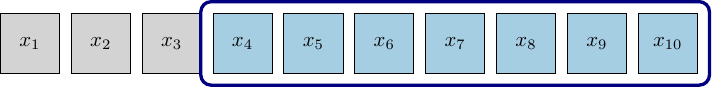}}
    \newlength{\bottomwidth}
    \setlength{\bottomwidth}{\wd\boxB}
    \makebox[\bottomwidth][r]{\usebox{\boxA}}
    \vspace{3mm}
    \rule{\bottomwidth}{0.5pt}
    \vspace{3mm}
    \usebox{\boxB}
    \caption{\textbf{Deletion process, front truncation}. A large language model with context window $\cxtsize = 7$ in navy blue, processing sequences of different lengths. \textbf{Top.} A sequence of length $4$.  \textbf{Bottom.} Front truncation of a sequence of length $10$.}
    \label{fig:context_window}
\end{figure}

As shown in \cref{fig:context_window}, only the last $\cxtsize$ tokens of a long input sequence are used. This is why we speak of \textit{deletion}, since we ignore the first tokens.

\begin{figure}[htbp]
    \centering

    \includegraphics{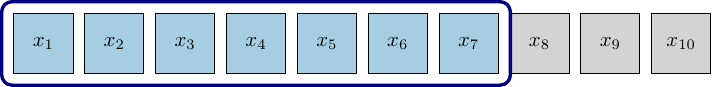}

    \caption{\textbf{Back truncation}. A large language model with context window $\cxtsize = 7$ in navy blue, processing back truncation of a sequence of a sequence of length $10$.}
    \label{fig:back_truncation}
\end{figure}

Note that it is possible to implement other kinds of truncation, but large language models usually do not \citep{brown2020LLMfewshot,touvron2023llama}, however, in models like BERT \citep{devlin2019bert}, which are not autoregressive, back truncation as described in \cref{fig:back_truncation} is also an option.

\subsection{Autoregressive Transformer-Based LLMs}
The most popular autoregressive LLMs rely on the transformer architecture~\citep{vaswani2017attention} which we describe below following~\citep{brown2020LLMfewshot, edelman2022inductive, zhang2023whathowicl}. An autoregressive transformer-based LLM takes as input a sequence of length $\inputsize$, with $\inputsize \leq \cxtsize$ and $\cxtsize$ is the context window, tokens with values in a vocabulary $\vocabspace$ of size $\vocabsize$. The tokens are embedded into a $\embdim$-dimensional space and the input can be written as $\mbf{S} \in \RR^{\embdim \times \inputsize}$. We consider a transformer model with $L$ layers and $h$ heads. The output of the $\ell$-th layer writes $\mbf{S}^{(\ell)}$ and is fed as input of the $(\ell+1)$-th layer. The input of the whole model is $\mbf{S}^{(0)} = \mbf{S}$. Below, we describe the operations performed by the model, including the embeddings of the tokens.

\begin{itemize}[leftmargin=*]
    \item \textbf{Token embeddings.} The tokens are embedded in a $\embdim$-dimensional space via an embedding layer $\mbf{W}_E$ which results in an input of the form $\mbf{S}^{\embdim \times \inputsize}$;
    \item \textbf{Positional embeddings.} (Learnable) positional embeddings are added to each token depending on its position in the input sequence. This breaks the permutation-invariance of the transformer architecture and leads, by abuse of notation, to an output $\mbf{S} \in \RR^{\embdim \times \inputsize}$;
    \item \textbf{Multi-head attention (MHA).} Given an input sequence $\mbf{S} \in \RR^{\embdim \times \inputsize}$, query, key, and value matrices $\mbf{W}_Q, \mbf{W}_K, \mbf{W}_V\in \RR^{\embdim \times \embdim}$ (here the value and output matrices are merged for ease of notations), the self-attention module computes
          \begin{equation*}
              \mcal{A}\mleft(\mbf{S}; \mbf{W}_Q, \mbf{W}_K, \mbf{W}_V\mright) \coloneq \softmax{\mbf{W}_Q\mbf{S}\mleft( \mbf{W}_K\mbf{S}\mright)^\top/ \sqrt{\embdim}}\mleft(\mbf{W}_V\mbf{S}\mright) \in \RR^{\embdim \times \inputsize},
          \end{equation*}
          with $\mrm{softmax} \colon \mbf{x} \in \RR^n \to \exp(\mbf{x}) / \sum_i\exp(\mbf{x})_i \in \Delta_n$. The operation described below corresponds to single-head self-attention. In practice, multi-head attention (MHA) is used with $H$ heads and the query and key matrices are in $\RR^{\frac{\embdim}{H} \times \frac{\embdim}{H}}$ and the value matrix is in $\RR^{\frac{\embdim}{H} \times \embdim}$ ($\embdim, H$ are taken such that $\frac{\embdim}{H}$ is an integer). The MHA module concatenates on the row dimension the outputs of $\mcal{A}$ for each head and then projects it back to the embedding dimension $\embdim$ with an output matrix $\mbf{W}_O \in \RR^{\embdim \times \embdim}$. By abuse of notation, we also denote by $\mcal{A}$ this operation which results in an output of dimension $\embdim \times \inputsize$, and we include the output matrix in the argument of the operator. The $\ell$-th layer of the transformer applies attention with layer-specific weight matrices and a residual connection that leads to an output
          \begin{equation*}
              \mbf{Z}^{(\ell)} = \mbf{S}^{(\ell-1)} +  \mcal{A}\mleft(\mbf{S}^{(\ell-1)}; \mbf{W}_Q^{(\ell)}, \mbf{W}_K^{(\ell)}, \mbf{W}_V^{(\ell)}, \mbf{W}_O^{(\ell)} \mright).
          \end{equation*}
          This is followed by a layer normalization~\citep{zhang2019rms} that projects each token into the $\ell_2$-unit ball, i.e., each column $\mbf{S}^{(\ell)}_{\cdot, n}$ has an $\ell_2$-norm lower than $1$;
    \item \textbf{Feed-forward block (FF).} Finally, a feed-forward block is applied, consisting of a two-layer MLP with hidden dimension $\hiddendim$, layer-specific weight matrices $\mbf{W}_1^{(\ell)} \in \RR^{\hiddendim \times \embdim}, \mbf{W}_2^{(\ell)} \in \RR^{\embdim \times \hiddendim}$ and ReLU activation denoted by $\relu{x} = \max\{0, x\}$ and applied entry-wise. The output of this layer reads
          \begin{equation*}
              \mbf{Y}^{(\ell)} = \mbf{W}_2^{(\ell)} \relu{\mbf{W}_1^{(\ell)}\mbf{Z}^{(\ell)}}.
          \end{equation*}
          It is followed by a residual connection to produce the output
          \begin{equation*}
              \mbf{S}^{(\ell)} =\mbf{Z}^{(\ell)} +  \mbf{W}_2^{(\ell)} \relu{\mbf{W}_1^{(\ell)}\mbf{Z}^{(\ell)}} \in \RR^{\embdim \times \inputsize},
          \end{equation*}
          on which layer normalization~\citep{zhang2019rms} is applied ensuring that each column $\mbf{S}^{(\ell)}_{\cdot, n}$ has an $\ell_2$-norm lower than $1$.
    \item \textbf{softmax output layer.} In the autoregressive setting, the model outputs a probability distribution on the vocabulary $\vocabspace$. To that end, the output $\mbf{S}^{(L)} \in \RR^{\embdim \times \inputsize}$ of the final layer is projected back to the vocabulary size by an \say{unembedding layer} $\mbf{W}_U \in \RR^{\vocabsize \times \embdim}$ and averaged over the columns to obtain a vector in $\RR^{\vocabsize}$. A softmax layer is finally applied on top of it to obtain the probability distribution of the next token $\Pbemp{\params}{\cdot}{\mbf{S}}$. Formally, we have
          \begin{equation*}
              \Pbemp{\params}{\cdot}{\mbf{S}} = \mrm{softmax}\mleft( \frac{1}{\inputsize \tau}\mbf{W}_U \mbf{S}^{(L)} \bbm{1}_\inputsize\mright) \in \Delta_{\vocabsize},
          \end{equation*}
          $n$ is the length (i.e., number of columns) of the input sequence $\mbf{S}$  (and thus of the last layer output $\mbf{S}^{(L)}$), $\params$ denotes the parameters of the whole network that subsume the parameters of each layer and each block and $\tau$ is the softmax temperature~\citep{hinton2015distilling}.
\end{itemize}

\paragraph{Theory faithful to the practice.} The architecture described above is used in most of the transformer-based autoregressive LLM~\citep{brown2020LLMfewshot, dubey2024llama3, google2024gemini, jiang2023mistral}. In the theoretical analysis of \cref{sec:theoretical_analysis}, and unless specified otherwise, we remain faithful to their practical implementation and only make the following mild assumption: we assume that the unembedding layer is bounded. The class of parameters and the class of neural networks it generates respectively writes
\begin{equation*}
    \paramspace = \{ \params \mid \lVert \mbf{W}_U^\top \rVert_{2, 1} \leq B_U\} \quad \text{ and } \quad \funcspace = \{ \model_{\params} \mid \params \in \paramspace \}.
\end{equation*}
It should be noted that this assumption is significantly weaker than what is usually done in the literature~\citep{zhang2023whathowicl, edelman2022inductive}.

\section{Background on Markov Chains}
\label{app:background_mc}
We recall below some important notions related to Markov chains based on~\citep{robert2004MC, paulin2015ineqMC} and that will be used in our proofs.

\subsection{Basic Notions}

Consider two distribution probabilities $\PP$ and $\QQ$ defined on a measurable space $\measurable$.

\begin{boxdef}
    The total variation between $\PP$ and $\QQ$ is defined as
    \begin{equation*}
        \TV{\PP}{\QQ} \coloneqq \sup_{A \in \mcal{F}} \lvert \PP\mleft(A\mright) - \QQ\mleft(A\mright) \rvert.
    \end{equation*}
\end{boxdef}

In the setting considered in the main paper, we consider Markov chains with finite discrete state space $\polish$. In this section, we refer to $\polish$ as a general Polish space, whose elements are referred to as \textit{states}.

Informally, a discrete-time, time-homogeneous Markov chain with state space $\polish$ is a sequence of random variables $\MC{X}$ taking values in $\polish$ such that the next observation is independent of the past given the present. This property is referred to as the Markov property and is defined below.
\begin{boxdef}
    A sequence of random variables $\MC{X}$ is said to satisfy the Markov property if for all $n \geq 1$ and any $\mleft(x_1, \ldots, x_{n+1}\mright) \in \polish^{n+1}$
    \begin{equation*}
        \Pb{\mbf{X}_{n+1} = x_{n+1}}{\mbf{X}_n = x_n, \cdots, \mbf{X}_1 = x_1} = \Pb{\mbf{X}_{n+1} = x_{n+1}}{\mbf{X}_n = x_n}.
    \end{equation*}
\end{boxdef}

To a given Markov chain, we associate its \emph{transition kernel} $\mathbf{Q}:\polish^2 \to  [0,1]$ which collects the transition probabilities from one state to another
\begin{equation*}
    \forall n \in \NN, (x, y) \in \polish^2, \quad \mathbf{Q}(x, y) = \Pb{\mbf{S}_{n+1} = y}{\mbf{S}_n = x}.
\end{equation*}
In the main text, we refer to $\mbf{Q}$ as a transition \emph{matrix} as the Markov chains we consider are of finite state space.

\begin{boxdef}
    A distribution $\pi$ on $\polish$ is said to be a stationary distribution if the action of the transition kernel leaves $\pi$ unchanged, that is
    \begin{equation*}
         (\pi\mathbf{Q})(A) := \int_{y \in A} \mathbf{Q}(x,y)d\pi(x) = \pi(A)
    \end{equation*}
    for all $A \in \mathcal{F}$.
\end{boxdef}

A natural question is whether such a distribution exists for a generic Markov chain. Before stating an existence theorem, we introduce a classification of states below.

\paragraph{Class of states.} All definitions below are borrowed from~\citep{gallager}

\begin{boxdef}[Accessibility and communication]
    \label{def:acc_comm}
    A state $x$ is accessible from $y$ (abbreviated as $x\rightarrow y$) if there exists $n>0$ such that $\mathbf{Q}^n(x,y) >0$.
    Two distinct states $x$ and $y$ communicate (abbreviated $x\leftrightarrow y$) if $x$ is accessible from $y$ and $y$ is accessible from $x$.
\end{boxdef}

Accessibility and communication concepts define how states can reach each other within a Markov chain. This leads to an important classification of states into transient and recurrent categories.

\begin{boxdef}[Recurrent and transient states]
    \label{def:rec_trans}
    For finite-state Markov chains, a recurrent state is a state $i$ that is accessible from all states that are accessible from $i$ ($i$ is recurrent if $i\rightarrow j$ implies that $j\rightarrow i$).
    A transient state is a state that is not recurrent.
\end{boxdef}

With the distinction between recurrent and transient states established, we can now group states into classes based on their communication properties.

\begin{boxdef}[Class of states]\label{def:class}
    A class $\class$ of states is a non-empty set of states such that each $i\in\class$ communicates with every other state $j\in\class$ and communicates with no $j\notin\class$
\end{boxdef}

\paragraph{Aperiodicity and Ergodicity.}

Aperiodicity ensures that the system does not exhibit cyclic behavior, which is a key condition for understanding the asymptotic behavior of states.

\begin{boxdef}[Aperiodicity]
    \label{def:aperiodicity}
    The period of a state $i$, denoted $d(i)$, is the greatest common divisor (gcd) of those values of $n$ for which $\mathbf{Q}^n(i,i) >0$.
    If the period is $1$, the state is aperiodic.
\end{boxdef}

Under some conditions on the Markov chain (aperiodicity and irreducibility~\citep{robert2004MC}), it can be proven that the chain converges to its stationary distribution i.e. for any $x \in \polish$, $\lim_{n \to \infty} \TV{Q^n\mleft(x, \cdot\mright)}{\pi} = 0$, where $Q^n\mleft(x, \cdot\mright)$ denotes the probability of $\mbf{S}_n$ conditioned on $\mbf{S}_1 = x$.

We recall below the notion of mixing time that assesses the time taken by the Markov chain to be $\varepsilon$-close to its stationary distribution (see \cref{def:mixing_time}).

\begin{boxdef}[Mixing time for time-homogeneous Markov chains~\citep{paulin2015ineqMC}]
    \label{def:mixing_time}
    Let $X \coloneqq \MC{X}$ be a time-homogeneous Markov chain with a state space $\polish$, a transition kernel $Q$, and a stationary distribution $\pi$. Its mixing time is defined for any $\varepsilon \in [0, 1]$ as
    \begin{equation*}
        \tmix{\varepsilon} \coloneqq \min{\{t \mid d(t) \leq \varepsilon\}} \text{ where } d(t) \coloneqq \sup_{x \in \polish} \TV{Q^t\mleft(x, \cdot\mright)}{\pi}.
    \end{equation*}
\end{boxdef}

We also introduce the quantity
\begin{equation*}
    t_\mrm{min} \coloneqq \inf_{0 \leq \varepsilon < 1} \tmix{\frac{\varepsilon}{2}} \cdot \mleft( \frac{2 - \varepsilon}{1-\varepsilon}\mright)^2
\end{equation*}
which will be useful later on.

\begin{rmk}[Well-posedness of $t_\mrm{min}$]
    \label{rmk:t_min}
    As we only consider finite state-space Markov chains in our work, we know that a stationary distribution always exists. However, its uniqueness and the convergence to it require additional assumptions (see \cref{app:background_ergodic_unichain}). In particular, not all Markov chains admit a finite $t_{\mrm{mix}}(\varepsilon)$, $t_\mrm{min}$ for some $\varepsilon < \frac{1}{2}$. In such case, $t_\mrm{min}$ can be infinite. In our practical experimentation, this is never the case despite considering various Markov chains.
\end{rmk}

\subsection{Ergodic Unichains}
\label{app:background_ergodic_unichain}

We are now ready to state the following theorem, which formalizes the classification of states into recurrent, transient, and aperiodic classes.

\begin{boxthm}[Recurrent and transient classes]\label{thm:rec_trans_class}
    For finite state Markov chains, either all states in a class are transient or all are recurrent. We refer to these classes as transient and recurrent, respectively.

    For any Markov chain, all states in a class have the same period. If the period is 1, the class is said to be aperiodic
\end{boxthm}

Having categorized states into recurrent, transient, and aperiodic classes, we can now define ergodicity.

\begin{boxdef}[Ergodicity]\label{def:ergodicity}
    For a finite-state Markov chain, an ergodic class of states is a class that is both recurrent and aperiodic. A Markov chain consisting entirely of one ergodic class is
    called an ergodic chain.
\end{boxdef}

\paragraph{Unichains.} We now introduce the concept of unichains.

\begin{boxdef}[Unichains and ergodic unichains]
    \label{def:unichain}
    A unichain is a finite-state Markov chain containing a single recurrent class and transient states.
    An ergodic unichain is a unichain for which the recurrent class is ergodic.
\end{boxdef}

\subsection{Marton Couplings}
\label{app:background_marton}

While we consider Markov chain inputs in \cref{sec:icl}, we consider less structured inputs during the pre-training phase \cref{sec:pretraining}.

More specifically, we model the sequences of tokens used during pre-training as generic dependent random variables. To derive meaningful results, we rely on the notion of Marton couplings introduced by~\citet{marton2003coupling}. A Marton coupling can be seen as a weak dependency structure between random variables. The associated notion of the mixing matrix, analogous to the mixing time of a Markov chain, is used to assess the strength of the dependence between those variables.

This minimal modeling choice is made to remain as faithful as possible to the pre-training considered in practical applications of LLMs, for which the pre-training data is not public and may contain arbitrary data points such as video, code snippets, text and images~\citep{dubey2024llama3, touvron2023llama, google2024gemini, jiang2023mistral, openai2024gpt4technicalreport, brown2020LLMfewshot}.

As shown in~\citet[Remark 2.2.]{paulin2015ineqMC}, considering sequences of random variables linked through a Marton coupling is a weaker assumption than what is usually done in the literature on generalization bounds, which typically relies on independent random variables and Markov chains~\citep{wolfer2019minimax,zhang2023whathowicl,hu2024statscot, marion2023generalization}.

In particular, the results stated in \cref{sec:pretraining} encompass the case where the pre-training input sequences of tokens are independent random variables~\citep{kim2024transformersminimaxoptimalnonparametric} or Markov chains~\citep{zhang2023whathowicl}. We also note that Markov chains can model bigrams used in natural language~\citep{jurafski2024nlp, bietti2023birth}. We do not provide an exhaustive review of Marton couplings. We will simply recall its definition and introduce the associated mixing matrix. We refer the interested reader to~\citet{marton2003coupling} and~\citet{paulin2015ineqMC}. Consider a sequence of dependent random variables $S = \MCseq{S}{N}$ taking values in a polish space $\polish = \polish_1 \times \ldots \times \polish_N$. We will denote by $\PP\MCseq{S}{N}$ the distribution of $S$.

\begin{boxdef}[Marton coupling]
    \label{def:marton}
    We define a \emph{Marton coupling} for $S$ as a set of couplings
    \[\left(S^{(s_1,\ldots,s_i,s_i')}, {S'}^{(s_1,\ldots,s_i,s_i')}\right) \in \Omega\times \Omega,\]
    for every $i\in [N]$, every $s_1\in \Omega_1,\ldots,s_i\in \Omega_i, s_i'\in \Omega_i$, satisfying the following conditions.
    \vspace{2mm}
    \begin{enumerate}[leftmargin=*, label=(\roman*)]
        \item $\begin{aligned}[t]\mbf{S}^{(s_1,\ldots,s_i,s_i')}_{1} & =s_1, \hspace{20mm} \ldots, \quad                                               & \mbf{S}^{(s_1,\ldots,s_i,s_i')}_{i}=s_i,   \\
               \mbf{S}'^{(s_1,\ldots,s_i,s_i')}_{1}   & =s_1,  \quad \ldots, \quad \mbf{S}'^{(s_1,\ldots,s_i,s_i')}_{i-1}=s_{i-1},\quad & \mbf{S}'^{(s_1,\ldots,s_i,s_i')}_{i}=s_i'.\end{aligned}$
              \vspace{2mm}
        \item $\begin{aligned}[t] & \left(\mbf{S}^{(s_1,\ldots,s_i,s_i')}_{i+1},\ldots, \mbf{S}^{(s_1,\ldots,s_i,s_i')}_{N}\right)                 \\
                   & \sim \PP(\mbf{S}_{i+1},\ldots,\mbf{S}_{N}\mid\mbf{S}_1=s_1,\ldots,\mbf{S}_i=s_i),                              \\
                   & \left({\mbf{S}'}^{(x_1,\ldots,x_i,x_i')}_{i+1},\ldots, {\mbf{S}'}^{(x_1,\ldots,x_i,x_i')}_{N}\right)           \\
                   & \sim \PP(\mbf{S}_{i+1},\ldots, \mbf{S}_{N}\mid\mbf{S}_1=x_1,\ldots,\mbf{S}_{i-1}=x_{i-1}, \mbf{S}_{i}=x_{i}').
                  \end{aligned}$
              \vspace{2mm}
        \item If $x_i=x_i'$, then $S^{(x_1,\ldots,x_i,x_i')}=S'^{(x_1,\ldots,x_i,x_i')}$.
    \end{enumerate}
\end{boxdef}

\begin{boxdef}[Mixing matrix~\citep{paulin2015ineqMC}]
    \label{def:coupling_matrix}
    For a Marton coupling, we define
    the \emph{mixing matrix} $\mixmat \in \RR^{N \times N}$ as an upper diagonal matrix with
    \begin{equation*}
        \forall 1 \leq i < j \leq N, \quad
        \begin{cases}
             & \mixmat_{i,i}:=1, \quad                   \\
             & \mixmat_{j,i}:=0 \quad                    \\
             & \mixmat_{i,j}:=\sup_{s_1,\ldots,s_i,s_i'}
            \PP\left[\mbf{S}^{(s_1,\ldots,s_i,s_i')}_{j}\ne {\mbf{S}'}^{(s_1,\ldots,s_i,s_i')}_{j}\right]
        \end{cases}.
    \end{equation*}
\end{boxdef}
For independent random variables, one can define a Marton coupling with a mixing matrix equal to the identity~\citep[see][Remark 2.2]{paulin2015ineqMC}. In particular, it means that for independent variables, we have the operator norm of the mixing matrix equal to $1$, i.e., $\lVert \mixmat \rVert = 1$.

\subsection{An (Almost) Distance between Markov Chains}
\label{sec:details_K}
In \cref{thm:restate_icl_risk_bound_llm},
We state elementary properties of $\mcal{K}$ in the proposition below.
\begin{boxprop}[Properties of $\mcal{K}$]
    \label{prop:properties_K}
    $\mcal{K}$ is an \emph{almost}-distance between transition matrices in the sense that it satisfies the properties below:
    \begin{enumerate}[topsep=0pt,itemsep=3pt,parsep=3pt,leftmargin=15pt]
        \item \textbf{Non-negativity}. For any $\params_1, \params_2$, $\prediv{\params_1}{\params_2} \geq 0$.
        \item \textbf{Almost sure positivity}. $\prediv{\params_1}{\params_2} = 0\iff \forall n \in [N], \Pbemp{\params_1}{\cdot}{\mbf{S}_n}= \Pbemp{\params_2}{\cdot}{\mbf{S}_n} \text{ a.s.}$.
        \item \textbf{Symmetry}. For any $\params_1, \params_2$, $\prediv{\params_1}{\params_2} =  \prediv{\params_1}{\params_2}$.
        \item \textbf{Triangular inequality.}. For any $\params_1, \params_2, \params_3$, $\prediv{\params_1}{\params_3} \leq  \prediv{\params_1}{\params_2} + \prediv{\params_2}{\params_3} $.
    \end{enumerate}
\end{boxprop}
\begin{proof}[Proof of \cref{prop:properties_K}]
    We first recall the following technical lemma.
    \begin{boxlem}[Proposition~2.16 in \cite{folland1999real}]
        \label{lem:nonneg_rv}
        Let $Y$ be a non-negative random variable defined on a probability space $\Omega$ with probability function $\PP$. If $\EE\mleft[ Y \mright] = 0$, then $Y=0$ almost surely, i.e., $$\PP \mleft( \{ \omega \in \Omega \mid Y(\omega) = 0\} \mright) = 1$$
    \end{boxlem}

    The non-negativity and symmetry of $\mcal{K}$ directly come from the symmetry and non-negativity of the total variation distance. The triangular inequality follows from the fact that the total variation is a distance and that the expectation respects inequalities. For the almost positivity, consider $\params_1, \params_2$ such that $\prediv{\params_1}{\params_2} = 0$. By non-negativity of all the terms in the sum, it means that for all $n \in [N]$, we have
    \begin{equation*}
        \Exp{\mbf{S}_n}{\TV{\Pbemp{\params_1}{\cdot}{\mbf{S}_n}}{\Pbemp{\params_2}{\cdot}{\mbf{S}_n}}} = 0.
    \end{equation*}
    As the total variation is a distance, we know that the random variable under the expectation is non-negative. Applying \cref{lem:nonneg_rv} leads to
    \begin{equation*}
        \TV{\Pbemp{\params_1}{\cdot}{\mbf{S}_n}}{\Pbemp{\params_2}{\cdot}{\mbf{S}_n}} = 0 \quad \text{ almost surely}.
    \end{equation*}
    On the probability space, deprived of the set where the distance is non-zero (which is of null measure), the total variation is equal to zero and as a distance between probability distributions, it means that on this subset of the probability space, the probabilities are equal. Again, as the set on which they are not equal is of null measure, we have
    \begin{equation*}
        \Pbemp{\params_1}{\cdot}{\mbf{S}_n} = \Pbemp{\params_2}{\cdot}{\mbf{S}_n} \quad \text{ almost surely}.
    \end{equation*}
    Putting everything together, we have
    \begin{equation}
        \label{eq:prob_equal_all}
        \forall n \in [N], \Pbemp{\params_1}{\cdot}{\mbf{S}_n} = \Pbemp{\params_2}{\cdot}{\mbf{S}_n} \quad \text{ a.s.},
    \end{equation}
    which concludes the direct sense. The converse sense is proved by assuming that Eq.~\eqref{eq:prob_equal_all} holds and using the distance properties of the total variation. This concludes the proof.
\end{proof}

\section{Additional Experiments}
\label{app:add_exp}
In this section, we present additional numerical experiments that confirm that our theory correctly predicts the in-context learning behavior of LLMs.

\subsection{Experimental Setup and Tokenization}
\label{app:tokenization}

\paragraph{Experimental setup.} To ensure a fair validation of our theoretical results, we conduct our experiments on some of the most recent and widely used LLMs: \texttt{Gemma 2B}~\citep{gemmateam2024gemmaopenmodelsbased}, \texttt{Llama2 7B \& 13B}~\citep{touvron2023llama2openfoundation}, \texttt{Llama3 8B}, \texttt{Llama3.2 1B \& 3B}~\citep{dubey2024llama3}, and \texttt{Mistral 7Bv0.1}~\citep{jiang2023mistral}.

\paragraph{Tokenization.} As the models we consider have different tokenizations, we need to do this step with extra care as it is a crucial part of the experimental procedure.
Indeed, LLMs' ability to handle numerical values has been proved to be dependent on the tokenization algorithm \citep{singh2024tokenization, ali2024tokenizer, gruver2024large}.
The most widely used tokenization algorithm to-date, \emph{BPE} \citep{sennrich2016neural}, tends to assign tokens to arbitrary $3$-digits numbers based on their occurrences in large-scale corpora, and the tokenizer's vocabulary size. As highlighted by \citep{gruver2024large}, this artifact severely hinders LLMs' ability to predict numerical values in-context.
This is the case for popular LLMs such as \texttt{GPT-3} \citep{brown2020LLMfewshot}. Newer models (\texttt{LLama3}, \texttt{GPT-3.5}, \texttt{GPT-4}) however, tend to have hard-coded rules on top of \emph{BPE}, making them able to encode all $3$-digits numbers with their own token. Although this feature would accelerate the ICL procedure by eliminating the need for the \emph{Hierarchy-PDF} algorithm in \citep{liu2024llmslearngoverningprinciples},the under-representability of larger numbers in the training data could be an issue. Other tokenization techniques that are numerical values-focused has been presented in the literature \citep{golkar2023xval, wu2024pre}, paving the way for another research direction that may benefit our method.

\paragraph{Rodmap.} In the rest of this section, we extend our experiments to study the following setups:
\begin{itemize}
    \item In \cref{app:markov_scaling}: impact of the number of states $\mcsize$;
    \item In \cref{app:exp_gen_mc}: extension to Markov chains with $\pmin=0$;
    \item In \cref{app:2024models}: impact of the tokenization;
    \item In \cref{app:dynamical_systems}: extension to dynamical systems.
\end{itemize}

\subsection{Impact of the Number of States $\mcsize$}
\label{app:markov_scaling}

\begin{figure}[htbp]
    \centering
    \includegraphics[width=0.4\textwidth]{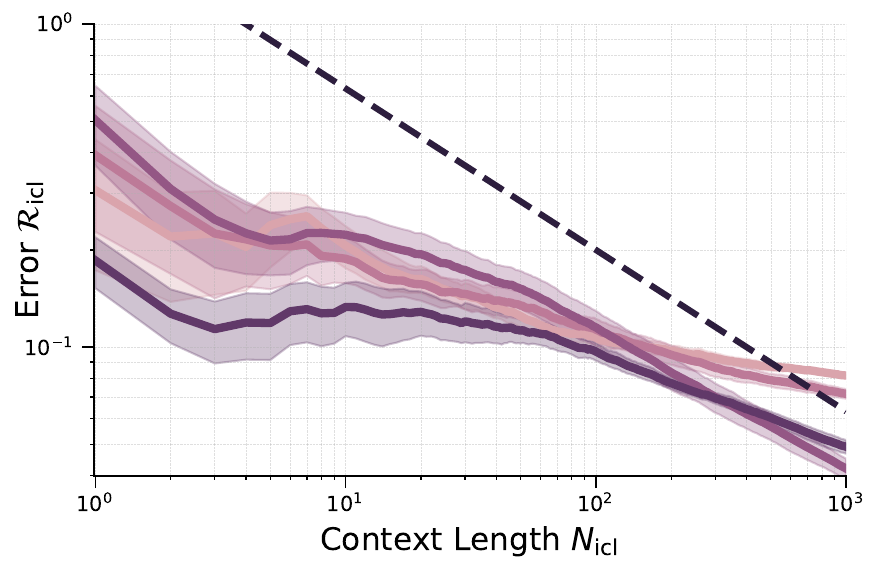}
    \includegraphics[width=0.4\textwidth]{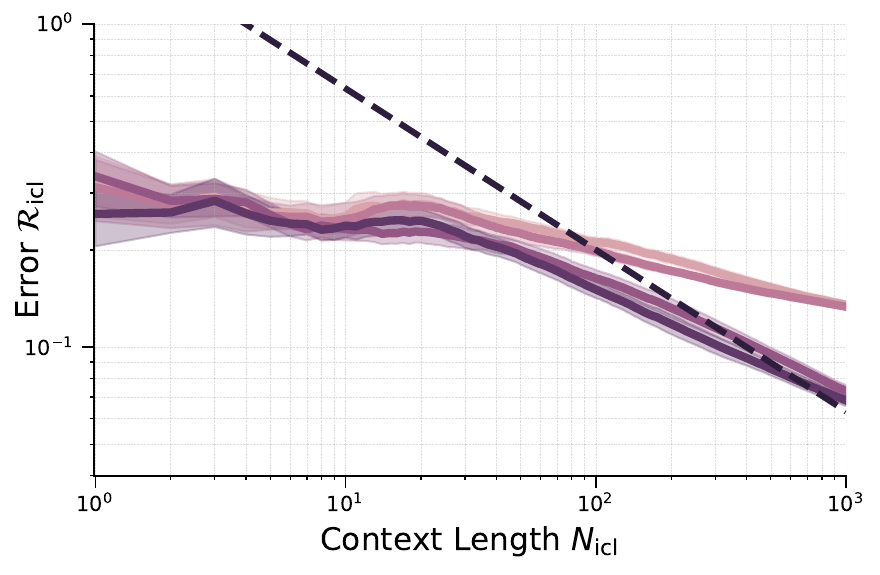}
    \includegraphics[width=0.4\textwidth]{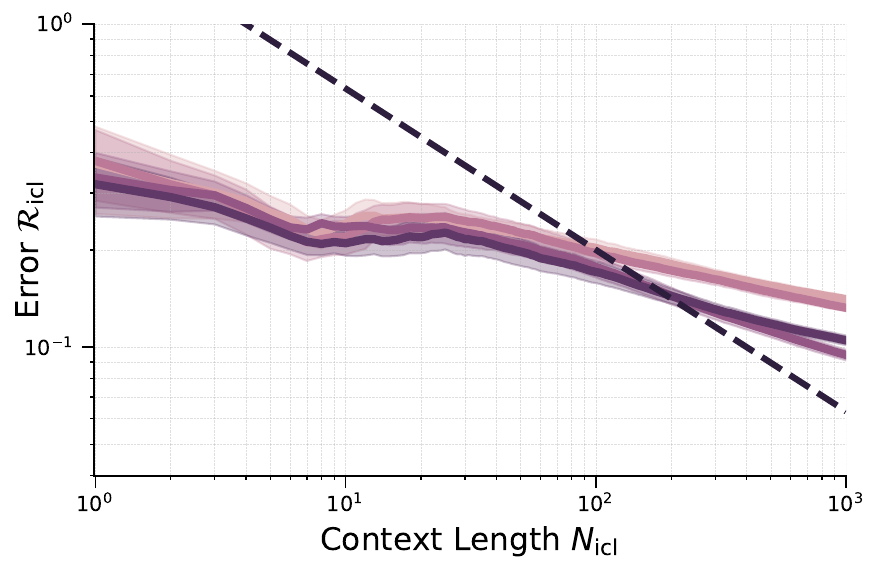}
    \includegraphics[width=0.4\textwidth]{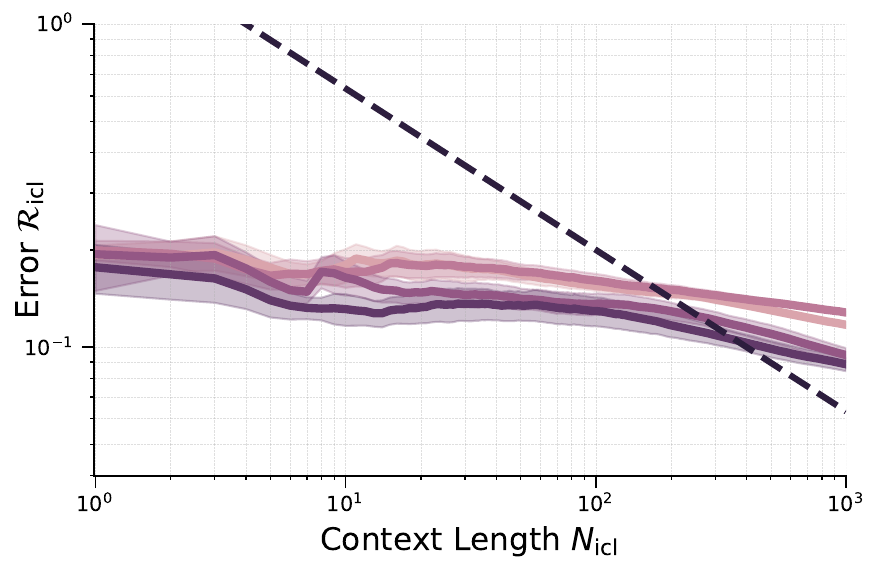}

    \hspace{0.02\textwidth}
    \includegraphics[width=0.9\textwidth]{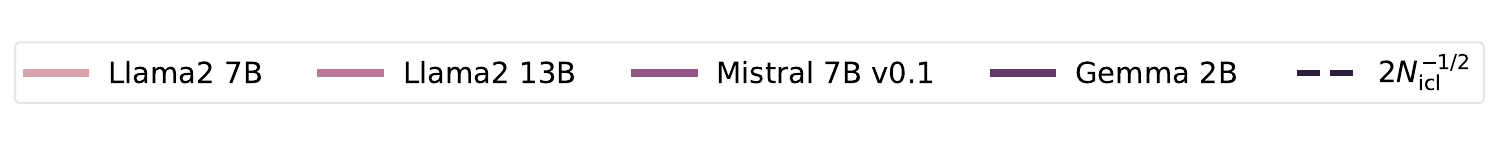}

    \caption{\textbf{Impact of the number of states $\mcsize$.} We plot the risk $\mcal{R}_{\ICL}$ as functions of $N_{\ICL}$, with $95\%$ confidence intervals. \textbf{Upper Left.} $2-$states Markov transition matrices. \textbf{Upper Right.} $4-$states Markov transition matrices. \textbf{Lower Left.} $6-$states Markov transition matrices. \textbf{Lower Right.} $8-$states Markov transition matrices.}
    \label{fig:random_dstates_mc}
\end{figure}

We further analyze the effect of the number of states $\mcsize$ on the risk and consider randomly generated $\mcsize$-state transition matrices in \cref{fig:random_dstates_mc}. After a first stage of stagnation, the risk tends to take the correct scaling law coefficient. As in \citep{liu2024llmslearngoverningprinciples}, we notice that considering randomly generated transition matrices seems to be difficult for an LLM to learn when there are more than $9$ states. We interpret this behavior as the distribution shift term in \cref{thm:risk_bound_llm}. Indeed, the lack of structure in these transition matrices can hinder the correct decay of this term. Note also that the increase in $\mcsize$ tends to implicitly increase $t_{\mrm{min}}$, which could have an impact on the upper bound on $\mcal{R}_{\ICL}$ (both in the generalization term and in the distribution shift term). We will now consider more structured Markov chains, and look at their impact on decay.

\subsection{More Structured Markov Chains}
\label{app:exp_gen_mc}
In this section, we empirically verify our theoretical results on more general Markov chains that do not verify $\pmin > 0$.

\subsubsection{Random Walks}
Random walks are a simple example of more structured Markov chains. Although we still have the possibility of discretizing the kernel of 
Markov chains with infinite state spaces as it is done in \citep{liu2024llmslearngoverningprinciples}, we consider two types of random walks on finite state spaces.

\begin{figure}[htbp]
    \centering
    \centering
    \includegraphics[width=0.6\textwidth]{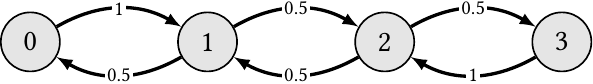}
    \caption{Constrained random walk with $\mcsize = 3$.}
    \label{fig:rw_line}
\end{figure}

\begin{figure}[htbp]
    \centering
    \includegraphics[width=0.4\textwidth]{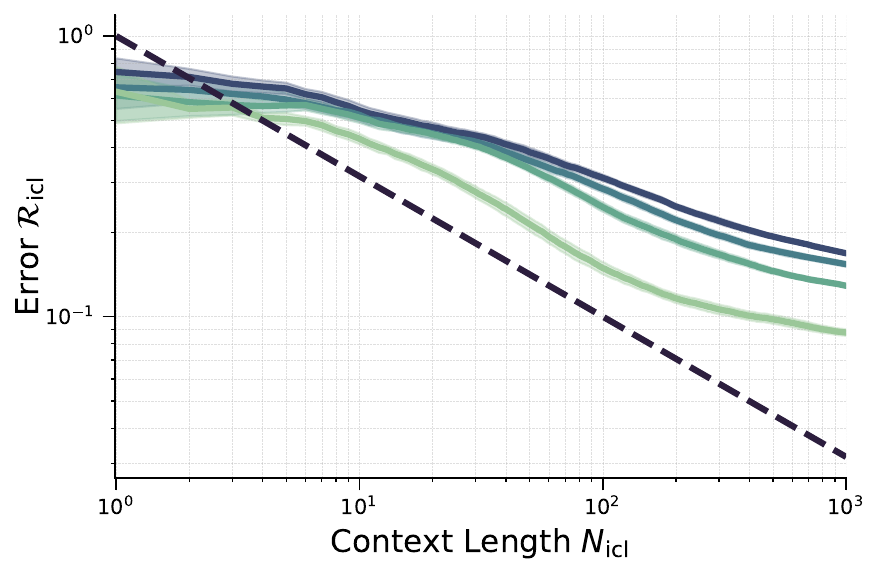}
    \includegraphics[width=0.4\textwidth]{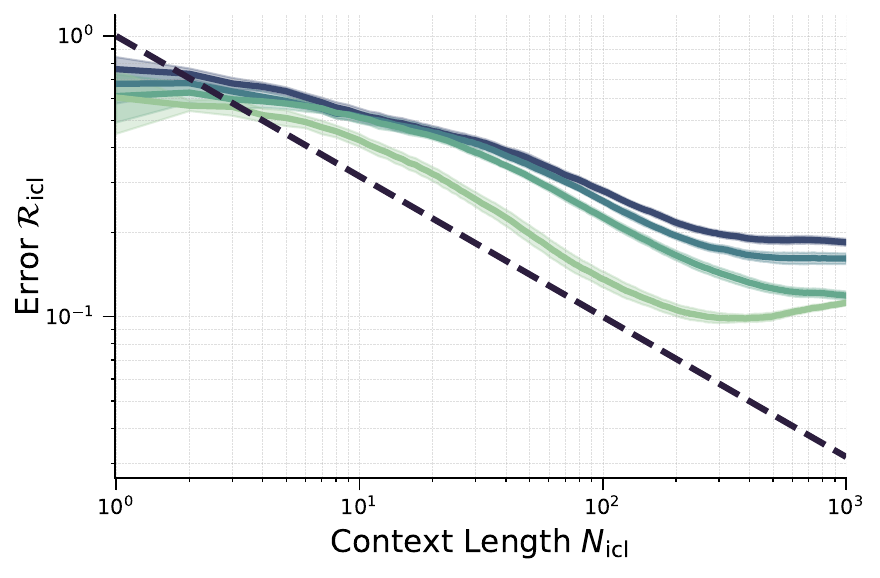}
    \includegraphics[width=0.4\textwidth]{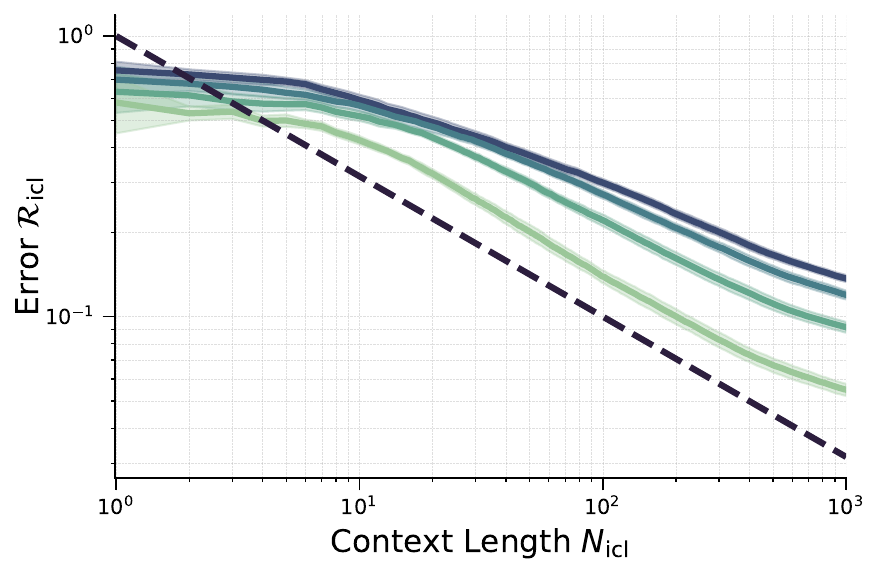}
    \includegraphics[width=0.4\textwidth]{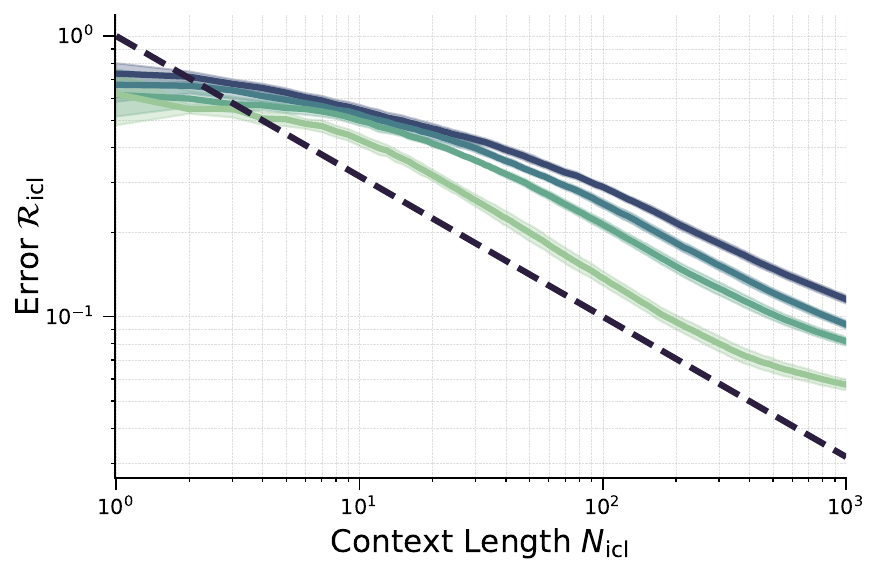}
    \hspace{0.05\textwidth}
    \includegraphics[width=0.8\textwidth]{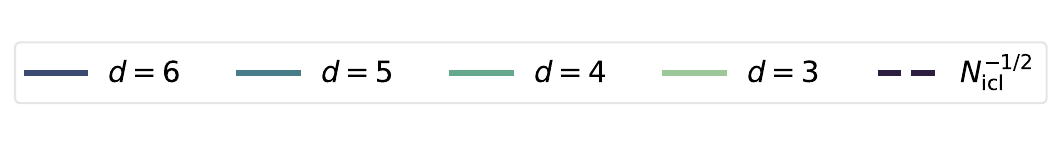}
    \caption{\textbf{Constrained random walks.} We plot the risk $\mcal{R}_{\ICL}$ as functions of $N_{\ICL}$, with $99\%$ confidence intervals. We consider different size $\mcsize$. \textbf{Upper Left.} \texttt{Llama2 7B} \textbf{Upper Right.} \texttt{Llama2 13B} \textbf{Lower Left.} \texttt{Mistral 7Bv0.1} \textbf{Lower Right.} \texttt{Gemma 2B}}
    \label{fig:rw_llama7_mistral7}
\end{figure}

\paragraph{Constrained random walk.} We define the transition matrix $P$ of a constrained random walk of $\mcsize$ states as in \cref{eq:cons_rw}. We draw the probabilistic graph in \cref{fig:rw_line} for the case $\mcsize = 3$.
\begin{equation}
P_{ij} =\begin{cases} 
1, & \text{if } i = 0 \text{ and } j = 1, \\
1, & \text{if } i = \text{\mcsize} - 1 \text{ and } j = \text{\mcsize} - 2, \\
0.5, & \text{if } 1 \leq i \leq \text{\mcsize} - 2 \text{ and } j = i - 1, \\
0.5, & \text{if } 1 \leq i \leq \text{\mcsize} - 2 \text{ and } j = i + 1, \\
0, & \text{otherwise}.\end{cases}
\label{eq:cons_rw}
\end{equation}
\cref{fig:rw_llama7_mistral7} highlights the scaling laws of \cref{thm:risk_bound_llm}, as well as the $\log(\mcsize)$ dependency. As before, the best-performing models generalize almost perfectly.

\begin{figure}[htbp]
    \centering
    \includegraphics[width=0.4\textwidth]{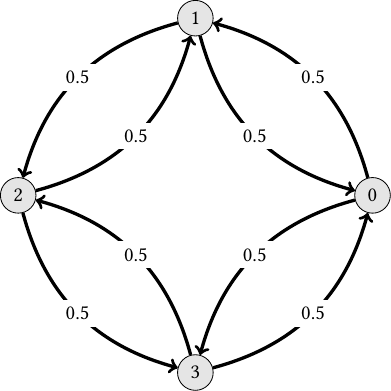}
    \caption{Polygonal random walk with $\mcsize = 4$.}
    \label{fig:rw_poly}
\end{figure}

\begin{figure}[htbp]
    \centering
    \includegraphics[width=0.4\textwidth]{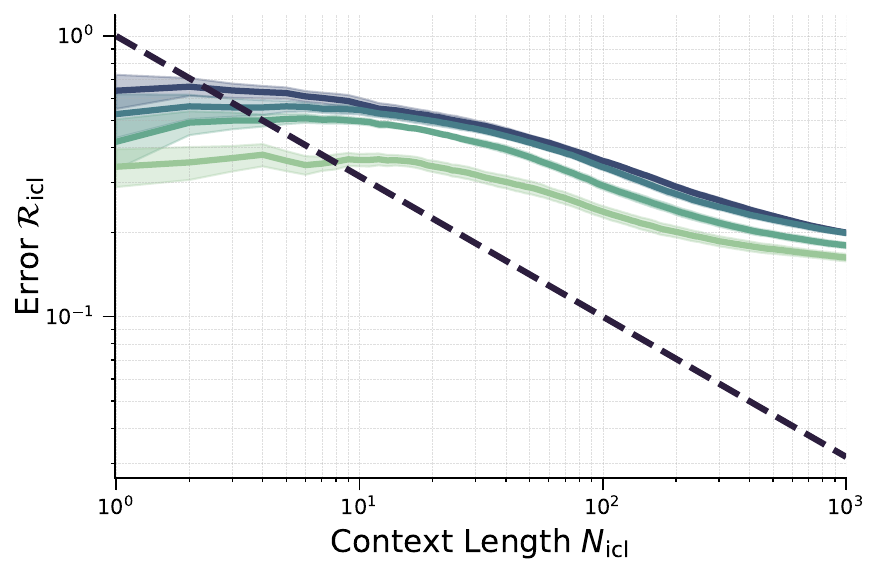}
    \includegraphics[width=0.4\textwidth]{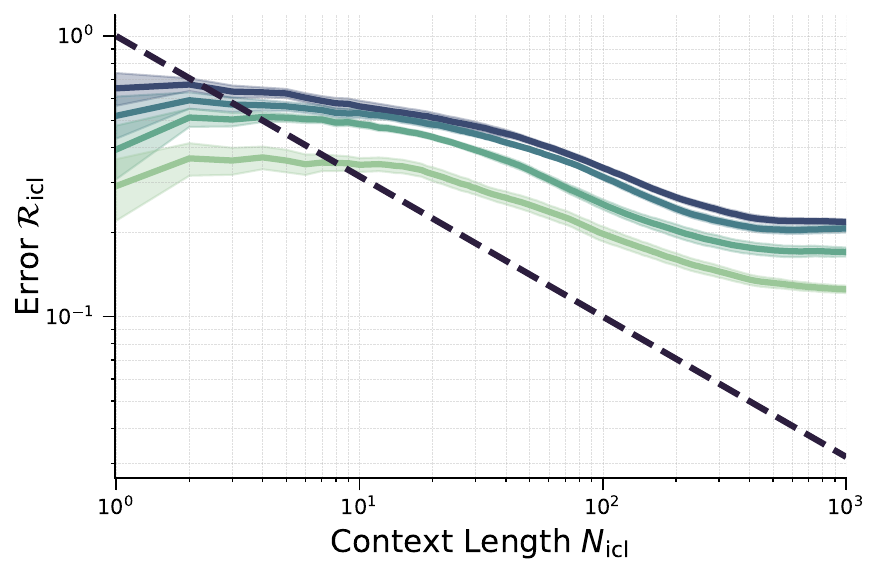}
    \includegraphics[width=0.4\textwidth]{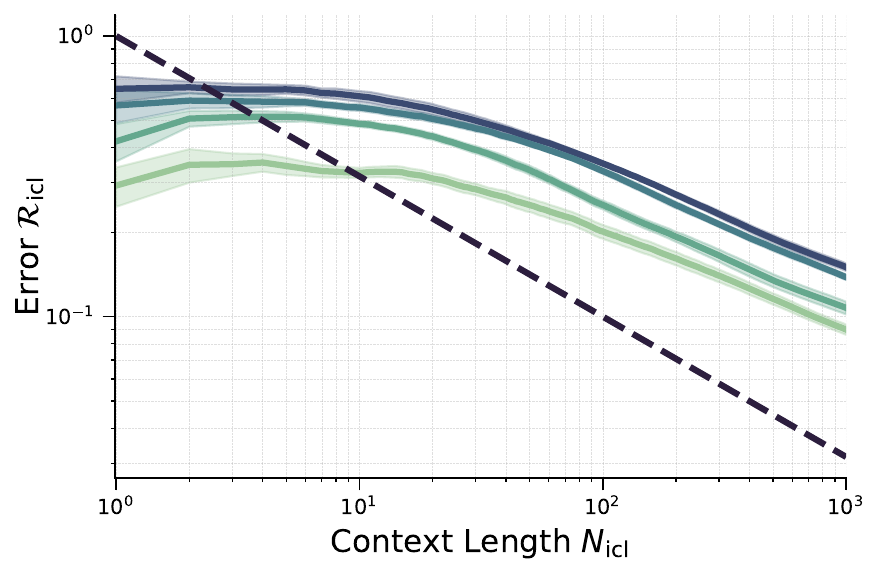}
    \includegraphics[width=0.4\textwidth]{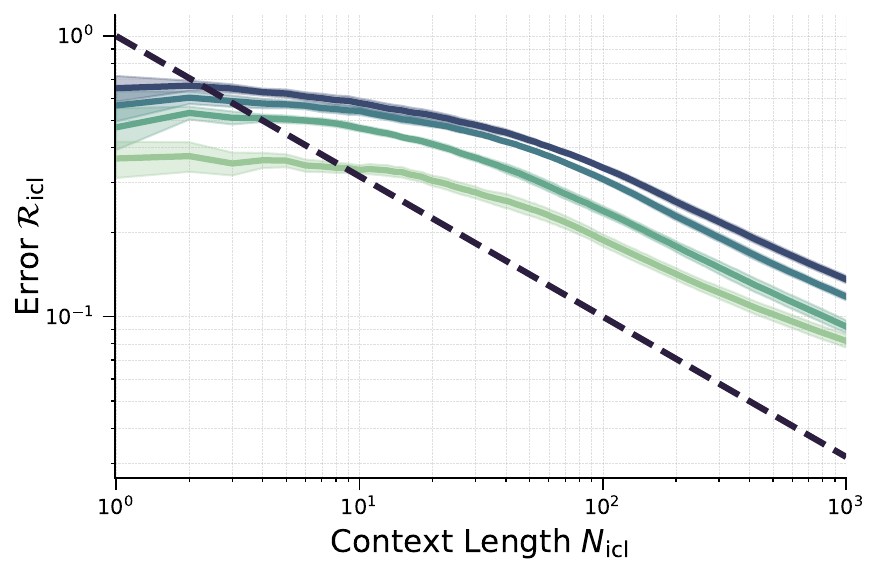}
    \hspace{0.05\textwidth}
    \includegraphics[width=0.8\textwidth]{appendix_experiments/regular_mc/rw_legend_horizontal.pdf}
    \caption{\textbf{Polygonal random walks.} We plot the risk $\mcal{R}_{\ICL}$ as functions of $N_{\ICL}$, with $99\%$ confidence intervals. We consider different size $\mcsize$. \textbf{Upper Left.} \texttt{Llama2 7B} \textbf{Upper Right.} \texttt{Llama2 13B} \textbf{Lower Left.} \texttt{Mistral 7Bv0.1} \textbf{Lower Right.} \texttt{Gemma 2B}}
    \label{fig:rw_poly_llama7_mistral7}
\end{figure}

\paragraph{Polygonal random walk.} We define the transition matrix $\PP$ of a polygonal random walk of $\mcsize$ states as in \cref{eq:poly_rw}. We draw the probabilistic graph in \cref{fig:rw_poly} for the case $\mcsize = 4$.
\begin{equation}
P_{ij} =\begin{cases} 
0.5, & \text{if } j = (i + 1) \mod \text{\mcsize} \text{ (clockwise transition)}, \\
0.5, & \text{if } j = (i - 1) \mod \text{\mcsize} \text{ (counterclockwise transition)}, \\
0, & \text{otherwise}.
\end{cases}
\label{eq:poly_rw}
\end{equation}
We draw the same conclusions as above for this second type of random walk, in \cref{fig:rw_poly_llama7_mistral7}.

\subsubsection{Inner Cliques and Outer Rims}

\begin{figure}[htbp]
    \centering
    \includegraphics[width=0.5\textwidth]{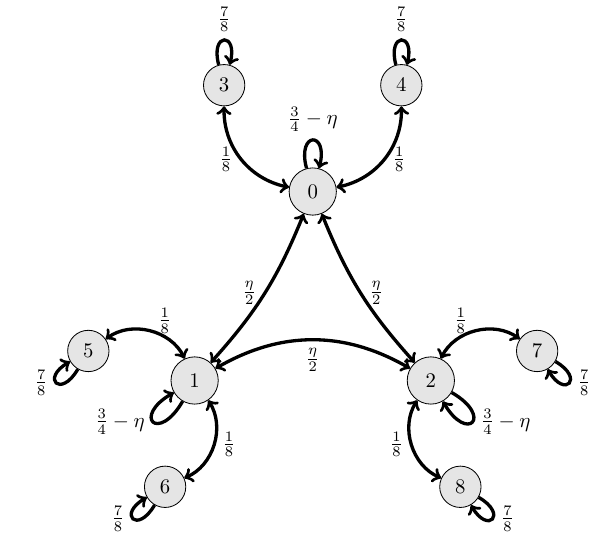}
    \caption{Probabilistic graph of $\mc_{\eta,\bold{0}}$ when $d =9$.}
    \label{fig:ic}
\end{figure}

\paragraph{Inner Cliques and Outer Rims.} We also want to test our method on the class of Markov chain put forward in \citep{wolfer2019minimax} to derive their lower bound. Let $\eta > 0$ and $d=3k$ for some $k \in \NN$, and define the collection of Markov matrices $\mcal{H}_\eta = \{\mc_{\eta,\btau}: \btau \in \{0,1\}^{d/3}\}$. Every element of this set consists of an \textit{inner clique}
and an \textit{outer rim}. $\mc_{\eta,\btau}$ is the block matrix defined as follows,
\begin{equation*}
\mc_{\eta,\btau}= 
\begin{pmatrix}
C_\eta & R_{\btau} \\
R_{\btau}\trn & L_{\btau}
\end{pmatrix},
\end{equation*}
where
$C_\eta\in\RR^{d/3\times d/3}$,
$  L_{\btau}  \in\RR^{2d/3\times2d/3}$,
and
$R_{\btau} \in\RR^{d/3\times 2d/3}$
are given by
$$
  L_{\btau} = \frac{1}{8} \diag\left( 7 - 4 \tau_1 \eps, 7 + 4 \tau_1 \eps, \dots, 7 - 4 \tau_{d/3} \eps, 7 + 4 \tau_{d/3} \eps  \right)
,$$
\begin{equation*}
C_\eta=
\begin{pmatrix}
\frac{3}{4} - \eta & \frac{\eta}{d/3 - 1} & \hdots & \frac{\eta}{d/3 - 1} \\
\frac{\eta}{d/3 - 1} & \frac{3}{4} - \eta & \ddots & \vdots \\
\vdots & \ddots & \ddots & \frac{\eta}{d/3 - 1} \\
\frac{\eta}{d/3 - 1} & \hdots & \frac{\eta}{d/3 - 1} & \frac{3}{4} - \eta \\
\end{pmatrix}
,
\end{equation*}
\begin{equation*}
R_{\btau} = \frac{1}{8}
\begin{pmatrix}
1 + 4 \tau_1 \eps & 1 - 4 \tau_1 \eps  & 0 & \hdots & \hdots & \hdots & 0 \\
0 & 0 & 1 + 4 \tau_2 \eps & 1- 4 \tau_2 \eps  & 0 & \hdots & 0 \\
\vdots & \vdots & \vdots & \vdots & \vdots & \vdots & \vdots \\
0 & \hdots & \hdots & \hdots & 0 & 1 + 4 \tau_{d/3} \eps & 1- 4 \tau_{d/3} \eps 
\end{pmatrix}.
\end{equation*}
We provide in \cref{fig:ic} a probabilistic graph of the case $\mc_{\eta,\bold{0}}$ and $d =9$.

\begin{figure}[htbp]
    \centering
    \begin{overpic}[width=0.4\textwidth]{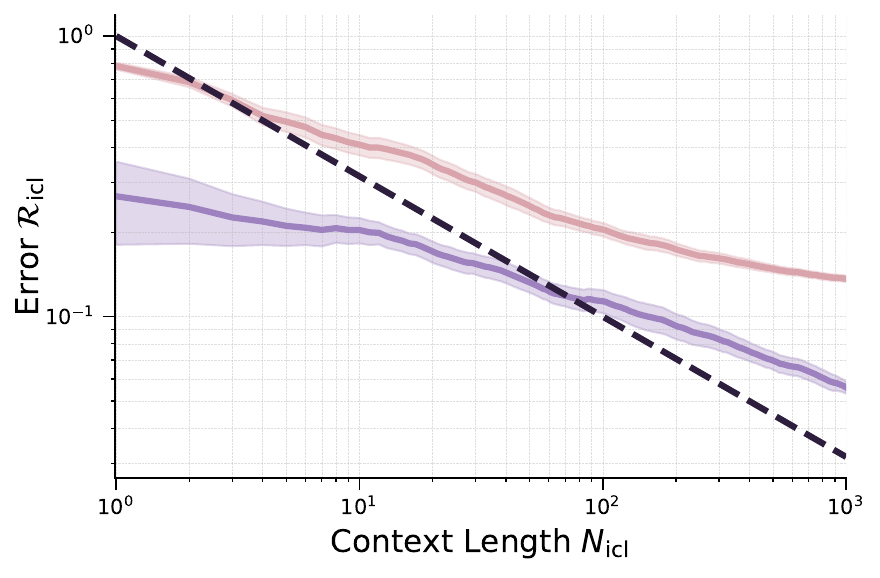}
    \end{overpic}
    \begin{overpic}[width=0.4\textwidth]{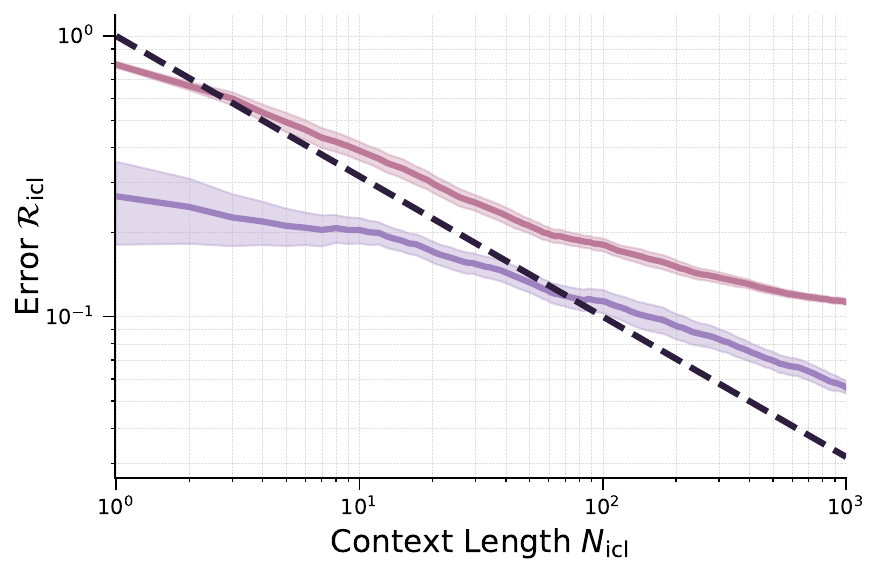}
    \end{overpic}
    \begin{overpic}[width=0.4\textwidth]{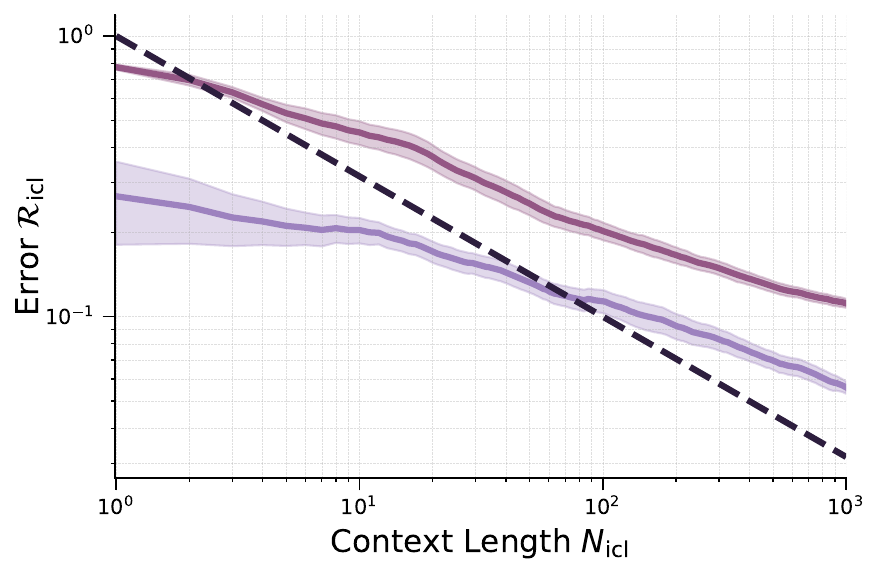}
    \end{overpic}
    \begin{overpic}[width=0.4\textwidth]{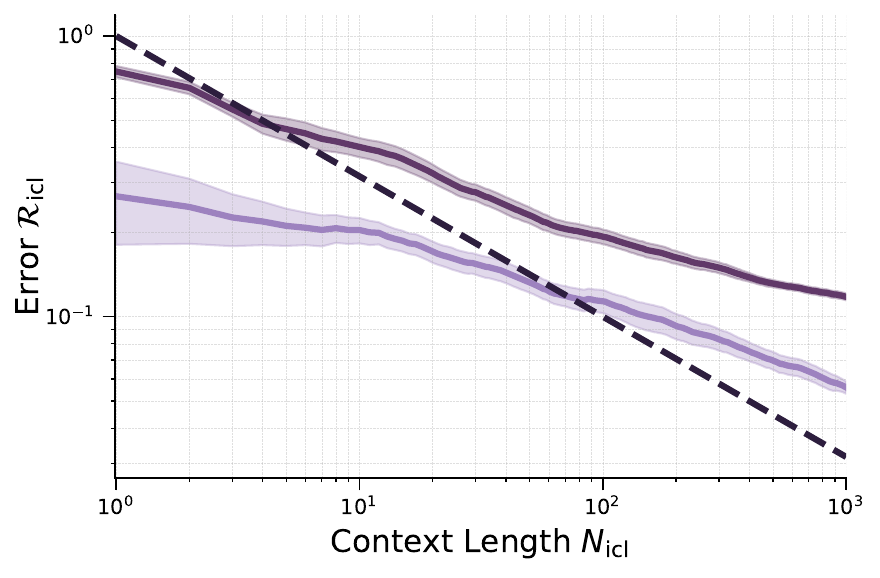}
\end{overpic}
    \hspace{0.02\textwidth}
    \includegraphics[width=0.9\textwidth]{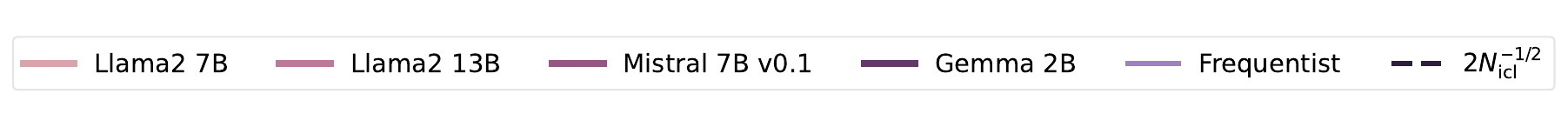}
    \caption{We plot the risk $\mcal{R}_{\ICL}$ as functions of $N_{\ICL}$, with $95\%$ confidence intervals. \textbf{Upper Left.} \texttt{Llama2 7B} \textbf{Upper Right.} \texttt{Llama2 13B} \textbf{Lower Left.} \texttt{Mistral 7Bv0.1} \textbf{Lower Right.} \texttt{Gemma 2B}}
    \label{fig:ic_llms}
\end{figure}

\cref{fig:ic_llms} compares different LLMs with the frequentist method, on the case depicted in \cref{fig:ic_llms} with $\eta = 0.02$. Although the frequentist method achieves a lower loss, the power laws seem to be the same with LLMs.

\subsection{Recent Models: Impact of the Tokenization}\label{app:2024models}

\begin{figure}[htbp]
    \centering
    \begin{overpic}[width=0.4\textwidth]{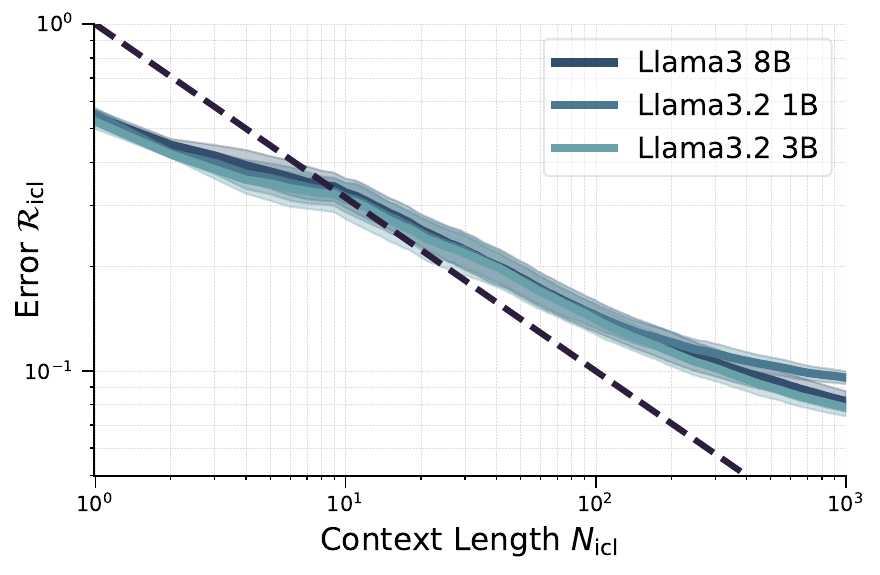}
    \put(26,57){\rotatebox{-33}{\small $\mcal{O}(N_{\ICL}^{-1/2})$}}
    \end{overpic}
    \caption{\textbf{In-context scaling laws for \texttt{LLama3} herd of models.} We plot the risk $\mcal{R}_{\ICL}$ as functions of $N_{\ICL}$, with $95\%$ confidence intervals.}
    \label{fig:icl_scaling_laws_2024models}
\end{figure}

As explained in \cref{app:tokenization}, models like Llama 3 tokenize 3-digit numbers with a single token. This saves a lot of inference compute time, but not necessarily in terms of performance when considering Markov chains with a few number of states $\mcsize$, since we have to separate the states by a comma to force tokenization into a single digit (e.g. the transitions $\texttt{1}\to\texttt{0}\to\texttt{1}$ will be prompted as $\texttt{1,0,1}$ ($5$ tokens) instead of $\texttt{101}$ ($1$ token). In \cref{fig:icl_scaling_laws_2024models}, we reproduce the same experiment as in \cref{fig:icl_scaling_laws}(left), but with Llama 3 models. The scaling laws are quite good, but much less so than those obtained with \texttt{Gemma 2B} and \texttt{Mistral 7Bv0.1} on the same inputs. On the other hand, with these models, it can be extremely interesting to consider Markov chains with many states, as we did in \cref{fig:LLM_vs_freq}(right). In the next section, we will use \texttt{LLama3} to learn other dynamic systems presented in \cite{liu2024llmslearngoverningprinciples}.

\subsection{Dynamical Systems}
\label{app:dynamical_systems}

\begin{figure}[htbp]
    \centering
    \begin{overpic}[width=0.4\textwidth]{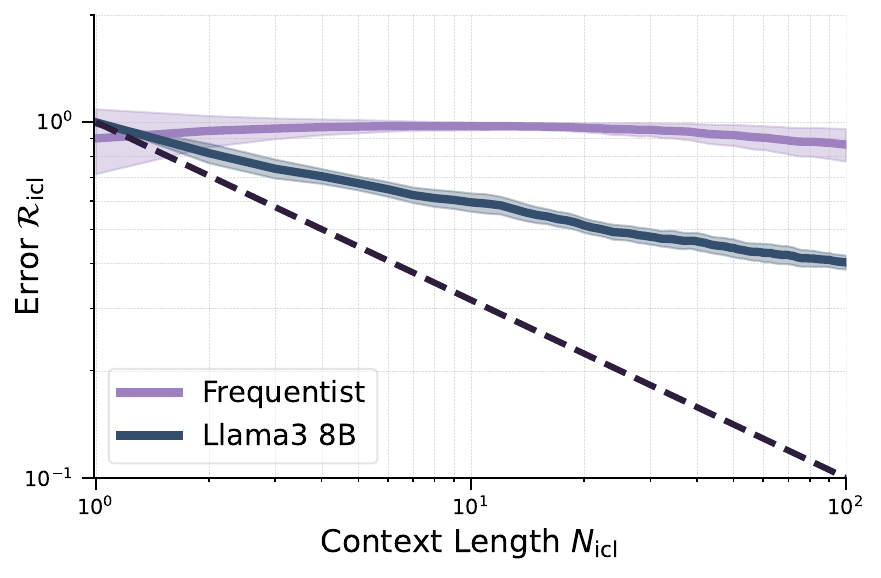}
    \put(68,29){\rotatebox{-28}{\small $\mcal{O}(N_{\ICL}^{-1/2})$}}
    \end{overpic}
    \begin{overpic}[width=0.4\textwidth]{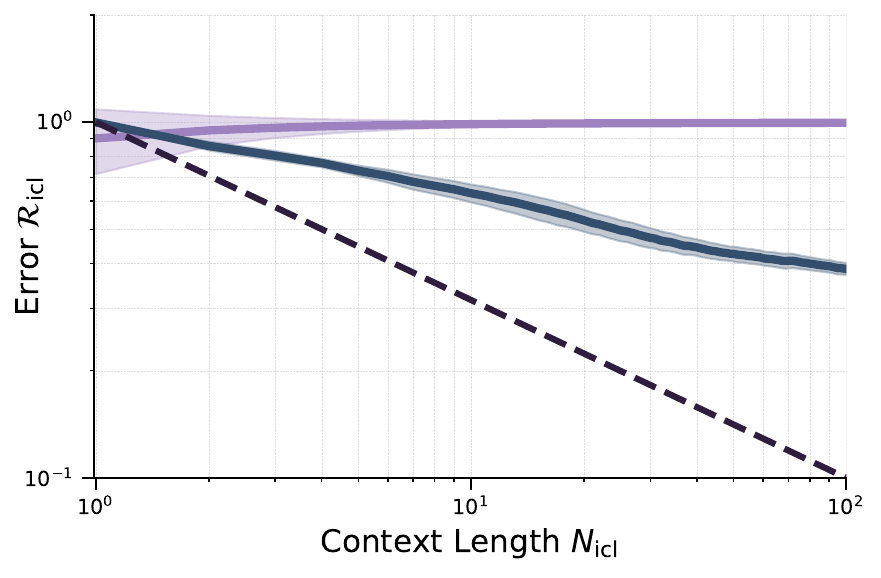}
    \put(68,29){\rotatebox{-28}{\small $\mcal{O}(N_{\ICL}^{-1/2})$}}
    \end{overpic}
    \begin{overpic}[width=0.4\textwidth]{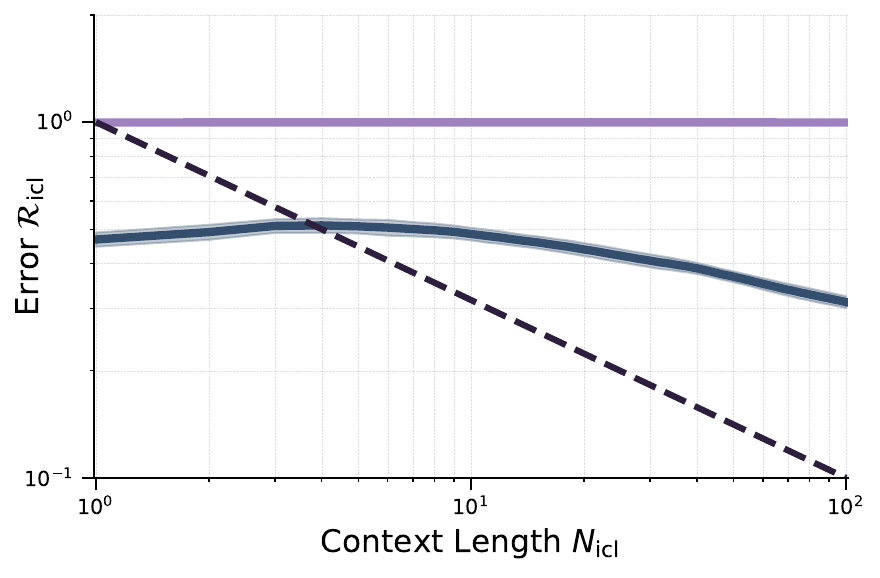}
    \put(68,29){\rotatebox{-28}{\small $\mcal{O}(N_{\ICL}^{-1/2})$}}
    \end{overpic}
    \begin{overpic}[width=0.4\textwidth]{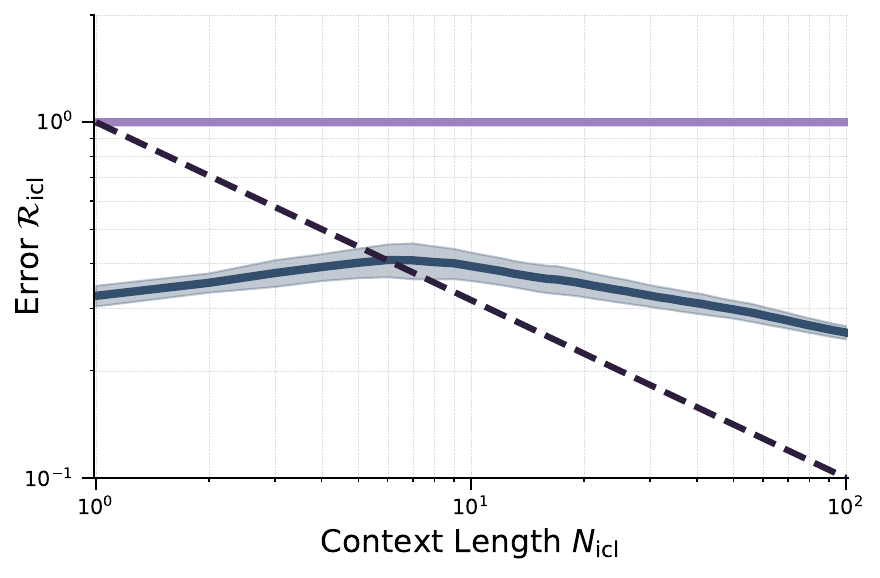}
    \put(68,29){\rotatebox{-28}{\small $\mcal{O}(N_{\ICL}^{-1/2})$}}
    \end{overpic}
    \caption{\textbf{\texttt{LLama3 8B} on dynamical systems.} We plot the risks $\mcal{R}_{\ICL}$ as functions of $N_{\ICL}$ for \texttt{LLama3 8B} and the frequentist approach~\citep{wolfer2019minimax} with $95\%$ confidence intervals. \textbf{Upper Left.} Geometric Brownian motion. \textbf{Upper Right.} Correlated Gaussian. \textbf{Lower Left.} Uncorrelated Gaussian. \textbf{Lower Right.} Uncorrelated Uniform.}
    \label{fig:dynaical systems}
\end{figure}

We consider four of the dynamic systems highlighted in \citep{liu2024llmslearngoverningprinciples} : a geometric Brownian motion, a correlated Gaussian, an uncorrelated Gaussian, and an uncorrelated uniform processe. We display in \cref{fig:dynaical systems} the risks of \texttt{LLama3 8B} and the frequentist method, which once again highlights the emerging capacity of in-context learning.

\section{Additional Theoretical Results}
\label{app:add_thm}
In this section, we present additional theoretical results on the sample complexity and generalization capabilities of LLMs. 

\subsection{Sample Complexity}
In this section, we provide more details to \cref{sec:sample_complexity}, including the experimental setting, the connections to the Markov chain literature, and the extension of \cref{cor:sample_complexity} to the in-context learning setting. 

\subsubsection{Pre-Training}
\label{app:sample_complexity_exp}
We recall that \cref{cor:sample_complexity} states 
\begin{boxprop}[Restatement of \cref{cor:sample_complexity}]
\label{cor:restate_sample_complexity}
   Let $\delta \in [0, 1]$ and $\epsilon > 0$. Assuming a perfect pre-training of $\model_{\params}$ and $N_{\train} $ pre-training tokens with $N_{\train} \geq N^* \coloneqq \lceil\frac{4\bar{B}^2}{\epsilon^2}\log{\mleft( \frac{2}{\delta}\mright)}\rceil$, we have with probability at least $1-\delta$,
    \begin{equation*} 
        \EE_{\mbf{S} \sim \PP_{\mcal{L}}}\lVert \langmat\mleft(\mbf{S}, \cdot \mright) - \transmat\mleft(\mbf{S}, \cdot \mright)\rVert_1
        \leq \epsilon.
    \end{equation*}
    with a constant depending on the problem's parameters
    \[\bar{B}=2\lVert \mixmat \rVert\max\{\log{(\vocabsize)} + 2B_U/\tau, \log{(1/c_0)}\}^{1/2}.\]
\end{boxprop}
To determine the approximation error of open-source LLM, it suffices to plug $N^* = N_{\train}$ in \cref{cor:sample_complexity}. Then, the approximation error writes
\[
\epsilon = \frac{2\bar{B}}{\sqrt{N_{\train}}} \sqrt{\log{\mleft( \frac{2}{\delta}\mright)}}.
\]
We recall that $\bar{B} = 2\lVert \mixmat \rVert \sqrt{\max\{\log{(\vocabsize)} + 2B_U/\tau, \log{(1/c_0)}\}}$. To numerically compute $\epsilon$, we proceed to the following simplifications. Since we do not have access to the training data of the Gemmas and Llamas, we cannot compute $\mixmat$ that captures the data dependency. Even if we had access to the data, the amount of it prevents us from numerically computing $\mixmat$ for all the pre-training sequences. The same goes for $c_o$. We circumvent these issues using the fact that $\vocabsize$ is very high in practice and compute $\bar{B} = \mcal{O}(2\sqrt{\log{(\vocabsize)} + 2B_U/\tau})$. Finally, we notice that the softmax temperature is of order $1$ in those models (we note that it evolves in $[0.2-1]$ depending on the downstream tasks~\citep{dubey2024llama3}) and that $B_U \sim \vocabsize \sqrt{\embdim}$. This comes from the fact that $B_U$ controls the norm of the unembedding matrix $\mbf{W}_U \in \RR^{\vocabsize \times \embdim}$ (see \cref{app:background_transformer}), i.e.
\[
\lVert \mbf{W}_U^\top \rVert_{2, 1} \leq B_U.
\]
When looking at this norm for the unembedding layer of Llama3 8B, we observe that it behaves like $\vocabsize \sqrt{\embdim}$. This is not surprising since the layer normalizations project the outputs' columns on the unit-ball (notably for the RMSNorm~\citep{zhang2019rmsnorm} used in the Llamas models~\citep{touvron2023llama2openfoundation}) and $\mbf{W}_U$ is initialized following $\mcal{N}(0, 1)$ (in PyTorch~\citep{paszke2019pytorch}). This should lead the entries of $\mbf{W}_U$ to be in $[0,1]$ at the end of training which would imply, using the fact that $\mbf{W}_U^\top \in R^{\embdim \times \vocabsize}$,
\[
\lVert\mbf{W}_U^\top \rVert_{2, 1} = \sum_{j=1}^{\vocabsize} \sqrt{\sum_{i=1}^{\embdim} (\mbf{W}_U^\top)_{ij}^2} = \sum_{j=1}^{\vocabsize} \sqrt{\sum_{i=1}^{\embdim} \underbrace{(\mbf{W}_U)_{ji}^2}_{\leq 1}} \leq \vocabsize \sqrt{\embdim}.
\]
In summary, we can indeed consider $B_U = \vocabsize \sqrt{\embdim}$ and we obtain 
\[
\epsilon = \frac{2\bar{B}}{\sqrt{N_{\train}}} \sqrt{\log{\mleft( \frac{2}{\delta}\mright)}}.
\]
where
\[
\bar{B} = 2\sqrt{\log{(\vocabsize)} + 2\vocabsize \sqrt{\embdim}/\tau}.
\]
In the technical reports of the Gemmas~\citep{gemmateam2024gemmaopenmodelsbased, gemmateam2024gemma2improvingopen} and Llamas~\citep{touvron2023llama, touvron2023llama2openfoundation, dubey2024llama3} models, the vocabulary sizes $\vocabsize$,  the embedding dimension $\embdim$, the number of pre-training tokens are given and the MMLU results are given. We summarize it in \cref{tab:exp_details_sample_complexity}. This enables us to plot the evolution of the MMLU with respect to the predicted approximation error $\epsilon$ in \cref{fig:mmlu_epsilon_separate}.

\begin{table*}[!h]
    \centering
    \scalebox{0.9}{
    \begin{tabular}{lccc}
    \toprule[\thick pt]
         \textbf{Model} & \textbf{Nb. Pre-Training Tokens $N_{\train}$} & \textbf{Vocabulary Size $\vocabsize$} & \textbf{Embedding Dimension $\embdim$}\\
        \midrule[\thick pt]
        Llama 7B~\citep{touvron2023llama} & $10^{12}$ & $32000$ & $4096$\\
        Llama2 7B~\citep{touvron2023llama2openfoundation} & $2 \times 10^{12}$& $32000$& $4096$\\
        Llama3 8B~\citep{dubey2024llama3} & $1.5 \times 10^{13}$& $128000$& $4096$\\
        Llama3.2 3B~\citep{dubey2024llama3} & $1.5 \times 10^{13}$& $128000$& $3072$\\
        Gemma 2B~\citep{gemmateam2024gemmaopenmodelsbased} & $3 \times 10^{12}$& $256128$& $2048$\\
        Gemma 7B~\citep{gemmateam2024gemmaopenmodelsbased} & $6 \times 10^{12}$& $256128$& $3072$\\
        Gemma2 9B~\citep{gemmateam2024gemma2improvingopen} & $8 \times 10^{12}$& $256128$ & $3584$\\
        Gemma2 27B~\citep{gemmateam2024gemma2improvingopen} & $1.3 \times 10^{13}$ & $256128$& $4608$\\
        \bottomrule[\thick pt]
    \end{tabular}
    }
    \caption{LLMs' parameters reported in the technical reports of the Llamas and Gemmas models.}
    \label{tab:exp_details_sample_complexity}
\end{table*}

\subsubsection{Connection of \cref{cor:sample_complexity} to Markov Chain Literature}
\label{app:sample_complexity_mc}
\cref{cor:sample_complexity} allows us to contextualize LLMs' ability to learn Markov chains with respect to the existing literature. To the best of our knowledge, the only existing approach with theoretical guarantees for learning Markov chains is the frequentist method~\citep{wolfer2019minimax}: counting the number of occurrences of different states to fill in the matrix $\transmat$. ~\citet{wolfer2019minimax} show that the sample complexity of approximating $\langmat$ up to $\epsilon$ with such approach is $N^* = \mcal{O}(\max\{|\mcal{V}_\cxtsize^*|/\epsilon^2 \gamma_s, 1/\gamma_s \pi^*\} \log{\mleft( \frac{1}{\delta}\mright)})$ samples, where $|\mcal{V}_\cxtsize^*| = \mcal{O}(\vocabsize^\cxtsize)$ in our setting, $\gamma_s$ is a (pseudo) spectral gap of the Markov chain and $\pi^*$ is the smallest element of its stationary distribution. The authors state that the frequentist approach is minimax optimal (up to logarithmic factors). Our sample complexity in \cref{cor:restate_sample_complexity} has a dependence that behaves as $\smash{\bar{B}^2 = \mcal{O}(\max\{\log{(\vocabsize)} + \frac{2\vocabsize\sqrt{\embdim}}{\tau}, \log{\mleft(1/{c_0}\mright)}\})}$, using $B_U \sim \vocabsize \sqrt{\embdim}$. Given that in practice $\vocabsize>\embdim$, it then simplifies to $N ^* = \mcal{O}(\max\{\vocabsize/\epsilon^2\tau, 1/\epsilon^2\}\log{\mleft( \frac{1}{\delta}\mright)})$. In particular, our LLMs' sample complexity is linear in the vocabulary size $\vocabsize$, which is remarkable compared to the sample complexity of the frequentist approach, which scales as $\mcal{O}(\vocabsize^\cxtsize)$ with $\cxtsize$ the context window. 

\paragraph{Extension to In-Context Learning.}
Using \cref{thm:risk_bound_llm}, we can extend \cref{cor:sample_complexity} to the in-context learning framework. The following proposition gives the sample complexity of in-context learning.
\begin{boxprop}[Sample complexity of in-context learning]
\label{cor:sample_complexity_icl}
   Let $\delta \in [0, 1]$ and $\epsilon > 0$. The model $\model_{\params}$ receives as inputs a $\mcsize-$state Markov chain $X = \MCseq{X}{N_{\ICL}}$ with transition matrix $\mbf{Q}$. Assuming that there is no distribution shift and $N_{\ICL} \geq N^* \coloneqq \lceil\frac{4\bar{B}^2}{\epsilon^2}\log{\mleft( \frac{2}{\delta}\mright)}\rceil$, we have with probability at least $1-\delta$,
    \begin{equation*} 
        \EE_{\mbf{S} \sim \PP_{\mcal{L}}}\lVert \mbf{Q}\mleft(\mbf{S}, \cdot \mright) - \transmat\mleft(\mbf{S}, \cdot \mright)\rVert_1
        \leq \epsilon.
    \end{equation*}
    with a constant depending on the problem's parameters 
    \[\bar{B}= 2\max\{\log{(\mcsize)} + 2B_U/\tau, \log{(1/\pmin)}\}^{1/2}.
    \]
\end{boxprop}
\begin{proof}
We recall that the probability distribution associated with the input Markov chain is $\PP$ and its transition matrix writes $\QQ$ (see \cref{app:background_mc} for the connection between $\PP$ and $\QQ$). We first note that by definition of the total variation distance~\citep{wolfer2019minimax}, we have
\begin{align*}
    \EE_{\mbf{S} \sim \PP}\lVert \mbf{Q}\mleft(\mbf{S}, \cdot \mright) - \transmat\mleft(\mbf{S}, \cdot \mright)\rVert_1 &= \EE_{\mbf{S} \sim \PP} \mleft[2 \cdot \TV{\QQ\mleft(\mbf{S}, \cdot \mright)}{\transmat\mleft(\mbf{S}, \cdot \mright)} \mright]\\
    &= 2\cdot\EE_{\mbf{S} \sim \PP} \mleft [ \TV{\QQ\mleft(\mbf{S}, \cdot \mright)}{\transmat\mleft(\mbf{S}, \cdot \mright)} \mright]\\
    &= 2\cdot\riskicl{\params} \tag{by definition of the risk \cref{eq:estimation_error_def}}.
\end{align*}
Applying \cref{thm:risk_bound_llm}, we know that
\begin{equation*}
    \riskicl{\params} \leq \inf_{\mcparams \in \mcparamspace} \{ \smallriskicl{\mcparams} + \mcal{K}(\mcparams, \params)\} +  \frac{\bar{B}}{\sqrt{N_{\ICL}}}\sqrt{\log{\mleft(\frac{2}{\delta}\mright)}},
\end{equation*}
where $\bar{B}$ is formally defined in \cref{thm:risk_bound_llm}. Assuming no distribution shift amounts to considering the infimum attained and equal to $0$. We denote by $N^*$ the integer such that the error is equal to $\frac{\epsilon}{2}$, i.e.,
\begin{equation*}
    \frac{\bar{B}}{\sqrt{N^*}}\sqrt{\log{\mleft(\frac{2}{\delta}\mright)}} = \frac{\epsilon}{2} \iff \frac{\bar{B}^2}{N^*}\log{\mleft(\frac{2}{\delta}\mright)} = \frac{\epsilon^2}{4} 
    \iff N^* = \mleft(\frac{2\bar{B}}{\epsilon}\mright)^2\log{\mleft(\frac{2}{\delta}\mright)}.
\end{equation*}
Taking the ceiling function ensures that $N^*$ is an integer. Hence, taking $N_{\train} \geq N^* = \lceil \mleft(\frac{2\bar{B}}{\epsilon}\mright)^2\log{\mleft(\frac{2}{\delta}\mright)} \rceil$ ensures that
\begin{align*}
    \frac{\bar{B}}{\sqrt{N_{\ICL}}}\sqrt{\log{\mleft(\frac{2}{\delta}\mright)}} \leq \frac{\bar{B}}{\sqrt{N^*}}\sqrt{\log{\mleft(\frac{2}{\delta}\mright)}} = \frac{\epsilon}{2}.
\end{align*}
Putting everything together, taking $N_{\ICL} \geq N^*$ leads to
\begin{equation*}
    \EE_{\mbf{S} \sim \PP}\lVert \QQ\mleft(\mbf{S}, \cdot \mright) - \transmat\mleft(\mbf{S}, \cdot \mright)\rVert_1 \leq 2 \cdot \riskicl{\params} \leq 2 \cdot \frac{\epsilon}{2} = \epsilon,
\end{equation*}
which concludes the proof.
\end{proof}
Similarly to the analysis above, we observe that our sample complexity of in-context learning scales logarithmically with the number of states $\mcsize$ while the frequentist's one~\citep{wolfer2019minimax} scales in $\mcal{O}(\mcsize)$. In \cref{fig:LLM_vs_freq} of \cref{sec:experiment}, we show that this is confirmed experimentally: LLM's ability to learn Markov chains exceeds the frequentist approach for Markov chains with a large state space.

\subsection{Depth-Dependent Generalization Bounds}
\label{app:depth_dependent_bound}
We extend \cref{thm:pre_training_risk_bound_llm} to make its dependency on $\model_{\params}$ more fine-grained. Rather than assuming that only the norm of the embedding layer's matrix is bounded, we follow the setting of prior work~\citep{zhang2023whathowicl, furuya2024transformers, marion2023generalization, edelman2022inductive} and consider the parameter space defined as follows:
\begin{equation*}
\begin{split}
    \widetilde{\paramspace} = &\{ \params \in \paramspace\mid \forall \ell \in [L],\lVert \mbf{W}_V^{(\ell)} \rVert_{\infty} \leq B_V,\\
    &\lVert \mbf{W}_O^{(\ell)} \rVert_{\infty} \leq B_O, \lVert \mbf{W}_1^{(\ell)} \rVert_{\infty} \leq B_1, \lVert \mbf{W}_2^{(\ell)} \rVert_{\infty} \leq B_2\}.
    \end{split}
\end{equation*}
The definition of $\widetilde{\paramspace}$ concerns the query, key, and value matrices of all layers and heads. 
Similarly to~\citet[Assumption 5.1]{zhang2023whathowicl}, we assume that each token has an $\ell_1$-norm bounded by $\ubtok$. We have the following generalization bound, whose proof is deferred to \cref{app:pre_training_risk_bound_llm_depth}.
\begin{boxcor}[Depth-dependent bound]
    \label{cor:pre_training_risk_bound_llm_depth}
    Consider an LLM $\model_{\params} \in \tilde{\funcspace} := \{ \model_{\params} \mid \params \in \tilde{\paramspace} \}$. With the same assumptions as in \cref{thm:pre_training_risk_bound_llm}, we have
    \begin{equation*}
        \risktrain{\params} \leq \smallrisktrain{\params} +  \frac{\bar{B}}{\sqrt{N_{\train}}}\sqrt{\log{\mleft(\frac{2}{\delta}\mright)}},
    \end{equation*}
    with constants depending on the problem's parameters 
    \begin{align*}
    \bar{B} &= 2\lVert \mixmat \rVert\max\{\log{(\vocabsize)} +2(\ubmod)^L/\tau, \log{(1/c_0)}\}^{1/2},\\
    \ubmod &= [( 1+ \embdim \hiddendim B_1B_2) ( 1 + \frac{\embdim^3}{H} B_OB_V )]( \ubtok B_U)^{1/L}.
    \end{align*}
\end{boxcor}
We note that $\bar{B}$ exhibits an exponential dependence on the depth of the transformer, which also amplifies the hidden dimensionality (width) of the embedding layer $r$. This contrasts with the dependency in $m$, the hidden dimensionality of the MLP block, which is linear. All these factors are commonly associated with higher expressive power of transformers suggesting that they should contribute to a better minimization of $\smash{\smallrisktrain{\params}}$ at the expense of requiring more training data. The number of heads $H$ can be used as a counterbalance to increasing the width in the cubic term $r^3$, suggesting that a good balance between these parameters may lead to more data-efficient models.

\subsection{Generalization Bounds with the KL Divergence}
\label{app:kl_thm}
As explained in \cref{rmk:choice_risk}, the total variation is the natural choice to define the risks in \cref{eq:estimation_error_def}. Another possibility in the Markov chain literature is to use the KL divergence to compare probability distributions~\citep{hao2018mc}. This is an interesting candidate as the KL divergence is naturally connected to the cross-entropy loss commonly used to train neural networks (the cross-entropy corresponds to the KL divergence between the true distribution and the predicted softmax distribution~\citep{blondel2019fenchel}. In this section, we discuss the extension of the theoretical results of \cref{sec:theoretical_analysis} by replacing the TV distance with the KL divergence in the risks' definition, i.e., 
\begin{equation}
    \label{eq:estimation_error_def_kl_extension}
    \risk{\params} \coloneqq \EE_{\mbf{S} \sim \PP_{\mcal{L}}} \mleft[ \KL{\langmat\mleft(\mbf{S}, \cdot\mright)}{\transmat\mleft(\mbf{S}, \cdot\mright)}\mright],\, \smallrisk{\params} \coloneqq \frac{1}{N}\sum_{n=1}^N \KL{\Pbl{\cdot}{\mbf{S}_n}}{\Pbemp{\params}{\cdot}{\mbf{S}_n}}.
\end{equation}

\cref{thm:pre_training_risk_bound_llm} and \cref{cor:pre_training_risk_bound_llm_depth} related to the pre-training phase in \cref{sec:pretraining} can be obtained similarly if the risks are defined with the KL divergence following \cref{eq:estimation_error_def_kl_extension}. Indeed, the key step to derive the proofs is to obtain a similar result to \cref{lem:ineq_tv_hellinger} but with the KL divergence. The next lemma provides this result.
\begin{boxlem}
    \label{lem:ineq_tv_hellinger_kl}
    Consider two probability distributions $\PP, \QQ$ defined on a measure space $\measurable$ and a $\sigma$-finite measure $\nu$ on $\measurable$. Let $p, q$ be the corresponding probabilities densities, i.e., we have $\PP(d\omega) = q(\omega)\nu(d\omega)$ and $\QQ(d\omega) = p(\omega)\nu(d\omega)$. If there exists a non-negative constant $B$ such that for any $z \in \Omega$, $\left |\log{\sqrt{\frac{\PP(z)}{\QQ(z)}}}\right | \leq B$,
    then we have
    \begin{equation*}
        \KL{\PP}{\QQ} \leq B.
    \end{equation*}
\end{boxlem}
\begin{proof}
We have
\begin{align*}
0 \leq \KL{\PP}{\QQ} &= \lvert \KL{\PP}{\QQ} \rvert \\
&= \left | \int \PP(z) \log(\frac{\PP(z)}{\QQ(z)})dz\right | \\
&\leq \int \lvert \PP(z) \rvert \lvert \log(\frac{\PP(z)}{\QQ(z)}) \rvert dz \\
&\leq B  \int \lvert \PP(z) \rvert dz \\
&= B \int \PP(z) dz \\
&= B,
\end{align*}
which concludes the proof.
\end{proof}

We can now state the results similar to \cref{thm:pre_training_risk_bound_llm}, \cref{cor:pre_training_risk_bound_llm_depth} and \cref{cor:sample_complexity} from the pre-training phase when the risk is defined according to \cref{eq:estimation_error_def_kl_extension}. 

\begin{boxthm}[Pre-training generalization bound]
    \label{thm:pre_training_risk_bound_llm_kl}
    Consider an LLM $\model_{\params} \in \funcspace$. We denote by $\mixmat$ the mixing matrix of the pre-training sequences of tokens $\MCseq{S}{N_{\train}}$. Let $0<\delta<1$, then with probability at least $1-\delta$,
    \begin{equation*}
        \risktrain{\params} \leq \smallrisktrain{\params} +  \frac{\bar{B}}{\sqrt{N_{\train}}}\sqrt{\log{\mleft(\frac{2}{\delta}\mright)}},
    \end{equation*}
    where $\bar{B}=\sqrt{2}\lVert \mixmat \rVert\max\{\log{(\vocabsize)} + 2B_U/\tau, \log{(1/c_0)}\}$ is a constant depending on the parameters of the problem.
\end{boxthm}
\begin{proof}
The proof simply follows from the proof of \cref{thm:pre_training_risk_bound_llm} by replacing the upper bound $\sqrt{2B}$ by $B$ (with the appropriate upper-bound $B$) when \cref{lem:ineq_tv_hellinger} is used in the proof.
\end{proof}

\begin{boxcor}[Depth-dependent bound]
    \label{cor:pre_training_risk_bound_llm_depth_kl}
    Consider an LLM $\model_{\params} \in \tilde{\funcspace} := \{ \model_{\params} \mid \params \in \tilde{\paramspace} \}$. With the same assumptions as in \cref{thm:pre_training_risk_bound_llm}, we have
    \begin{equation*}
        \risktrain{\params} \leq \smallrisktrain{\params} +  \frac{\bar{B}}{\sqrt{N_{\train}}}\sqrt{\log{\mleft(\frac{2}{\delta}\mright)}},
    \end{equation*}
    where $\bar{B} = \sqrt{2}\lVert \mixmat \rVert\max\{\log{(\vocabsize)} +2(\ubmod)^L/\tau, \log{(1/c_0)}\}$ is a constant depending on the parameters of the problem, and $\ubmod = [( 1+ \embdim \hiddendim B_1B_2) ( 1 + \frac{\embdim^3}{H} B_OB_V )]( \ubtok B_U)^{1/L}$.
\end{boxcor}
\begin{proof}
The proof simply follows from the proof of \cref{thm:pre_training_risk_bound_llm} by replacing the upper bound $\sqrt{2B}$ by $B$ (with the appropriate upper-bound $B$) when \cref{lem:ineq_tv_hellinger} is used in the proof.
\end{proof}

\paragraph{Limitations.} We recall from \cref{rmk:choice_risk} that the TV distance is a natural choice to compare transition matrices in the Markov chain literature. In addition, while the KL divergence can be used to compare probability distributions, it does not define a metric space. Hence, we cannot straightforwardly extend \cref{thm:risk_bound_llm} with the KL divergence because the proof relies on the use of the triangular inequality. As \cref{thm:risk_bound_llm} is one of our main results and enables us to show that the theory and the practice align (\cref{sec:experiment}), this also contributed to our preference for the TV distance instead of the KL divergence. We also note that the proof of \cref{cor:sample_complexity} relies on properties of the total variation and hence we cannot extend it straightforwardly with the KL divergence.

\section{Proofs}
\label{app:proofs}

\subsection{Proof of \cref{prop:LLM_formal_def}}\label{app:autoregrssive_markov}

We detail below the proof of \cref{prop:LLM_formal_def}.

\begin{proof}[Proof of \cref{prop:LLM_formal_def}]

\textbf{Step 1: Large language models as Markov chains.} Given an input $v_i \in \mcal{V}_\cxtsize^*$ of $p$ tokens, a large language model outputs a probability mass function $\model_{\params}^{\vocabsize,\cxtsize}(v_i)$ over the discrete vocabulary space. As the temperature is positive, i.e., $\tau > 0$, and as the exponential is positive, we know that all the tokens in the vocabulary will be given a positive mass.

A next sequence $v_j \in \mcal{V}_\cxtsize^*$ is then sampled according to $\model_{\params}^{\vocabsize,\cxtsize}(v_i)$. But the $v_j$ sequences that fit necessarily contain the $v_i$ sequence (except possibly the first element of $v_i$, thanks to \cref{def:deletion_process}), i.e. $\forall l, (v_j)_l = (v_i)_{l+1}$. Note also the size of $v_j$ is $p +1$ when $p<k$ and $k$ when $p=k$. All other sequences $v_j$ that do not satisfy this condition are not suitable.

In that sense, $\model_{\params}^{\vocabsize,\cxtsize}$ can be represented by a Markov chain $\text{MC}(\mcal{V}_\cxtsize^*, \transmat)$ with transition kernel $\transmat \in \mathbb{R}^{|\mcal{V}_\cxtsize^*|\times |\mcal{V}_\cxtsize^*|}$, as defined in \cref{prop:LLM_formal_def}.

\textbf{Step 2: Proportion of non-zero elements.} We denote by $\recurrent$ the set of states of length $\cxtsize$. The set of states of length strictly less than equal $\cxtsize$ is denoted by $\transient$. We can construct a transition matrix $P_\recurrent \in \RR^{\vocabsize^\cxtsize\times\vocabsize^\cxtsize}$ with the states of this class, containing the probabilities of moving from one state of $\recurrent$ to another. $P_\recurrent$ corresponds to the blue block in \cref{fig:exp_matrix} while green rectangle blocks correspond to part of $P_\transient$ and $P_{\transrec}$ in the following description of large language models as Markov chains,
    \begin{equation}\label{eq:matrix_q_blocks}
        \transmat = \begin{pNiceMatrix}[
                columns-width = auto,
                hvlines,
                cell-space-limits = 4pt
            ]
            P_\transient & P_{\transrec} \\
            0            & P_\recurrent  \\
        \end{pNiceMatrix}.\end{equation}

    Now, let us count the number of non-zero elements in each of these $4$ large blocks.

    \underline{$P_\transient$ block :} The size of this block is $\displaystyle \Big[\frac{\vocabsize}{\vocabsize-1}(\vocabsize^{\cxtsize-1} - 1)\Big] \times \Big[\frac{\vocabsize}{\vocabsize-1}(\vocabsize^{\cxtsize-1} - 1)\Big]$. There are $\cxtsize-2$ green blocks contained in $P_{\transient}$. The block number $i \in [\cxtsize-2]$ is of size $\vocabsize^{i} \times \vocabsize^{i+1}$. Since each sentence of size $i$ can be completed with non-zero probability, by any other token, there are a total of $\sum_{p=1}^{\vocabsize^{i}}\vocabsize = \vocabsize^{i+1}$ non-zero elements. There are therefore $\sum_{i=1}^{\cxtsize-2}\vocabsize^{i+1}$ non-zero elements in the entire $P_\transient$ block.

    \underline{$P_{\transrec}$ block :} The size of this block is $\displaystyle \Big[\frac{\vocabsize}{\vocabsize-1}(\vocabsize^{\cxtsize-1} - 1)\Big] \times \vocabsize^\cxtsize$. The green block contained in $P_{\transrec}$ that contains non-zero elements is of size $\vocabsize^{\cxtsize-1} \times \vocabsize^\cxtsize$. Since each sentence of size $\cxtsize-1$ can be completed with non-zero probability, by any other token, there are a total of $\sum_{p=1}^{\vocabsize^{\cxtsize-1}}\vocabsize = \vocabsize^{\cxtsize}$ non-zero elements.

    \underline{$P_\recurrent$ block :} The size of this block is $\vocabsize^\cxtsize \times \vocabsize^\cxtsize$. Each sentence $v = (v_1,\hdots,v_\cxtsize)$ of size $\cxtsize$ is mapped to another sentence $v' = (v_1',\hdots,v_\cxtsize')$ of size $\cxtsize$ with non-zero probability, if and only if $v_1' = v_2, v_2' = v_3, \hdots, v'_{k-1} = v_\cxtsize$. The final token $v'_{\cxtsize}$ can by any other token in the vocabulary. It means that there are a total of $\sum_{p=1}^{\vocabsize^{\cxtsize}}\vocabsize = \vocabsize^{\cxtsize+1}$ non-zero elements.

    \underline{$0's$ block :} There are no non-zero elements in this block.

    Finally, there are 
    \[\sum_{i=1}^{\cxtsize-2}\vocabsize^{i+1} + \vocabsize^{\cxtsize} + \vocabsize^{\cxtsize+1} = \sum_{i=1}^{\cxtsize}\vocabsize^{i+1} = \vocabsize^{2}\left(\frac{\vocabsize^{\cxtsize} - 1}{\vocabsize-1}\right)\]
 non-zero elements. This means that the proportion of non-zero elements in the matrix is exactly 
 \[\frac{\vocabsize^{2}\left(\frac{\vocabsize^{\cxtsize} - 1}{\vocabsize-1}\right)}{ \left(\vocabsize\left(\frac{\vocabsize^{\cxtsize} - 1}{\vocabsize-1}\right)\right)^2} = \frac{\vocabsize - 1}{\vocabsize^{\cxtsize}-1}.\] 
 Note that for large $\vocabsize$ and $\cxtsize$ we have that 
 \[\frac{\vocabsize - 1}{\vocabsize^{\cxtsize}-1}\sim\frac{1}{\vocabsize^{\cxtsize-1}}.\]
\end{proof}

\subsection{Proof of \cref{prop:ergodic_unichains}} \label{app:ergodic_unichains}

We begin with a preliminary lemma.

\begin{boxlem}[Powers of $\transmat$ greater than $\cxtsize$]\label{lem:llm_power_k}
    For any initial state $i$, the following hold:
    \begin{itemize}
        \item $\forall k \geq \cxtsize, \forall j \in \transient, (\transmat^k)_{i,j} = 0$,
        \item $\forall k \geq \cxtsize, \forall j \in \recurrent, (\transmat^k)_{i,j} > 0$.
    \end{itemize}
\end{boxlem}
\begin{proof}
    By considering $\transmat$ as defined in \eqref{eq:matrix_q_blocks}, we can compute its powers. For any $k \geq 1,$
    \begin{equation*}
        \transmat^k = \begin{pNiceMatrix}[
                columns-width = auto,
                hvlines,
                cell-space-limits = 4pt
            ]
            P_\transient^k & B_k            \\
            0              & P_\recurrent^k \\
        \end{pNiceMatrix},\end{equation*}
    where $B_k = \sum_{m=0}^{k-1} P_\transient^{m} P_\transrec P_\recurrent^{k-1-m}$.

    To prove the first item, we focus on the blocks on the left of $\transmat$. Since the lower left block is zero, we have that $\forall k \geq 1, \forall i \in \recurrent, \forall j \in \transient, (\transmat^k)_{i,j} = 0$. In the upper left block, the element $(P_\transient^k)_{i,j}$ designates the probability of moving from one transient state $i\in\transient$ to another transient state $j\in\transient$ after $k$ iterations. According to \cref{def:generation_process}, if state $i$ is a sequence of $p\geq1$ tokens, state $j$ is necessarily a sequence of $\min\{\cxtsize,p+\cxtsize\} = \cxtsize$ elements. Thus, $P_\transient$ is a nilpotent matrix and $\forall k \geq \cxtsize, \forall i,j, (P_\transient^k)_{i,j} = 0$. This proves that $\forall k \geq \cxtsize, \forall j \in \transient, (\transmat^k)_{i,j} = 0$.

    We now move on to the second item. From the above, $\forall k \geq K, B_k = \sum_{m=0}^{\cxtsize-1} P_\transient^{m} P_\transrec P_\recurrent^{k-1-m}$. Note that this sum is finite, but there is still a dependence on $k$, in the powers of the matrix $P_\recurrent$. In the lower right block, the element $(P_\recurrent^k)_{i,j}$ designates the probability of moving from one recurrent state $i\in\recurrent$ to another recurrent state $j\in\recurrent$ after $k$ iterations. According to the definition of $\transmat$ in \cref{prop:LLM_formal_def} and \cref{def:deletion_process}, $\forall k \geq \cxtsize, \forall i,j \in \recurrent^2, (P_\recurrent^k)_{i,j}$ is nonzero. Exploiting this also in $B_k$, we obtain the result, i.e. $\forall k \geq \cxtsize, \forall j \in \recurrent, (\transmat^k)_{i,j} > 0$.
\end{proof}

We are now ready to prove \cref{prop:ergodic_unichains}, which is inspired by ~\citet{gallager}.

\begin{proof}[Proof of \cref{prop:ergodic_unichains}]
    The states of length strictly less than equal $\cxtsize$ (elements of $\transient$) are transient, because of \cref{def:generation_process}. To discuss the nature of states of length $\cxtsize$ (elements of $\recurrent$), let us introduce a result regarding the powers of the $\transmat$ matrix as defined in \eqref{eq:matrix_q_blocks}. Thanks to \cref{lem:llm_power_k}, the set of states $\recurrent$ (i.e. the states of length $\cxtsize$) form a class. \cref{lem:llm_power_k} gives us also that $\recurrent$ is ergodic. In fact, every state in $\recurrent$ only communicates with all the other states in $\recurrent$, which proves the recurrence. Since $\forall i,j\in\recurrent^2, (\transmat^\cxtsize)_{i,j} > 0$, we can move between any two states in exactly $\cxtsize$ steps, regardless of the initial position. This ensures that $\recurrent$ is aperiodic because the transition probabilities do not depend on a specific cycle, and states can be revisited at various time steps, not just multiples of a particular number. More simply, by considering a token $x$, the state defined as $i = \underbrace{xx\hdots x}_{\text{$\cxtsize$ times}}$ has period $1$, i.e. $(\transmat)_{i,i} > 0$. This is a consequence of \cref{def:deletion_process} and  \cref{prop:LLM_formal_def}. Thanks to \cref{thm:rec_trans_class}, it means that the whole class $\recurrent$ is aperiodic. Finally, this means that $\text{MC}(\mcal{V}_\cxtsize^*, \transmat)$ are ergodic unichains, in the sense of  \cref{def:unichain}.
\end{proof}

\subsection{Proof of \cref{prop:stationary_distrib}}
\label{app:stationary_distrib}

We start by introducing three technical lemmas that will be useful in the proof of \cref{prop:stationary_distrib}. We start with the Chapman--Kolmogorov equation.

\begin{boxlem}[Chapman-Kolmogorov equation]\label{lem:chapman}
    Let $P$ be a matrix of size $d$. Then, $P$ satisfies $$\forall i,j \in [d]^2, \forall n,n' \in \NN^2, (P^{n+n'})_{i,j} = \sum_{k=1}^{d} (P^{n})_{i,k}(P^{n'})_{k,j}.$$
\end{boxlem}
\begin{proof}
    The result follows from the fact that $\forall n,n' \in \NN^2, P^{n+n'} = P^{n}P^{n'}$.
\end{proof}

Then, we introduce a simple but useful result of monotonicity.

\begin{boxlem}[Lemma 3.3.1. in ~\citet{gallager}]\label{lem:non_decreasing_map}
    Let the transition matrix $P$ of a finite state Markov chain.
    Then, for all states $j$ and $n \geq 1$, we have
    \[\underset{i}{\max}~(P^{n+1})_{i,j}\leq \underset{i}{\max}~(P^{n})_{i,j},\quad \text{and}\quad
        \underset{i}{\min}~(P^{n+1})_{i,j}\geq \underset{i}{\min}~(P^{n})_{i,j}.
    \]
\end{boxlem}

We now refer to a result on Markov chains with positive transition matrices.

\begin{boxlem}[Lemma 3.3.2. in~\citet{gallager}]\label{lem:minimal_power}
    Let the transition matrix $P$ of a finite state Markov chain satisfy $\forall i,j, P_{i,j} > 0$, and let $\alpha = \underset{i,j}{\min}~P_{i,j} >0$. Then, for all states $j$ and $n \geq 1$,
    \begin{align*}
        \underset{i}{\max}~(P^{n})_{i,j}-\underset{i}{\min}~(P^{n})_{i,j} & \leq (1-2\alpha)\mleft(\underset{i}{\max}~(P^{n})_{i,j}-\underset{i}{\min}~(P^{n})_{i,j}\mright),                                       \\
        \underset{i}{\max}~(P^{n})_{i,j}-\underset{i}{\min}~(P^{n})_{i,j} & \leq (1-2\alpha)^n,                                       \\
        \lim_{n \to \infty} \underset{i}{\max}~(P^n)_{i,j}                & = \lim_{n \to \infty} \underset{i}{\min}~(P^n)_{i,j} > 0.
    \end{align*}

\end{boxlem}

We are now ready to prove \cref{prop:stationary_distrib} using a similar argument as in~\cite{gallager}.

\begin{proof}[Proof of \cref{prop:stationary_distrib}]
    Let $\transient$ and $\recurrent$ denote respectively the sets of transient and recurrent states. For any state $j$, we define $\pi_j := \lim_{n \to \infty} \max_i~(\transmat^n)_{i,j} = \lim_{n \to \infty} \min_i~(\transmat^n)_{i,j}$. Then $\bm{\pi} = (\pi_j)_{j\in\polish}$ is the stationary distribution for $\transmat$.

    \paragraph{Step 1: Stationary distribution for transient states.}
    \cref{lem:llm_power_k} gives us that $\forall i, \forall k \geq \cxtsize, \forall j \in \transient, (\transmat^k)_{i,j} = 0$. This means that $\forall j\in \transient, \pi_j = 0$ and hence the limit is reached at most after $\cxtsize$ iteration.

    \paragraph{Step 2: Stationary distribution for recurrent states.}
    \cref{lem:llm_power_k} gives us $\forall i,j \in \recurrent^2, (\transmat^\cxtsize)_{i,j} >0$. By defining $\varepsilon := \underset{i,j\in \recurrent^2}{\min}~(\transmat^\cxtsize)_{i,j}$, \cref{lem:minimal_power} shows that for any integer $\ell \geq 1,$ \begin{align}
        \underset{i\in\recurrent}{\max}~(\transmat^{\ell \cxtsize})_{i,j}-\underset{i\in\recurrent}{\min}~(\transmat^{\ell \cxtsize})_{i,j} & \leq (1-2\varepsilon)\mleft(\underset{i\in\recurrent}{\max}~(\transmat^{\ell \cxtsize})_{i,j}-\underset{i\in\recurrent}{\min}~(\transmat^{\ell \cxtsize})_{i,j}\mright),                                                \\
        \underset{i\in\recurrent}{\max}~(\transmat^{\ell \cxtsize})_{i,j}-\underset{i\in\recurrent}{\min}~(\transmat^{\ell \cxtsize})_{i,j} & \leq (1-2\varepsilon)^\ell,\label{eq:bound_nu}                                                \\
        \lim_{\ell \to \infty} \underset{i\in\recurrent}{\max}~(\transmat^{\ell \cxtsize})_{i,j}                                            & = \lim_{\ell \to \infty} \underset{i\in\recurrent}{\min}~(\transmat^{\ell \cxtsize})_{i,j} >0.
        \label{eq:limit_statio}\end{align}
    Thanks to \cref{lem:non_decreasing_map}, $\underset{i}{\max}~(\transmat^{n+1})_{i,j}$ is non-decreasing in $n$, so the limit on the left in \cref{eq:limit_statio} can be replaced with a limit in $n$. The same argument for the limit on the right gives that, $\forall j \in \recurrent$,
    \begin{align*}
        \underset{i\in\recurrent}{\max}~(\transmat^{n})_{i,j}-\underset{i\in\recurrent}{\min}~(\transmat^{n})_{i,j} & \leq (1-2\varepsilon)^{\lfloor n/\cxtsize\rfloor},                              \\
        \lim_{n \to \infty} \underset{i\in\recurrent}{\max}~(\transmat^{n})_{i,j}                                   & = \lim_{n \to \infty} \underset{i\in\recurrent}{\min}~(\transmat^{n})_{i,j} >0,
        \end{align*} 
        where we have taken the floor function to also convert the result of \eqref{eq:bound_nu}. Since $\pi_j$ lies between the minimum and maximum $(\transmat^{n})_{i,j}$ for each $n$, we have that $\forall i,j \in\recurrent^2$, $$\lvert (\transmat^n)_{i,j} - \pi_{j} \rvert \leq (1-2\varepsilon)^{\lfloor\frac{n}{\cxtsize}\rfloor}.$$

    It means that $\forall i,j\in \recurrent^2, \pi_j = \lim_{n \to \infty}(\transmat^n)_{i,j}$. This also gives us the convergence rate when the initial state $i$ is recurrent. In the next step, we consider the general convergence rate, regardless of the nature of the initial state $i$.

    \paragraph{Step 3: Convergence bound.}

    We proceed to the remaining case, i.e. the case where the initial state $i\in\transient$ and the final state $j\in\recurrent$. \cref{lem:chapman} says that $\forall n \geq \cxtsize$, $$(\transmat^n)_{i,j} = \sum_{k\in\transient}(\transmat^\cxtsize)_{i,k}(\transmat^{n-\cxtsize})_{k,j} + \sum_{k\in\recurrent}(\transmat^\cxtsize)_{i,k}(\transmat^{n-\cxtsize})_{k,j}.$$
    We then have that $\forall i \in \transient, \forall n \in \NN,$\begin{align*}
        \lvert (\transmat^n)_{i,j} - \pi_{j} \rvert & \leq \Big\lvert \sum_{k\in\transient}(\transmat^\cxtsize)_{i,k}\big[(\transmat^{n-\cxtsize})_{k,j} - \pi_j\big] + \sum_{k\in\recurrent}(\transmat^\cxtsize)_{i,k}\big[(\transmat^{n-\cxtsize})_{k,j} - \pi_j\big] \Big\rvert \\
        & \leq \sum_{k\in\transient}(\transmat^\cxtsize)_{i,k}\big\lvert(\transmat^{n-\cxtsize})_{k,j} - \pi_j\big\rvert + \sum_{k\in\recurrent}(\transmat^\cxtsize)_{i,k}\big\lvert(\transmat^{n-\cxtsize})_{k,j} - \pi_j\big\rvert   \\
        & \leq \sum_{k\in\transient}(\transmat^\cxtsize)_{i,k} + \sum_{k\in\recurrent}(\transmat^\cxtsize)_{i,k}\big\lvert(\transmat^{n-\cxtsize})_{k,j} - \pi_j\big\rvert                                                             \\
        & \leq (1-2\varepsilon)^{\lfloor\frac{n-\cxtsize}{\cxtsize}\rfloor},
    \end{align*}
    where the first sum vanishes and $\sum_{k\in\recurrent}(\transmat^\cxtsize)_{i,k} \leq 1$.  Finally, $\forall i \in \transient, \forall n \geq \cxtsize,$ $$\lvert (\transmat^n)_{i,j} - \pi_{j} \rvert \leq (1-2\varepsilon)^{\lfloor\frac{n}{\cxtsize}\rfloor - 1}.$$
    Combining this with the result of Step $2$ concludes the proof.
\end{proof}

\subsection{Proof of \cref{cor:sample_complexity}}
\label{app:sample_complexity}
We detail the proof of \cref{cor:sample_complexity} below.
\begin{proof}
We first note that by definition of the total variation distance~\citep{wolfer2019minimax}, we have
\begin{align*}
    \EE_{\mbf{S} \sim \PP_{\mcal{L}}}\lVert \langmat\mleft(\mbf{S}, \cdot \mright) - \transmat\mleft(\mbf{S}, \cdot \mright)\rVert_1 &= \EE_{\mbf{S} \sim \PP_{\mcal{L}}} \mleft[2 \cdot \TV{\langmat\mleft(\mbf{S}, \cdot \mright)}{\transmat\mleft(\mbf{S}, \cdot \mright)} \mright]\\
    &= 2\cdot\EE_{\mbf{S} \sim \PP_{\mcal{L}}} \mleft [ \TV{\langmat\mleft(\mbf{S}, \cdot \mright)}{\transmat\mleft(\mbf{S}, \cdot \mright)} \mright]\\
    &= 2\cdot\risktrain{\params} \tag{by definition of the risk \cref{eq:estimation_error_def}}.
\end{align*}
Applying \cref{thm:pre_training_risk_bound_llm} (a similar result can be derived using \cref{cor:pre_training_risk_bound_llm_depth}), we know that
\begin{equation*}
    \risktrain{\params} \leq \smallrisktrain{\params} +  \frac{\bar{B}}{\sqrt{N_{\train}}}\sqrt{\log{\mleft(\frac{2}{\delta}\mright)}},
\end{equation*}
where $\bar{B}$ is formally defined in \cref{thm:pre_training_risk_bound_llm} (respectively \cref{cor:pre_training_risk_bound_llm_depth}). Assuming a perfect pre-training error amounts to consider $\smallrisktrain{\params} = 0$. We denote by $N^*$ the integer such that the error is equal to $\frac{\epsilon}{2}$, i.e.,
\begin{equation*}
    \frac{\bar{B}}{\sqrt{N^*}}\sqrt{\log{\mleft(\frac{2}{\delta}\mright)}} = \frac{\epsilon}{2} \iff \frac{\bar{B}^2}{N^*}\log{\mleft(\frac{2}{\delta}\mright)} = \frac{\epsilon^2}{4} 
    \iff N^* = \mleft(\frac{2\bar{B}}{\epsilon}\mright)^2\log{\mleft(\frac{2}{\delta}\mright)}.
\end{equation*}
Taking the ceiling function ensures that $N^*$ is an integer. Hence, taking $N_{\train} \geq N^* = \lceil \mleft(\frac{2\bar{B}}{\epsilon}\mright)^2\log{\mleft(\frac{2}{\delta}\mright)} \rceil$ ensures that
\begin{align*}
    \frac{\bar{B}}{\sqrt{N_{\train}}}\sqrt{\log{\mleft(\frac{2}{\delta}\mright)}} \leq \frac{\bar{B}}{\sqrt{N^*}}\sqrt{\log{\mleft(\frac{2}{\delta}\mright)}} = \frac{\epsilon}{2}.
\end{align*}
Putting everything together, taking $N_{\train} \geq N^*$ leads to
\begin{equation*}
    \EE_{\mbf{S} \sim \PP_{\mcal{L}}}\lVert \langmat\mleft(\mbf{S}, \cdot \mright) - \transmat\mleft(\mbf{S}, \cdot \mright)\rVert_1 \leq 2 \cdot \risktrain{\params} \leq 2 \cdot \frac{\epsilon}{2} = \epsilon,
\end{equation*}
which concludes the proof.
\end{proof}

\subsection{Proof of \cref{thm:pre_training_risk_bound_llm}}
\label{app:pre_training_risk_bound_llm}
In this section, we detail the proof of \cref{thm:pre_training_risk_bound_llm}. We provide below an overview of the proof before detailing it.

\paragraph{Overview of the proof.} We are going to use McDiarmid's inequality for dependent random variables of~\citet[Theorem 2.9]{paulin2015ineqMC}. To adapt the arguments of~\citet[Theorem 2.9]{paulin2015ineqMC} to our setting, we bound the total variation between the true probability of the next token and the one estimated by the LLM. The rest of this section is organized as follows. First in \cref{app:thm_pretraining_paulin}, we adapt the concentration inequality of~\citet[Theorem 2.9]{paulin2015ineqMC}. Then in \cref{app:thm_pretraining_bound}, we show how to bound the total variation between the true and the estimated probability of the next token. , in \cref{app:thm_pretraining_conclude}, we restate \cref{thm:pre_training_risk_bound_llm} and conclude the proof.

\subsubsection{Concentration Inequalities for Dependent Random Variables}
\label{app:thm_pretraining_paulin}
We first state a concentration inequality for dependent random variables that will be used to obtain our final bound. 

\begin{boxprop}[McDiarmid's inequality for dependent random variables]
    \label{prop:mcdiarmid_ineq_dependent_rv}
    Let $S \coloneqq \MCseq{S}{N}$ be a sequence of random variables that take values in $\polish = \polish_1 \times \ldots \times \polish_N$. Assume there exists a Marton coupling for $S$ with mixing matrix $\mixmat$. Let $\lVert \mixmat \rVert$ be the operator norm of $\mixmat$. If $f \colon \polish \to \RR$ is such that there exists $\mbf{c} \in \RR^N$ satisfying
    \begin{equation*}
        \forall \mbf{x}, \mbf{y} \in \polish, \quad f(\mbf{x}) - f(\mbf{y}) \leq \sum_{i=1}^N \mbf{c}_i \bbm{1}_{\{\mbf{x}_i \neq \mbf{y}_i\}},
    \end{equation*}
    then we have for any $u \geq 0$,
    \begin{equation*}
        \PP \mleft( \lvert f(S) - \EE_S \mleft[ f(S)\mright] \rvert \geq u \mright) \leq 2\exp{\mleft( \frac{-2u^2}{\lVert \mixmat \rVert^2 \lVert \mbf{c}\rVert_2^2} \mright)}.
    \end{equation*}
\end{boxprop}

\begin{proof}
    Consider a function $f$ verifying the properties of \cref{prop:mcdiarmid_ineq_dependent_rv}. \citet[Theorem 2.9]{paulin2015ineqMC} ensures that for a partition $\hat{S}$ of $S$~\citep[see][Definition 2.3]{paulin2015ineqMC} the following inequality holds
    \begin{equation}
        \label{eq:ineq_mc_non_homogeneous}
        \forall u \geq 0, \quad \PP \mleft( \lvert f(\hat{S}) - \EE \mleft[ f(\hat{S})\mright] \rvert \geq u \mright) \leq 2\exp{\mleft( \frac{-2u^2}{\lVert \mixmat \cdot C(\mbf{c})\rVert_2^2}\mright)},
    \end{equation}
    where $C(\mbf{c})$ is a vector of $\RR^N$ whose $i$-th element is the sum of the $\mbf{c}_j$ such that $j$ is an index of the elements of $\hat{\mbf{S}}_i$. Taking the trivial partition $\hat{S} = S$ implies that the index of the elements in $\hat{\mbf{S}}_i$ are reduced to $\{i\}$. Hence the $i$-th entry of $C(\mbf{c})$ is equal to $\mbf{c}_i$ and $C(\mbf{c}) = \mbf{c}$. By definition of the operator norm (naturally induced by the $\ell_2$-norm), we have
    \[
        \lVert \mixmat\cdot \mbf{c}\rVert_2  =  \frac{\lVert \mixmat \mbf{c}\rVert_2}{\lVert \mbf{c} \rVert_2}  \cdot \lVert \mbf{c} \rVert_2                                                              
                                         \leq \underbrace{\sup_{\mbf{x} \neq 0} \frac{\lVert \mixmat \mbf{x}\rVert_2}{\lVert \mbf{x} \rVert_2}}_{=\lVert \mixmat \rVert} \cdot \lVert \mbf{c} \rVert_2 
                                             \leq \lVert \mixmat \rVert \cdot \lVert \mbf{c}\rVert_2,
    \]
    where the first inequality comes from the fact that $\mbf{c}$ is non-zero (otherwise the only possible $f$ is the zero function which is not of great interest). Using the fact that the function $x \to \exp{(-\frac{2u^2}{x})}$ is increasing, we obtain
    \begin{equation*}
        \exp{\mleft(\frac{-2u^2}{\lVert \mixmat\cdot \mbf{c}\rVert_2^2}\mright)} \leq \exp{\mleft(\frac{-2u^2}{\lVert \mixmat \rVert^2 \cdot \lVert \mbf{c}\rVert_2^2}\mright)},
    \end{equation*}
    which concludes the proof.
\end{proof}
By looking at the definition of the risk $\smallrisktrain{\params}$, we can see that applying \cref{prop:mcdiarmid_ineq_dependent_rv} to the function
\begin{equation*}
    f \colon \MCseq{S}{N_{\train}} = \frac{1}{N_{\train}} \sum_{n=1}^{N_{\train}} \TV{\Pbl{\cdot}{\mbf{S}_n}}{\Pbemp{\params}{\cdot}{\mbf{S}_n}},
\end{equation*}
would lead to the desired bound as we already know $S$ admits a Marton coupling with mixing matrix $\mixmat$. We investigate in the next section how to find the bounding vector $\mbf{c}$ to apply \cref{prop:mcdiarmid_ineq_dependent_rv}.

\subsubsection{Finding the Bounding Vector}
\label{app:thm_pretraining_bound}

\paragraph{Technical lemmas.} We first recall the following important notions from~\citep{tsybakov2008nonparam}. Let $\measurable$ be a measure space and consider two probability distributions $\PP, \QQ$ defined on $\measurable$. For any $\sigma$-finite measure $\nu$ on $\measurable$ such that $\PP, \QQ$ are absolutely continuous with respect to $\nu$, we can define $p = \frac{d\PP}{d\nu}, q=\frac{d\QQ}{d\nu}$ which can also be written as $\PP(d\omega) = q(\omega)\nu(d\omega)$ and $\QQ(d\omega) = p(\omega)\nu(d\omega)$. We will adopt both notations interchangeably. It should be noted that there always exists at least one such measure $\nu$ as one can take $\nu = \PP + \QQ$. With these notations, the squared Hellinger distance between $\PP$ and $\QQ$ is defined as
\begin{equation*}
    \Helling{\PP}{\QQ}^2 \coloneqq \int_{\omega \in \Omega} \mleft( \sqrt{p(\omega)} - \sqrt{q(\omega)}\mright)^2 \nu(d\omega) = \int_{\omega \in \Omega} \mleft( \sqrt{\PP(d\omega)} - \sqrt{\QQ(d\omega)}\mright)^2.
\end{equation*}

The lemma below shows that the total variation between two probability distributions is controlled from above by the absolute value of the logarithm of their ratio.
\begin{boxlem}
    \label{lem:ineq_tv_hellinger}
    Consider two probability distributions $\PP, \QQ$ defined on a measure space $\measurable$ and a $\sigma$-finite measure $\nu$ on $\measurable$. Let $p, q$ be the corresponding probabilities densities, i.e., we have $\PP(d\omega) = q(\omega)\nu(d\omega)$ and $\QQ(d\omega) = p(\omega)\nu(d\omega)$, the total variation between $\PP$ and $\QQ$ satisfies
    \begin{equation*}
        \TV{\PP}{\QQ} \leq \mleft( 2\int_{\omega \in \Omega} \left |\log{\sqrt{\frac{\PP(d\omega)}{\QQ(d\omega)}}}\right | q(\omega)d\nu(d\omega)\mright)^{1/2}.
    \end{equation*}
    If there exists a non-negative constant $B$ such that for any $z \in \Omega$, $\left |\log{\sqrt{\frac{\PP(z)}{\QQ(z)}}}\right | \leq B$,
    then we have
    \begin{equation*}
        \TV{\PP}{\QQ} \leq \sqrt{2B}.
    \end{equation*}
\end{boxlem}

\begin{proof}
    We have the following relation between the total variation and the Hellinger distance~\citep[cf.][Lem.~2.3, Chapt.~2, p.~86]{tsybakov2008nonparam}:
    \begin{equation}
        \label{eq:ineq_tv_hellinger}
        \TV{\PP}{\QQ}^2 \leq \Helling{\PP}{\QQ}^2 \cdot \mleft ( 1 - \underbrace{\Helling{\PP}{\QQ}^2/4}_{\geq 0}\mright) \leq \Helling{\PP}{\QQ}^2,
    \end{equation}
    where the last inequality uses the positivity of the Hellinger distance. Inspired by the decomposition of the Hellinger distance in~\citep[Lem.~25]{agarwal2020flambe}, we have
    \begin{align*}
        \Helling{\PP}{\QQ}^2 & = \int_{\omega \in \Omega} \mleft( \sqrt{\PP(d\omega)} - \sqrt{\QQ(d\omega)}\mright)^2  = \int_{\omega \in \Omega} \mleft( \PP(d\omega) + \QQ(d\omega) - 2\sqrt{\PP(d\omega)}\sqrt{\QQ(d\omega)}\mright)                                          \\
                             & =2\cdot\mleft(1 -  \int_{\omega \in \Omega} \sqrt{\PP(d\omega)}\sqrt{\QQ(d\omega)} \mright)  =2\cdot\mleft(1 -  \int_{\omega \in \Omega} \sqrt{\frac{\PP(d\omega)}{\QQ(d\omega)}} \QQ(d\omega)\mright)                                                 \\
                             & =2\cdot\mleft(1 -  \int_{\omega \in \Omega} \sqrt{\frac{\PP(d\omega)}{\QQ(d\omega)}} q(\omega)d\nu(d\omega)\mright) \tag{by definition of $\QQ(d\omega)$} \\
                             & \leq - 2\log{\mleft( \int_{\omega \in \Omega} \sqrt{\frac{\PP(d\omega)}{\QQ(d\omega)}} q(\omega)d\nu(d\omega)\mright) } \tag{using $1-x \leq -\log{(x)}$}
    \end{align*}
    It follows using \cref{eq:ineq_tv_hellinger}
    \begin{align*}
        \TV{\PP}{\QQ}^2 & \leq \Helling{\PP}{\QQ}^2                                                                                                                                                                                  \\
                        & \leq 2\int_{\omega \in \Omega} -\log{\mleft(\sqrt{\frac{\PP(d\omega)}{\QQ(d\omega)}}\mright)} q(\omega)d\nu(d\omega) \tag{by Jensen as $-\log$ is convex}                                                  \\
                        & \leq 2\left | \int_{\omega \in \Omega} -\log{\mleft(\sqrt{\frac{\PP(d\omega)}{\QQ(d\omega)}}\mright)} q(\omega)d\nu(d\omega)\right |                                                                       \\
                        & \leq 2\int_{\omega \in \Omega} \left | -\log{\mleft(\sqrt{\frac{\PP(d\omega)}{\QQ(d\omega)}}\mright)}\right | q(\omega)d\nu(d\omega) \tag{by Jensen as $\lvert \cdot \rvert$ is convex}                    \\
                        & \leq 2 \int_{\omega \in \Omega} \underbrace{\left |\log{\mleft(\sqrt{\frac{\PP(d\omega)}{\QQ(d\omega)}}\mright)}\right |}_{\leq B} q(\omega)d\nu(d\omega) \tag{first part of \cref{lem:ineq_tv_hellinger}} \\
                        & \leq 2 B \underbrace{\int_{\omega \in \Omega} q(\omega)d\nu(d\omega)}_{=1}            \leq 2B \tag{second part of \cref{lem:ineq_tv_hellinger}}.
    \end{align*}
    This concludes both parts of the proof.
\end{proof}
The next lemma provides a lower bound on the softmax output if its input is upper-bounded (in $\ell_1$-norm).
\begin{boxlem}
    \label{lem:ineq_softmax}
    Let $\mbf{x} \in \RR^m$ be such that $\lVert \mbf{x} \rVert_1 \leq c_1$ for some $c_1 > 0$. Then, we have
    \begin{equation*}
        \softmax{\mbf{u}} \geq \frac{1}{m\exp{\mleft(2c_1\mright)}},
    \end{equation*}
    where the inequality holds for each component of $\softmax{\mbf{u}}$.
\end{boxlem}

\begin{proof}
    Using the fact that
    \begin{equation*}
        \lVert \mbf{x} \rVert_1 = \sum_{i=1}^m \lvert \mbf{x}_i \rvert \leq c_1,
    \end{equation*}
    we know that for any $i \in [m]$, we have
    \begin{equation*}
        -c_1 \leq \mbf{x}_i \leq c_1.
    \end{equation*}
    Hence, using the fact that the exponential is increasing, we have for any $i \in [m]$
    \begin{equation}
        \label{eq:ineq_exp}
        \exp{\mleft(-c_1\mright)} \leq \exp{\mleft(\mbf{x}_i\mright)} \leq \exp{\mleft(c_1\mright)}.
    \end{equation}
    Summing and taking the inverse leads to
    \begin{equation}
        \label{eq:ineq_exp_sum}
        \begin{split}
                 & \sum_{i=1}^m \exp{\mleft(-c_1\mright)} \leq \sum_{i=1}^m \exp{\mleft(\mbf{x}_j\mright)} \leq \sum_{i=1}^m \exp{\mleft(c_1\mright)}                               \\
            \iff & \frac{1}{\sum_{j=1}^m \exp{\mleft(c_1\mright)}} \leq \frac{1}{\sum_{j=1}^m \exp{\mleft(\mbf{x}_j\mright)}} \leq \frac{1}{\sum_{j=1}^m \exp{\mleft(-c_1\mright)}}.
        \end{split}
    \end{equation}
    Combining \cref{eq:ineq_exp} and \cref{eq:ineq_exp_sum} yields
    \begin{equation*}
        \frac{\exp{\mleft(-c_1\mright)}}{\sum_{j=1}^m \exp{\mleft(c_1\mright)}} \leq \frac{\exp{\mleft(\mbf{x}_i\mright)}}{\sum_{j=1}^m \exp{\mleft(\mbf{x}_j\mright)}} \leq \frac{\exp{\mleft(c_1\mright)}}{\sum_{j=1}^m \exp{\mleft(-c_1\mright)}}.
    \end{equation*}
    As we desire a lower bound, we only focus on the left-hand side of the previous inequality. Multiplying the numerator and denominator by $\exp{(c_1)}$ leads to
    \begin{equation*}
        \forall i \in [m], \quad \mrm{softmax}(\mbf{x})_i =  \frac{\exp{\mleft(\mbf{x}_i\mright)}}{\sum_{j=1}^m \exp{\mleft(\mbf{x}_j\mright)}} \geq \frac{1}{m \exp{\mleft(2c_1\mright)}},
    \end{equation*}
    which concludes the proof. While we only need the lower bound of Eq.~\eqref{eq:ineq_exp_sum} to obtain~\cref{lem:ineq_softmax}, both bounds can be used, for instance in~\citet[Lemma E.7]{xie2024mano} and~\citet[Lemma 2.1]{veličković2024softmaxsharp} to show that, under a mild assumption on $\mbf{x} \in \RR^m$, $\mrm{softmax}(\mbf{x})$ behaves as $\mcal{O}\mleft( \frac{1}{m}\mright)$ when $m$ grows to infinity.
\end{proof}

\paragraph{Upper-bounding the total variation.} We now proceed with finding an upper bound on the total variation between the true probability of the next token and the one estimated by the LLM $\model_{\params}$. It will enable us to find the bounding vector $\mbf{c}$. The next lemma shows that the input of the softmax layer of the model is bounded.
\begin{boxlem}
    \label{lem:ineq_last_layer_output}
    Consider an LLM $\model_{\params} \in \funcspace$. For any input sequence $\mbf{S} \in \RR^{\embdim \times \inputsize}$, the following inequality holds
    \begin{equation*}
        \lVert \frac{1}{\inputsize\tau}\mbf{W}_U \mbf{S}^{(L)} \bbm{1}_{\inputsize} \rVert_1 \leq \frac{1}{\tau} \lVert \mbf{W}_U^\top \rVert_{2, 1},
    \end{equation*}
    where $\tau$ is the temperature, $\mbf{W}_U$ is the unembedding matrix (which is bounded as stated in the definition of the parameters space $\paramspace$), and $\mbf{S}^{(L)}$ is the output of the last transformer layer.
\end{boxlem}
\begin{proof}
    We recall that the layer normalization ensures that at each layer, the tokens are in the unit $\ell_2$-ball. This is, in particular, the case for the output of the last layer $\mbf{S}^{(L)}$. It means that the columns of $\mbf{S}^{(L)}$ verifies
    \begin{equation}
        \label{eq:ineq_columns_last_layer}
        \forall k \in [\inputsize], \quad \rVert \mbf{S}^{(L)}_{\cdot, k} \lVert_{2} \leq 1,
    \end{equation}
    which implies
    \begin{equation}
        \label{eq:ineq_ln}
        \max_{1 \leq k \leq \inputsize}{\lVert \mbf{S}^{(L)}_{\cdot, i}\rVert_2} \leq 1.
    \end{equation}
    Recalling that the $L_{p, q}$-norm of a matrix $\mbf{A} \in \RR^{n \times m}$ can be rewritten as
    \begin{equation*}
        \lVert \mbf{A} \rVert_{p, q} \coloneqq \mleft(\sum_{j = 1}^m \mleft(\sum_{i=1}^n \lvert \mbf{A}_{ij}\rvert^p \mright)^{\frac{q}{p}}\mright)^\frac{1}{q} = \lVert \mleft( \lVert \mbf{A}_{\cdot, j} \rVert_p\mright)_{j=1}^m\rVert_q,
    \end{equation*}
which corresponds to 
\begin{equation}
\label{eq:def_l_2_1_norm}
\lVert \mbf{A} \rVert_{2, 1} = \sum_{j=1}^m \lVert \mbf{A}_{\cdot, j} \rVert_p
\end{equation}
with $(p, q) = (2, 1)$. Hence, the $\ell_1$-norm of the last layer before the softmax layer satisfies 
    \begin{align*}
        \lVert \frac{1}{\inputsize\tau}\mbf{W}_U \mbf{S}^{(L)} \bbm{1}_\inputsize \rVert_1 & = \frac{1}{\inputsize\tau} \sum_{i=1}^{\vocabsize} \left| \sum_{j=1}^{\embdim}\mbf{W}_{ij} \sum_{k=1}^{\inputsize}\mbf{S}_{jk}\right| = \frac{1}{\inputsize\tau} \sum_{i=1}^{\vocabsize} \left| \sum_{j=1}^{\embdim} \sum_{k=1}^{\inputsize}\mbf{W}_{ij} \mbf{S}_{jk}\right|                                                    \\
  & \leq \frac{1}{\inputsize\tau} \sum_{i=1}^{\vocabsize} \sum_{j=1}^{\embdim} \sum_{k=1}^{\inputsize} \left | \mbf{W}_{ij} \mbf{S}_{jk}
        \right | \tag{triangular inequality}  \\
& \leq \frac{1}{\inputsize\tau} \sum_{i=1}^{\vocabsize} \sum_{k=1}^{\inputsize} \left | \mbf{W}_{i}^\top \mbf{S}_{\cdot, k} \right |  \\
 & \leq \frac{1}{\inputsize\tau} \sum_{i=1}^{\vocabsize} \sum_{k=1}^{\inputsize} \lVert\mbf{W}_{i} \rVert_2 \lVert\mbf{S}_{\cdot, k} \rVert_2 \tag{Cauchy-Schwartz inequality}
                                                                                    \\
 & \leq \frac{1}{\inputsize\tau} \sum_{i=1}^{\vocabsize} \sum_{k=1}^{\inputsize} \lVert\mbf{W}_{i} \rVert_2 \max_{1 \leq k \leq \inputsize}{\lVert\mbf{S}_{\cdot, k} \rVert_2} \\
 &\leq \frac{1}{\inputsize\tau} \inputsize  \max_{1 \leq k \leq \inputsize}{\lVert\mbf{S}_{\cdot, k} \rVert_2}  \sum_{i=1}^{\vocabsize} \lVert\mbf{W}_{i} \rVert_2 \\
& \leq \frac{1}{\tau} \sum_{i=1}^{\vocabsize} \lVert\mbf{W}_{i} \rVert_2 \\
&\leq \frac{1}{\tau} \sum_{i=1}^{\vocabsize} \lVert(\mbf{W}_U^\top)_{\cdot, i} \rVert_2 \\
&\leq \frac{1}{\tau} \lVert \mbf{W}_U^\top \rVert_{2, 1} \tag{by \cref{eq:ineq_ln} and the def. of $L_{2, 1}$ in \cref{eq:def_l_2_1_norm}} 
    \end{align*}
    where we dropped the subscript and superscript on $\mbf{W}_U$ and $\mbf{S}^{(L)}$ to ease the notations. This concludes the proof.
\end{proof}


The previous lemma can be used to show that the logarithm of the ratio between the true probability of the next token and the one estimated by the LLM $\model_{\params}$ is upper bounded as a function of the vocabulary size $\vocabsize$, the temperature, the upper-bound on $\mbf{W}_U$ and some constant related to the ambiguity of language (see Eq.~\eqref{eq:ambiguity_language}).

\begin{boxprop}[Upper-bound on the logarithm]
    \label{prop:ub_ratio_log}
    Consider an LLM $\model_{\params} \in \funcspace$ with vocabulary size $\vocabsize$. We recall that $B_U$ is the upper bound on the norm of $\mbf{W}_U$ in the definition of parameter space $\paramspace$, $\tau$ is the softmax temperature and $c_0$ is the constant related to the ambiguity of language (see \cref{eq:ambiguity_language}). We have
    \begin{equation*}
        \forall n \in [N], \quad \left|\log{\mleft ( \frac{\Pbl{\mbf{X}_{n+1}}{\mbf{S}_n}}{\Pbemp{\params}{\mbf{X}_{n+1}}{\mbf{S}_n}}\mright)}\right| \leq \cst = \max\{\log{(\vocabsize)} + \frac{2B_U}{\tau}, \log{\mleft(\frac{1}{c_0}\mright)}\}.
    \end{equation*}
\end{boxprop}
\begin{proof}
    The main idea of the proof is to bound the probability ratio and use the fact that $\log$ is non-decreasing. Let $n \in [N]$. The model $\model_{\params}$ receives as input sequences of tokens $\mbf{S}_n$ of size $n \leq \cxtsize$. We first lower-bound each term of the probability ratio. From \cref{eq:ambiguity_language}, we have
    \begin{equation}
        \label{eq:ineq_pb}
        \Pbl{\mbf{X}_{n+1}}{\mbf{S}_n} \geq c_0.
    \end{equation}
    We want to obtain a similar inequality for $\Pbemp{\params}{\mbf{X}_{n+1}}{\mbf{S}_n}$. As the parameters $\params$ of the LLM are in $\paramspace$, we know that $\lVert \mbf{W}_U^\top \rVert_{2, 1} \leq B_U$. \cref{lem:ineq_last_layer_output} ensures that
    \begin{equation*}
        \lVert \frac{1}{\inputsize\tau}\mbf{W}_U \mbf{S}^{(L)} \bbm{1}_\inputsize \rVert_1 \leq \frac{1}{\tau} \lVert \mbf{W}_U^\top \rVert_{2, 1} \leq \frac{B_U}{\tau}.
    \end{equation*}
    We can then apply \cref{lem:ineq_softmax} with $c_1= \frac{B_U}{\tau}$ and given that $\frac{1}{\inputsize\tau}\mbf{W}_U \mbf{S}^{(L)} \bbm{1}_\inputsize \in \RR^{\vocabsize}$, it leads to
    \begin{equation*}
        \Pbemp{\params}{\cdot}{\mbf{S}_n} = \softmax{\frac{1}{\inputsize\tau}\mbf{W}_U \mbf{S}^{(L)} \bbm{1}_\inputsize} \geq \frac{1}{\vocabsize \exp{(2B_U/\tau)}},
    \end{equation*}
    where the inequality holds for each component of $\Pbemp{\params}{\cdot}{\mbf{S}_n}$. This is in particular the case for $\Pbemp{\params}{\mbf{X}_{n+1}}{\mbf{S}_n}$ which is the entry we are interested in, i.e., we have
    \begin{equation}
        \label{eq:ineq_pbemp}
        \Pbemp{\params}{\mbf{X}_{n+1}}{\mbf{S}_n} \geq \frac{1}{\vocabsize \exp{(2B_U/\tau)}}.
    \end{equation}
    Going back to the ratio of probability, consider the situation where we have
    \begin{equation*}
        \frac{\Pbl{\mbf{X}_{n+1}}{\mbf{S}_n}}{\Pbemp{\params}{\mbf{X}_{n+1}}{\mbf{S}_n}} \geq 1.
    \end{equation*}
    Then, using Eq.~\eqref{eq:ineq_pbemp}, we have
    \begin{equation*}
        1 \leq \frac{\Pbl{\mbf{X}_{n+1}}{\mbf{S}_n}}{\Pbemp{\params}{\mbf{X}_{n+1}}{\mbf{S}_n}} \leq \frac{1}{\Pbemp{\params}{\mbf{X}_{n+1}}{\mbf{S}_n}} \leq \vocabsize \exp{(2B_U/\tau)},
    \end{equation*}
    which implies, as the $\log$ is non-decreasing monotonically,
    \begin{equation}
        \label{eq:ineq_log_rat_pos}
        0 \leq \log{\mleft( \frac{\Pbl{\mbf{X}_{n+1}}{\mbf{S}_n}}{\Pbemp{\params}{\mbf{X}_{n+1}}{\mbf{S}_n}} \mright)} \leq \log{(\vocabsize \exp{(2B_U/\tau)})} = \log{(\vocabsize)} + \frac{2B_U}{\tau}.
    \end{equation}
    Similarly, consider the case where we have
    \begin{equation*}
        \frac{\Pbl{\mbf{X}_{n+1}}{\mbf{S}_n}}{\Pbemp{\params}{\mbf{X}_{n+1}}{\mbf{S}_n}} \leq 1.
    \end{equation*}
    Then, we have
    \begin{equation*}
        \frac{\Pbemp{\params}{\mbf{X}_{n+1}}{\mbf{S}_n}}{\Pbl{\mbf{X}_{n+1}}{\mbf{S}_n}} \geq 1,
    \end{equation*}
    and similarly to above, we can use Eq.~\eqref{eq:ineq_pb} to obtain
    \begin{equation*}
        1 \leq \frac{\Pbemp{\params}{\mbf{X}_{n+1}}{\mbf{S}_n}}{\Pbl{\mbf{X}_{n+1}}{\mbf{S}_n}} \leq \frac{1}{\Pbl{\mbf{X}_{n+1}}{\mbf{S}_n}} \leq \frac{1}{c_0}.
    \end{equation*}
    This implies
    \begin{equation*}
        0 \leq \log{\mleft( \frac{\Pbemp{\params}{\mbf{X}_{n+1}}{\mbf{S}_n}}{\Pbl{\mbf{X}_{n+1}}{\mbf{S}_n}} \mright)} \leq \log{\mleft(\frac{1}{c_0}\mright)},
    \end{equation*}
    which also rewrites
    \begin{equation}
        \label{eq:ineq_log_rat_neg}
        0 \leq - \log{\mleft( \frac{\Pbl{\mbf{X}_{n+1}}{\mbf{S}_n}}{\Pbemp{\params}{\mbf{X}_{n+1}}{\mbf{S}_n}} \mright)} \leq \log{\mleft(\frac{1}{c_0}\mright)}.
    \end{equation}
    By definition of the absolute value, combining Eq.~\eqref{eq:ineq_log_rat_pos} and Eq.~\eqref{eq:ineq_log_rat_neg} leads to
    \begin{equation*}
        \left | \log{\mleft( \frac{\Pbl{\mbf{X}_{n+1}}{\mbf{S}_n}}{\Pbemp{\params}{\mbf{X}_{n+1}}{\mbf{S}_n}} \mright)} \right | \leq \max\{\log{(\vocabsize)} + \frac{2B_U}{\tau}, \log{\mleft(\frac{1}{c_0}\mright)}\}.
    \end{equation*}
    This concludes the proof.
\end{proof}

We are now ready to upper-bound the total variation.

\begin{boxcor}[Upper-bound on the total variation]
    \label{cor:ub_tv}
    Consider an LLM $\model_{\params} \in \funcspace$ with vocabulary size $\vocabsize$. We recall that $B_U$ is the upper bound on the norm of $\mbf{W}_U$ in the definition of parameter space $\paramspace$, $\tau$ is the softmax temperature and $c_0$ is the constant related to the ambiguity of language (see \cref{eq:ambiguity_language}). For $n \in [N]$, we have
    \begin{equation}
        \label{eq:ub_tv_1}
        \TV{\Pbl{\cdot}{\mbf{S}_n}}{\Pbemp{\params}{\cdot}{\mbf{S}_n}} \leq \sqrt{2\max\{\log{(\vocabsize)} + \frac{2B_U}{\tau}, \log{\mleft(\frac{1}{c_0}\mright)}\}} \coloneqq c_2.
    \end{equation}
\end{boxcor}
\begin{proof}
    Using \cref{prop:ub_ratio_log}, we can directly apply \cref{lem:ineq_tv_hellinger} with $B = \max\{\log{(\vocabsize)} + \frac{2B_U}{\tau}, \log{\mleft(\frac{1}{c_0}\mright)}\}$ for any $n \in [N]$. This leads to
    \begin{equation*}
        \forall n \in [N], \quad \TV{\Pbl{\cdot}{\mbf{S}_n}}{\Pbemp{\params}{\cdot}{\mbf{S}_n}} \leq \sqrt{2 \max\{\log{(\vocabsize)} + \frac{2B_U}{\tau}, \log{\mleft(\frac{1}{c_0}\mright)}\}}.
    \end{equation*}
    This concludes the proof.
\end{proof}

\subsubsection{Concluding the Proof}
\label{app:thm_pretraining_conclude}
We are now ready to state our main result.

\begin{boxthm}[Restatement of \cref{thm:pre_training_risk_bound_llm}]
    \label{thm:restate_pre_training_risk_bound_llm}
    Consider an LLM $\model_{\params} \in \funcspace$ with vocabulary size $\vocabsize$. We denote by $\mixmat$ the mixing matrix of the pretraining sequences of tokens $\MCseq{S}{N_{\train}}$. Let $\delta > 0$. Then, with probability at least $1-\delta$,
    \begin{equation*}
        \risktrain{\params} \leq \smallrisktrain{\params} +  \frac{\bar{B}}{\sqrt{N_{\train}}}\sqrt{\log{\mleft(\frac{2}{\delta}\mright)}},
    \end{equation*}
    where $\bar{B}$ is a constant depending on the parameters of the problem. More precisely,
    \begin{equation*}
        \bar{B} = 2\lVert \mixmat \rVert\sqrt{\max\{\log{(\vocabsize)} +\frac{2B_U}{\tau}, \log{\mleft(\frac{1}{c_0}\mright)}\}}.
    \end{equation*}
\end{boxthm}

\begin{proof}[Proof of \cref{thm:pre_training_risk_bound_llm}]
    By definition of the risk, we have
    \begin{align*}
        \smallrisktrain{\params} & = \frac{1}{N_{\train}} \sum_{n=1}^{N_{\train}} \underbrace{\TV{\Pbl{\cdot}{\mbf{S}_n}}{\Pbemp{\params}{\cdot}{\mbf{S}_n}}}_{=g_n\mleft(\mbf{S}_n\mright)} = \frac{1}{N_{\train}} \sum_{n=1}^{N_{\train}} g_n\mleft(\mbf{S}_n\mright)                                                                               \\
                                 & = f\MCseq{S}{N_{\train}} = f\mleft(S\mright).
    \end{align*}
    Using \cref{cor:ub_tv}, we know that
    \begin{equation*}
        \lvert g_n\mleft(\mbf{S}_n\mright) \rvert \leq \sqrt{2\max\{\log{(\vocabsize)} + \frac{2B_U}{\tau}, \log{\mleft(\frac{1}{c_0}\mright)}\}} \coloneqq c_2.
    \end{equation*}
    By definition, each sequence of tokens $\mbf{S}_n$ takes its values in $\vocabspace^n$ (again by abuse of notation, $n = \min\{n, \cxtsize\}$) and $S$ takes its values in $\vocabspace^1 \times \ldots \times \vocabspace^{N_\train}$. 
    For any two sequences $\zeta, \Sigma$ with values in $\vocabspace^1 \times \ldots \times \vocabspace^{N_\train}$, we have
    \begin{align*}
        f\mleft(\zeta\mright) - f\mleft(\Sigma\mright) & =  \frac{1}{N_{\train}} \sum_{n=1}^{N_{\train}} \mleft(\underbrace{\TV{\Pbl{\cdot}{\bm{\zeta}_n}}{\Pbemp{\params}{\cdot}{\bm{\zeta}_n}}}_{=g_n\mleft(\bm{\zeta}_n\mright)} - \underbrace{\TV{\Pbl{\cdot}{\bm{\Sigma}_n}}{\Pbemp{\params}{\cdot}{\bm{\Sigma}_n}}}_{=g_n\mleft(\bm{\Sigma}_n\mright)} \mright) \\
                                                   & = \frac{1}{N_{\train}} \sum_{n=1}^{N_{\train}} \mleft( g_n\mleft(\bm{\zeta}_n\mright) - g_n\mleft(\bm{\Sigma}_n\mright)\mright)                                                                                                                                                                      \\
                                                   & = \frac{1}{N_{\train}} \sum_{n=1}^{N_{\train}} \mleft( g_n\mleft(\bm{\zeta}_n\mright) - g_n\mleft(\bm{\Sigma}_n\mright)\mright) \bbm{1}_{\{\bm{\zeta}_n \neq \bm{\Sigma}_n\}} \tag{removing the zero terms}                                                                                             \\
                                                   & \leq \left | \frac{1}{N_{\train}} \sum_{n=1}^{N_{\train}} \mleft( g_n\mleft(\bm{\zeta}_n\mright) - g_n\mleft(\bm{\Sigma}_n\mright)\mright) \bbm{1}_{\{\bm{\zeta}_n \neq \bm{\Sigma}_n\}} \right |                                                                                                       \\
                                                   & \leq \frac{1}{N_{\train}} \sum_{n=1}^{N_{\train}} \left | \mleft( g_n\mleft(\bm{\zeta}_n\mright) - g_n\mleft(\bm{\Sigma}_n\mright)\mright) \bbm{1}_{\{\bm{\zeta}_n \neq \bm{\Sigma}_n\}} \right |                                                                                                       \\
                                                   & \leq \frac{1}{N_{\train}} \sum_{n=1}^{N_{\train}} \left | g_n\mleft(\bm{\zeta}_n\mright) - g_n\mleft(\bm{\Sigma}_n\mright)\right | \bbm{1}_{\{\bm{\zeta}_n \neq \bm{\Sigma}_n\}}                                                                                                                        \\
                                                   & \leq \frac{1}{N_{\train}} \sum_{n=1}^{N_{\train}} \mleft(\underbrace{\left | g_n\mleft(\bm{\zeta}_n\mright) \right |}_{\leq c_2} + \underbrace{\left | g_n\mleft(\bm{\Sigma}_n\mright)\right |}_{\leq c_2} \mright) \bbm{1}_{\{\bm{\zeta}_n \neq \bm{\Sigma}_n\}} \tag{\cref{cor:ub_tv}}                \\
                                                   & \leq \frac{1}{N_{\train}} \sum_{n=1}^{N_{\train}} 2 c_2\bbm{1}_{\{\bm{\zeta}_n \neq \bm{\Sigma}_n\}}              = \sum_{n=1}^{N_{\train}} \mleft(\frac{2c_2}{N_{\train}}\mright) \bbm{1}_{\{\bm{\zeta}_n \neq \bm{\Sigma}_n\}},
    \end{align*}
    where $c_2 = \sqrt{2\max\{\log{(\vocabsize)} + \frac{2B_U}{\tau}, \log{\mleft(\frac{1}{c_0}\mright)}\}}$. As $\zeta$ and $\Sigma$ were taken arbitrary, choosing $\mbf{c} \in \RR^{N_{\train}}$ with all entries equal to $\frac{2c_2}{N_{\train}}$ ensures that $f$ verifies the condition in \cref{prop:mcdiarmid_ineq_dependent_rv}, i.e.,
    \begin{equation*}
        \forall \zeta, \Sigma, \quad f(\zeta) - f(\Sigma) \leq \sum_{n=1}^{N_{\train}} \mbf{c}_n \bbm{1}_{\{\bm{\zeta}_n \neq \bm{\Sigma}_n\}}.
    \end{equation*}
    We assumed in \cref{sec:pretraining} that the sequences $\mbf{S}_n$ were related via a Marton coupling with mixing matrix $\mixmat$. Putting everything together, we can apply \cref{prop:mcdiarmid_ineq_dependent_rv} which leads to
    \begin{equation}
        \label{eq:thm_pretraining_paulin_abs}
        \forall u \geq 0, \quad \PP \mleft( \lvert f(S) - \EE_S \mleft[ f(S)\mright] \rvert \geq u \mright) \leq 2\exp{\mleft( \frac{-2u^2}{\lVert \mixmat \rVert^2 \lVert \mbf{c}\rVert_2^2} \mright)}.
    \end{equation}
    Let $u \geq 0$. We have the following events ordering
    \begin{align*}
        \mleft( \EE_S \mleft[ f(S)\mright] - f(S) \geq u \mright) & \subseteq \mleft( \EE_S \mleft[ f(S)\mright] - f(S) \geq u \mright) \cup \mleft(f(S) - \EE_S \mleft[ f(S)\mright] \geq u \mright) \\
                                                                  & = \mleft(\left | f(S) - \EE_S \mleft[ f(S)\mright] \right | \geq u\mright).
    \end{align*}
    Hence, as $u$ was taken arbitrary and using \cref{eq:thm_pretraining_paulin_abs}, we have
    \begin{equation*}
        \forall u \geq 0, \quad \PP \mleft( \EE_S \mleft[ f(S)\mright] - f(S) \geq u \mright) \leq 2\exp{\mleft( \frac{-2u^2}{\lVert \mixmat \rVert^2 \lVert \mbf{c}\rVert_2^2} \mright)}.
    \end{equation*}
    We recall that by definition
    \begin{equation*}
        f(S) = \smallrisktrain{\params} \text{ and } \risktrain{\params} = \EE_S \mleft[ \smallrisktrain{\params}\mright].
    \end{equation*}
    Since the previous inequality holds for any $u \geq 0$, we can hence choose $u$ such that
    \begin{align*}
        \delta = 2\exp{\mleft( \frac{-2u^2}{\lVert \mixmat \rVert^2 \lVert \mbf{c}\rVert_2^2} \mright)} & \iff \frac{-2u^2}{\lVert \mixmat \rVert^2 \lVert \mbf{c}\rVert_2^2} = \log{\mleft( \frac{\delta}{2}\mright)} \iff u^2 = \frac{1}{2}\lVert \mixmat \rVert^2 \lVert \mbf{c}\rVert_2^2 \log{\mleft( \frac{2}{\delta}\mright)} \\
                                                                                                        & \iff u = \frac{1}{\sqrt{2}}\lVert \mixmat \rVert \lVert \mbf{c}\rVert_2 \sqrt{\log{\mleft( \frac{2}{\delta}\mright)}}.
    \end{align*}
    Using the fact that
    \[
        \lVert \mbf{c}\rVert_2 = \sqrt{\sum_{n=1}^{N_\train} \mbf{c}_n^2} = \sqrt{\sum_{n=1}^{N_\train} \mleft( \frac{2c_2}{N_{\train}}\mright)^2}= \sqrt{\sum_{n=1}^{N_\train} \frac{4c_2^2}{N_{\train}^2}}= \sqrt{\frac{4c_2^2}{N_{\train}}}= \frac{2c_2}{\sqrt{N_{\train}}}\]
    and using the fact that $c_2 = \sqrt{2\max\{\log{(\vocabsize)} + \frac{2B_U}{\tau}, \log{\mleft(\frac{1}{c_0}\mright)}\}}$ from \cref{cor:ub_tv}, we obtain
    \begin{align*}
        u & = \frac{1}{\sqrt{2}} \frac{2c_2}{\sqrt{N_{\train}}} \lVert \mixmat \rVert \sqrt{\log{\mleft( \frac{2}{\delta}\mright)}} = \frac{\sqrt{2}c_2}{\sqrt{N_{\train}}} \lVert \mixmat \rVert \sqrt{\log{\mleft( \frac{2}{\delta}\mright)}}                                          \\
          & = \frac{2\lVert \mixmat \rVert \sqrt{\max\{\log{(\vocabsize)} + \frac{2B_U}{\tau}, \log{\mleft(\frac{1}{c_0}\mright)}\}}}{\sqrt{N_{\train}}}  \sqrt{\log{\mleft( \frac{2}{\delta}\mright)}}= \frac{\bar{B}}{\sqrt{N_{\train}}}\sqrt{\log{\mleft( \frac{2}{\delta}\mright)}},
    \end{align*}
    where we define
    \begin{equation*}
        \bar{B} = 2\lVert \mixmat \rVert \sqrt{\max\{\log{(\vocabsize)} + \frac{2B_U}{\tau}, \log{\mleft(\frac{1}{c_0}\mright)}\}}.
    \end{equation*}
    Putting everything together, we have
    \begin{equation*}
        \PP \mleft( \risktrain{\params} - \smallrisktrain{\params} \geq \frac{\bar{B}}{\sqrt{N_{\train}}}\sqrt{\log{\mleft( \frac{2}{\delta}\mright)}} \mright) \leq \delta.
    \end{equation*}
    Taking the opposite event leads to the following inequality with probability at least $1-\delta$
    \begin{equation*}
        \risktrain{\params} \leq \smallrisktrain{\params} + \frac{\bar{B}}{\sqrt{N_{\train}}}\sqrt{\log{\mleft( \frac{2}{\delta}\mright)}},
    \end{equation*}
    which concludes the proof.
\end{proof}

\subsection{Proof of \cref{cor:pre_training_risk_bound_llm_depth}}
\label{app:pre_training_risk_bound_llm_depth}
 As the layer norm is not applied anymore, each token is no longer in the $\ell_2$-unit ball, and \cref{lem:ineq_last_layer_output} does not hold anymore. We want to provide an analogous lemma for our setting. We first prove the following technical lemmas.
\begin{boxlem}
    \label{lem:ineq_relu}
    The $\mrm{ReLU}$ is a norm-decreasing operator, i.e., we have
    \[
        \forall \mbf{A} \in \RR^{n \times m}, \quad \lVert \mrm{ReLU} \mleft(\mbf{A}\mright) \rVert_{1, 1} \leq \lVert \mbf{A} \rVert_{1, 1},
    \]
    where the $\mrm{ReLU}$ is applied entry-wise.
\end{boxlem}
\begin{proof}
    Recalling that $\mrm{ReLU}(x) = \max\{0, x\}$ is applied entry-wise, using the fact that $\lvert \max\{0, x\} \rvert \leq \lvert x \rvert$ and considering $\mbf{A}$ and $\tilde{\mbf{A}} = \mrm{ReLU}\mleft(\mbf{A}\mright)$, we have
    \begin{equation*}
        \lVert \tilde{\mbf{A}} \rVert_{1, 1} = \sum_{i, j} \lvert \tilde{\mbf{A}}_{i, j}\rvert = \sum_{i, j} \lvert \max\{0, \tilde{\mbf{A}}_{i, j} \}\rvert \leq \sum_{i, j} \lvert \mbf{A}_{i, j}\rvert \leq \lVert \mbf{A} \rVert_{1, 1},
    \end{equation*}
    which concludes the proof.
\end{proof}

\begin{boxlem}
    \label{lem:prop_l_11}
    The $L_{1, 1}$-norm verifies the following property:
    \begin{equation*}
        \forall \mbf{A} \in \RR^{n \times m}, \mbf{B} \in \RR^{m \times p}, \quad \lVert \mbf{AB} \rVert_{1, 1} \leq n \lVert \mbf{A}\rVert_{\infty} \lVert \mbf{B}\rVert_{1, 1}.
    \end{equation*}
\end{boxlem}
\begin{proof}
    We have
    \begin{align*}
        \lVert \mbf{AB} \rVert_{1, 1} & = \sum_{j=1}^p \sum_{i=1}^n \lvert \mleft(\mbf{AB}\mright)_{ij} \rvert = \sum_{j=1}^p \sum_{i=1}^n \lvert \sum_{k=1}^m \mbf{A}_{ik} \mbf{B}_{kj} \rvert \leq \sum_{j=1}^p \sum_{i=1}^n \sum_{k=1}^m \lvert \mbf{A}_{ik} \mbf{B}_{kj} \rvert \\
        & \leq \sum_{j=1}^p \sum_{i=1}^n \sum_{k=1}^m \lvert \mbf{A}_{ik} \rvert \lvert \mbf{B}_{kj} \rvert \leq \max_{ik}\lvert \mbf{A}_{ik} \rvert \sum_{j=1}^p \sum_{i=1}^n \sum_{k=1}^m \lvert \mbf{B}_{kj} \rvert                 \\
        & \leq n \lVert \mbf{A} \rVert_{\infty} \sum_{j=1}^p \sum_{k=1}^m  \lvert \mbf{B}_{kj} \rvert\leq n \lVert \mbf{A}\rVert_{\infty} \lVert \mbf{B}\rVert_{1, 1},
    \end{align*}
    which concludes the proof.
\end{proof}

\begin{boxlem}
    \label{lem:ineq_l_21_infty}
    The $L_{2, 1}$ and $L_{\infty, 1}$-norms verify the following relation
    \begin{equation*}
        \forall \mbf{A} \in \RR^{n \times m}, \quad \lVert \mbf{A} \rVert_{\infty, 1} \leq \lVert \mbf{A} \rVert_{2, 1}.
    \end{equation*}
\end{boxlem}
\begin{proof}
    By definition of the $L_{p, q}$-norm, we have
    \begin{align*}
        \lVert \mbf{A} \rVert_{\infty, 1} & = \sum_{j=1}^M \max_{1 \leq i \leq n} \lvert \mbf{A}_{ij} \rvert= \sum_{j=1}^M \sqrt{\max_{1 \leq i \leq n} \lvert \mbf{A}_{ij}^2} \rvert \tag{as $x \to x^2$ is increasing} \\
                                          & \leq \sum_{j=1}^M \sqrt{\sum_{i=1}^n \lvert \mbf{A}_{ij}^2} \rvert \leq \sum_{j=1}^M \lVert \mbf{A}_{\cdot, j}\rVert_2 \leq \lVert \mbf{A} \rVert_{2, 1},
    \end{align*}
    where the first inequality comes from adding non-negative terms.
\end{proof}
We are now ready to state the lemma analogous to \cref{lem:ineq_last_layer_output}.
\begin{boxlem}
    \label{lem:ineq_last_layer_output_depth}
    Consider an LLM $\model_{\params} \in \tilde{F}$ with $L$ layers. For any input sequence $\mbf{S} \in \RR^{\embdim \times \inputsize}$, the following inequality holds
    \begin{equation*}
        \lVert \frac{1}{\inputsize\tau}\mbf{W}_U \mbf{S}^{(L)} \bbm{1}_{\inputsize} \rVert_1 \leq \frac{c_3}{\tau} \lVert \mbf{W}_U^\top \rVert_{2, 1},
    \end{equation*}
    where $\tau$ is the temperature and $c_3$ is a constant depending on the parameters upper-bound. More precisely,
    \begin{equation*}
        c_3 = \mleft[\mleft( 1+ \embdim \hiddendim B_1B_2\mright) \cdot \mleft( 1 + \frac{\embdim^3}{H} B_OB_V \mright)\mright]^L \cdot \ubtok.
    \end{equation*}$\mbf{W}_U$ is the unembedding matrix (which is bounded as stated in the definition of the parameters space $\paramspace$), and $\mbf{S}^{(L)}$ is the output of the last transformer layer.
\end{boxlem}
\begin{proof}[Proof of \cref{lem:ineq_last_layer_output_depth}]
    Our model $f_{\params} \in \tilde{\funcspace}$ is given as input a sequence $\mbf{S} \in \RR^{\embdim \times \inputsize}$. With similar computations than in \cref{lem:ineq_last_layer_output}, we have
    \begin{align*}
        \frac{1}{\inputsize\tau}\lVert \mbf{W}_U \mbf{S}^{(L)} \bbm{1}_\inputsize \rVert_1 & = \frac{1}{\inputsize\tau} \sum_{i=1}^{\vocabsize} \left| \sum_{j=1}^{\embdim}\mbf{W}_{ij} \sum_{k=1}^{\inputsize}\mbf{S}_{jk}\right| = \frac{1}{\inputsize\tau} \sum_{i=1}^{\vocabsize} \left| \sum_{j=1}^{\embdim} \sum_{k=1}^{\inputsize}\mbf{W}_{ij} \mbf{S}_{jk}\right| \notag                                                   \\
                                                                                           & \leq \frac{1}{\inputsize\tau} \sum_{i=1}^{\vocabsize} \sum_{j=1}^{\embdim} \sum_{k=1}^{\inputsize} \left | \mbf{W}_{ij} \mbf{S}_{jk}
        \right | \tag{triangular inequality}                                                                                                                                                                                                                                                                                                                                                                       \\
                                                                                           & \leq \frac{1}{\inputsize\tau} \sum_{i=1}^{\vocabsize} \sum_{k=1}^{\inputsize} \left | \mbf{W}_{i}^\top \mbf{S}_{\cdot, k}\right |  \leq \frac{1}{\inputsize\tau} \sum_{i=1}^{\vocabsize} \sum_{k=1}^{\inputsize} \lVert\mbf{W}_{i} \rVert_{\infty} \lVert\mbf{S}_{\cdot, k} \rVert_1 \tag{H\"older inequality} \notag \\
                                                                                           & \leq \frac{1}{\inputsize\tau} \mleft(\sum_{i=1}^{\vocabsize} \lVert\mbf{W}_{i} \rVert_{\infty} \mright) \cdot \mleft(\sum_{k=1}^{\inputsize} \lVert\mbf{S}_{\cdot, k} \rVert_1\mright) \leq \frac{1}{\inputsize\tau} \lVert \mbf{W}_U^\top \rVert_{\infty, 1} \lVert \mbf{S}^{(L)}\rVert_{1, 1}                       \\
                                                                                           & \leq \frac{1}{\inputsize\tau} \lVert \mbf{W}_U^\top \rVert_{2, 1} \lVert \mbf{S}^{(L)}\rVert_{1, 1}, \tag{\cref{lem:ineq_l_21_infty}}
    \end{align*}
    where, again, we dropped the subscript and superscript on $\mbf{W}_U$ and $\mbf{S}^{(L)}$ to ease the notations. We obtain
    \begin{equation}
        \label{eq:ineq_output_depth_step_0}
        \lVert \frac{1}{\inputsize\tau}\mbf{W}_U \mbf{S}^{(L)} \bbm{1}_\inputsize \rVert_1 \leq \frac{1}{\inputsize\tau} \lVert \mbf{W}_U^\top \rVert_{2, 1} \lVert \mbf{S}^{(L)}\rVert_{1, 1}.
    \end{equation}
    As we do not use layer normalization, we want to find another way to bound $\mbf{S}^{(L)}$. To that end, we will first express $\mbf{S}^{(\ell)}$, the output of the $(\ell)$-th layer of the transformer, as a function of $\mbf{S}^{(\ell-1)}$, the output of the $(\ell-1)$-th layer. Using the definition of the transformer model (see \cref{app:background_transformer}), we have
    \begin{equation*}
        \begin{cases}
             & \mbf{Z}^{(\ell)} = \mbf{S}^{(\ell-1)} +  \mcal{A}\mleft(\mbf{S}^{(\ell-1)}; \mbf{W}_Q^{(\ell)}, \mbf{W}_K^{(\ell)}, \mbf{W}_V^{(\ell)}, \mbf{W}_O^{(\ell)} \mright), \\
             & \mbf{Y}^{(\ell)} = \mbf{W}_2^{(\ell)} \relu{\mbf{W}_1^{(\ell)}\mbf{Z}^{(\ell)}},                                                                                     \\
             & \mbf{S}^{(\ell)} = \mbf{Z}^{(\ell)} + \mbf{Y}^{(\ell)}.
        \end{cases}
    \end{equation*}
    We will compute each layer's $L_{1,1}$-norm.

    \textbf{Step 1: MHA.} By definition, denoting the number of heads by $H$, we know that $\mcal{A}\mleft(\mbf{S}^{(\ell-1)}; \mbf{W}_Q^{(\ell)}, \mbf{W}_K^{(\ell)}, \mbf{W}_V^{(\ell)}, \mbf{W}_O^{(\ell)} \mright) \in \RR^{\embdim \times \inputsize}$ multiplies $\mbf{W}^{(\ell)} \in \RR^{\embdim \times \embdim}$ with the concatenation on the rows of the $H$ softmax layers that each writes
    \begin{equation*}
        \softmax{\mbf{W}^{(\ell)}_Q\mbf{S}^{(\ell)}\mleft( \mbf{W}^{(\ell)}_K\mbf{S}^{(\ell-1)}\mright)^\top/ \sqrt{\embdim}}\mleft(\mbf{W}^{(\ell)}_V\mbf{S}^{(\ell-1)}\mright) \in \RR^{\frac{\embdim}{H} \times \inputsize},
    \end{equation*}
    We keep the notations $\ell$ without explicating the index of the head to ease notations. Denoting the concatenation on the rows by $\mbf{C}^{(\ell)} \in \RR^{\embdim \times \inputsize}$, we have
    \begin{align*}
        \lVert \mcal{A}\mleft(\mbf{S}^{(\ell-1)}; \mbf{W}_Q^{(\ell)}, \mbf{W}_K^{(\ell)}, \mbf{W}_V^{(\ell)}, \mbf{W}_O^{(\ell)} \mright) \rVert_{1, 1} & = \lVert \mbf{W}_O^{(\ell)} \mbf{C}^{(\ell)} \rVert_{1, 1} \leq r \cdot \lVert \mbf{W}_O^{(\ell)} \rVert_{\infty} \lVert \mbf{C}^{(\ell)} \rVert_{1, 1} \\
  & \leq rB_O \lVert \mbf{C}^{(\ell)} \rVert_{1, 1}. \tag{definition of $\tilde{\paramspace}$}
    \end{align*}
    Moreover, by definition of $\mbf{C}^{(\ell)}$, we have
    \begin{equation}
        \label{eq:softmax_head}
        \lVert \mbf{C}^{(\ell)} \rVert_{1, 1} = \sum_{j=1}^{\embdim} \sum_{i=1}^{\embdim} \lvert \mbf{C}^{(\ell)}_{ij} \rvert = \sum_{j=1}^{\embdim} \sum_{i=1}^{\embdim/H} \sum_{h=1}^H \lvert \mbf{C}^{(\ell, h)}_{ij} \rvert = \sum_{h=1}^H \lVert \mbf{C}^{(\ell, h)} \rVert_{1, 1},
    \end{equation}
    where $\mbf{C}^{(\ell, h)} \in \RR^{\frac{\embdim}{H} \times \inputsize}$ is the softmax matrix of the $h$-th layer. We recall that the softmax matrix is a row-stochastic matrix of $\RR^{\frac{\embdim}{H} \times \embdim}$ so it has all values lower than $1$. In the next computations, we drop the $h$ index on the query, key, and value matrices to ease the notations. Using \cref{lem:prop_l_11} on the softmax matrix and on the value matrix $\mbf{W}_V^{(\ell)} \in \RR^{\frac{\embdim}{H} \times \embdim}$, we have
    \begin{align*}
        \lVert \mbf{C}^{(\ell, h)} \rVert_{1, 1} & =\lVert \softmax{\mbf{W}^{(\ell)}_Q\mbf{S}^{(\ell)}\mleft( \mbf{W}^{(\ell)}_K\mbf{S}^{(\ell-1)}\mright)^\top/ \sqrt{\embdim}}\mleft(\mbf{W}^{(\ell)}_V\mbf{S}^{(\ell-1)}\mright) \rVert_{1, 1}                                                          \\
                                                 & \leq \frac{\embdim}{H} \cdot \lVert \softmax{\mbf{W}^{(\ell)}_Q\mbf{S}^{(\ell)}\mleft( \mbf{W}^{(\ell)}_K\mbf{S}^{(\ell-1)}\mright)^\top/ \sqrt{\embdim}} \rVert_{\infty}\cdot \lVert \mleft(\mbf{W}^{(\ell)}_V\mbf{S}^{(\ell-1)}\mright) \rVert_{1, 1} \\
                                                 & \leq \frac{\embdim}{H}  \cdot \lVert \mleft(\mbf{W}^{(\ell)}_V\mbf{S}^{(\ell-1)}\mright) \rVert_{1, 1} \tag{the softmax matrix is row-stochastic}                                                                                                       \\
                                                 & \leq \frac{\embdim}{H}  \cdot \frac{\embdim}{H} \lVert (\mbf{W}^{(\ell)}_V \rVert_{\infty} \lVert \mbf{S}^{(\ell-1)}\rVert_{1, 1}          \leq \mleft(\frac{\embdim}{H}\mright)^2 B_V \lVert \mbf{S}^{(\ell-1)}\rVert_{1, 1}. \tag{definition of $\tilde{\paramspace}$}
    \end{align*}
    Combining the previous inequality with \cref{eq:softmax_head} leads to
    \begin{equation*}
        \lVert \mbf{C}^{(\ell)} \rVert_{1, 1} \leq \frac{\embdim^2}{H} B_V \lVert \mbf{S}^{(\ell-1)}\rVert_{1, 1}.
    \end{equation*}
    In the end, the multi-head attention norm verifies
    \begin{equation*}
        \lVert \mcal{A}\mleft(\mbf{S}^{(\ell-1)}; \mbf{W}_Q^{(\ell)}, \mbf{W}_K^{(\ell)}, \mbf{W}_V^{(\ell)}, \mbf{W}_O^{(\ell)} \mright) \rVert_{1, 1} \leq \frac{\embdim^3}{H} B_OB_V \lVert \mbf{S}^{(\ell-1)}\rVert_{1, 1}.
    \end{equation*}
    Using the triangular inequality, we obtain
    \begin{equation}
        \label{eq:norm_11_z}
        \lVert \mbf{Z}^{(\ell)} \rVert_{1, 1} \leq \mleft( 1 + \frac{\embdim^3}{H} B_OB_V \mright) \cdot \lVert \mbf{S}^{(\ell-1)}\rVert_{1, 1}.
    \end{equation}

    \textbf{Step 2: FF.} We recall that $\mbf{W}_1 \in \RR^{\hiddendim \times \embdim}$ and $\mbf{W}_2 \in \RR^{\embdim \times \hiddendim}$. Using similar arguments to the above, we have
    \begin{align*}
        \lVert \mbf{Y}^{(\ell)} \rVert_{1, 1} & = \lVert \mbf{W}_2^{(\ell)} \relu{\mbf{W}_1^{(\ell)}\mbf{Z}^{(\ell)}} \rVert_{1, 1}                                                                                                     \\
                                              & \leq \embdim \cdot \lVert \mbf{W}_2^{(\ell)} \rVert_{\infty} \lVert \relu{\mbf{W}_1^{(\ell)}\mbf{Z}^{(\ell)}} \rVert_{1, 1} \tag{\cref{lem:prop_l_11}}                                  \\
                                              & \leq \embdim \cdot \lVert \mbf{W}_2^{(\ell)} \rVert_{\infty} \lVert \mbf{W}_1^{(\ell)}\mbf{Z}^{(\ell)} \rVert_{1, 1} \tag{\cref{lem:ineq_relu}}                                         \\
                                              & \leq \embdim \cdot \hiddendim \cdot \lVert \mbf{W}_2^{(\ell)} \rVert_{\infty} \lVert \mbf{W}_1^{(\ell)} \rVert_{\infty} \lVert\mbf{Z}^{(\ell)} \rVert_{1, 1} \tag{\cref{lem:prop_l_11}} \\
                                              & \leq \embdim \hiddendim B_1B_2 \lVert\mbf{Z}^{(\ell)} \rVert_{1, 1} \tag{definition of $\tilde{\paramspace}$}.
    \end{align*}

    \textbf{Step 3: output layer.} Again, applying the triangular inequality and using the previous inequality and \cref{eq:norm_11_z}, we have
    \begin{align*}
        \lVert \mbf{S}^{(\ell)} \rVert_{1, 1} & \leq \lVert \mbf{Z}^{(\ell)} \rVert_{1, 1} + \lVert \mbf{Y}^{(\ell)} \rVert_{1, 1} \leq \mleft( 1+ \embdim \hiddendim B_1B_2\mright)  \lVert\mbf{Z}^{(\ell)} \rVert_{1, 1}                                                          \\
                                              & \leq \mleft( 1+ \embdim \hiddendim B_1B_2\mright)  \mleft( 1 + \frac{\embdim^3}{H} B_OB_V \mright)  \lVert \mbf{S}^{(\ell-1)}\rVert_{1, 1}.
    \end{align*}
    Iterating through the layers, recalling that $\mbf{S}^{(0)} = \mbf{S}$, we finally obtain
    \begin{equation*}
        \lVert \mbf{S}^{(L)} \rVert_{1, 1} \leq \mleft[\mleft( 1+ \embdim \hiddendim B_1B_2\mright) \mleft( 1 + \frac{\embdim^3}{H} B_OB_V \mright)\mright]^L \lVert \mbf{S}\rVert_{1, 1},
    \end{equation*}
    where $\mbf{S}$ is the input sequence. Combining this inequality with \cref{eq:ineq_output_depth_step_0} leads to
    \begin{equation*}
        \lVert \frac{1}{\inputsize\tau}\mbf{W}_U \mbf{S}^{(L)} \bbm{1}_\inputsize \rVert_1 \leq \mleft[\mleft( 1+ \embdim \hiddendim B_1B_2\mright)  \mleft( 1 + \frac{\embdim^3}{H} B_OB_V \mright)\mright]^L  \frac{\lVert \mbf{S}\rVert_{1, 1}}{\inputsize}  \mleft(\frac{1}{\tau} \lVert \mbf{W}_U^\top \rVert_{2, 1}\mright).
    \end{equation*}
    Using the fact that each token has a $\ell_1$-norm bounded by $\ubtok$. Hence, each column of $\mbf{S}$ is too and we have
    \begin{equation*}
        \frac{1}{n}\lVert \mbf{S}\rVert_{1, 1} = \frac{1}{n}\sum_{j=1}^{\inputsize} \sum_{i=1}^{\embdim} \rvert\mbf{S}_{ij}\lvert = \frac{1}{n}\sum_{j=1}^{\inputsize} \underbrace{\lVert \mbf{S}_{\cdot, j} \rVert_1}_{\leq \ubtok} \leq \ubtok.
    \end{equation*}
    Combining the last two inequalities concludes the proof.
\end{proof}

We can now restate \cref{cor:pre_training_risk_bound_llm_depth}.
\begin{boxcor}[Restatement of \cref{cor:pre_training_risk_bound_llm_depth}]
    Consider an LLM $\model_{\params} \in \tilde{\funcspace}$ with vocabulary size $\vocabsize$ composed of $L$ transformer blocks and $H$ attention heads. We denote by $\mixmat$ the mixing matrix of the pretraining sequences of tokens $\MCseq{S}{N_{\train}}$. Let $\delta > 0$. Then, with probability at least $1-\delta$,
    \begin{equation*}
        \risktrain{\params} \leq \smallrisktrain{\params} +  \frac{\bar{B}}{\sqrt{N_{\train}}}\sqrt{\log{\mleft(\frac{2}{\delta}\mright)}},
    \end{equation*}
    where $\bar{B}$ is a constant depending on the parameters of the problem. More precisely,
    \begin{equation*}
        \bar{B} = 2\lVert \mixmat \rVert\sqrt{\max\{\log{(\vocabsize)} +\frac{2(\ubmod)^L}{\tau}, \log{\mleft(\frac{1}{c_0}\mright)}\}},
    \end{equation*}
    with $\ubmod = \mleft[\mleft( 1+ \embdim \hiddendim B_1B_2\mright)  \mleft( 1 + \frac{\embdim^3}{H} B_OB_V \mright)\mright]\mleft( \ubtok B_U\mright)^{1/L}$.
\end{boxcor}
\begin{proof}[Proof of \cref{cor:pre_training_risk_bound_llm_depth}]
    We first note that the only change from \cref{lem:ineq_last_layer_output} to \cref{lem:ineq_last_layer_output_depth} is the multiplicative constant $c_3 = \mleft[\mleft( 1+ \embdim \hiddendim B_1B_2\mright)  \mleft( 1 + \frac{\embdim^3}{H} B_OB_V \mright)\mright]^L\ubtok$ in front of $\frac{1}{\tau}\lVert \mbf{W}_U^\top \rVert_{2, 1}$. In particular, as we know that $\tilde{\paramspace} \subset \paramspace$, we also have
    $\lVert \mbf{W}_U^\top \rVert_{2, 1} \leq B_U$. Hence, we can apply the proof of \cref{thm:pre_training_risk_bound_llm} in a straightforward manner by changing $\frac{B_U}{\tau}$ by $c_3 \cdot \frac{B_U}{\tau}$. This concludes the proof.
\end{proof}

\subsection{Proof of \cref{thm:risk_bound_llm}}
\label{app:risk_bound_llm}
In this section, we detail the proof of \cref{thm:risk_bound_llm}. We first recall the problem setup.

\paragraph{Markov chains inputs.} In this section, we give as input of the model a single Markov chain $X = \MCseq{X}{N_{\ICL}}$ with finite, discrete state space $\polish$ of size $\mcsize$ with transition probability $\PP$. We assume the $\mbf{X}_n$ are already tokenized and thus we have $\polish \subset \vocabspace$. We denote the sequence of tokens the LLM receives by $\mbf{S}_n = \MCseq{X}{n}$ if $n \leq \cxtsize$ and  $\mbf{S}_n = \mleft(\mbf{X}_{n-K+1}, \ldots, \mbf{X}_n \mright)$ otherwise due to the deletion process (see~\cref{def:deletion_process}). In particular, the $\mbf{S}_n$ are elements of $\vocabspace^*_{\cxtsize}$. We note that $S = \MCseq{S}{N_{\ICL}}$ is also a Markov chain (see \cref{app:connection_X_S}). By definition of $\PP$, we know that for any $n \in [N_{\ICL}]$, the next token $\mbf{X}_{n+1}$ follows the distribution $\Pb{\cdot}{\mbf{S}_n}$. We assume that there exists a positive constant $\pmin$ that lower bounds all the transition probability between states, i.e., $\forall n \in [N_{\ICL}], \forall x, y \in \polish, \quad \Pb{\mbf{X}_{n+1} = y}{\mbf{X}_n = x} \geq \pmin > 0$. This is akin to the ambiguity of language constant $c_0$ considered in the previous section and in~\citet{zhang2023whathowicl, hu2024statscot, xie2022iclbayesian, wies2024learnability}.

\paragraph{Next token probability distribution.} An important difference with the setting considered in \cref{thm:pre_training_risk_bound_llm} is that here, we predict a probability distribution on the state space $\polish$ of the Markov chain and not on the vocabulary of the LLM $\vocabspace$. To that end, we restrict the predicted probability given the past tokens $\mbf{S}_n$ to the state space $\polish$. Formally, denoting the output of the last layer of $\model_{\params}$ by $\mbf{S}^{(L)}$, the last layer before the softmax outputs a vector $\mbf{u} = \frac{1}{n\tau} \mbf{W}_U \mbf{S}^{(L)} \bbm{1}_n \in \RR^{\vocabsize}$. We first extract the entries of $\mbf{u}$ whose index $i$ are such that the $i$-th element of the vocabulary space $\vocabspace$ is in $\polish$. This can be formalized as follows. We denote by $\II_{\mcsize} = \subspace$ the subset of $\mcsize$ distinct elements of $[\vocabsize]$ and consider the matrix $\mbf{M}_j = \mbf{e}_{i_j}^\top$, where $\mbf{e}_{i_j} \in \RR^{T}$ has value $1$ at entry $i_j \in \II$ and $0$ elsewhere. Extracting only the $\mcsize$ entries of $\mbf{u}$ that corresponds to the state space yields a vector in $\RR^{\mcsize}$ that writes $\mbf{v} = \frac{1}{n\tau} \mbf{M} \mbf{W}_U \mbf{S}^{(L)} \bbm{1}_n \in \RR^{\vocabsize}$. Similarly to \cref{app:background_transformer}, the probability distribution of next token $\mbf{X}_{n+1}$ provided by the LLM $\model_{\params}$ now writes
\begin{equation*}
    \Pbemp{\params}{\cdot}{\mbf{S}_n} = \softmax{\frac{1}{n\tau} \mbf{M} \mbf{W}_U \mbf{S}^{(L)} \bbm{1}_n} \in \Delta_{\mcsize}.
\end{equation*}
We aim to obtain a similar generalization bound than in \cref{thm:pre_training_risk_bound_llm} where the reference probability distribution is the Markov chain transition probability $\PP$ instead of the probability distribution of language $\PP_{\mcal{L}}$. In particular, $\PP$ will replace $\PP_{\mcal{L}}$ in the definition of the risks in \cref{eq:estimation_error_def}. We provide below an overview of the proof before detailing it.
\paragraph{Overview of the proof.} We are going to use McDiarmid's inequality for Markov chains of~\citet[Corollary 2.11]{paulin2015ineqMC}. To adapt their arguments to our setting, we bound the total variation between the true probability of the next token and the one estimated by the LLM. The rest of this section is organized as follows. First, in \cref{app:connection_X_S}, we show that $S = \MCseq{S}{N_{\ICL}}$ is a Markov chain. Then in \cref{app:thm_icl_paulin}, we adapt the concentration inequality of~\citet[Corollary 2.11]{paulin2015ineqMC}. Afterwards in \cref{app:thm_icl_bound}, we show how to bound the total variation between the true and the estimated probability of the next token. Finally \cref{app:thm_icl_conclude} concludes the proof.

\subsubsection{Connection between tokens and sequences of tokens Markov chains}
\label{app:connection_X_S}
We first show that $S = \MCseq{S}{N_{\ICL}}$ is also a Markov chain.

\begin{boxlem}
Consider a sequence (not necessarily a Markov chain) $X = \MCseq{X}{N}$ with values in $\polish$ and let $\mbf{S}_n = \MCseq{X}{n}$ if $n < \cxtsize$ and $\mbf{S}_n = \mleft(\mbf{X}_{n-K+1}, \ldots, \mbf{X}_n \mright)$ otherwise. Then, the sequence $S = \MCseq{S}{N}$ is a Markov chain with state space $\polish^*_{\cxtsize}$ that contains the sequence of elements in $\polish$ of length smaller than $\cxtsize$. 
\end{boxlem}

\begin{proof}
By definition of the $\mbf{S}_n$, we know that they take values in $\polish^*_K$. Let $x_{1}, \ldots, x_{n+1} \in \polish$. We first assume that $n > \cxtsize$ and denote $s_i = (x_{n-\cxtsize+1}, \ldots, x_i)$.
 We have
 \begin{align*}
     &\Pb{\mbf{S}_{n+1} = s_{n+1}}{\mbf{S}_n=s_n, \ldots, \mbf{S}_{n-\cxtsize+1} = s_{n-\cxtsize+1}} \\
     & = \Pb{\mbf{S}_{n+1} = s_{n+1}}{\mbf{X}_n = x_n, \ldots, \mbf{X}_{n-\cxtsize+1} = x_{n-\cxtsize+1}} \\
     & = \Pb{\mbf{S}_{n+1} = s_{n+1}}{\mbf{S}_n = s_n} \tag{by definition of $\mbf{S}_n$}.
 \end{align*}
Similarly, we assume $n < \cxtsize$ and denote $s_i = (x_{1}, \ldots, x_i)$. We have
 \begin{align*}
     &\Pb{\mbf{S}_{n+1} = s_{n+1}}{\mbf{S}_n=s_n, \ldots, \mbf{S}_{1} = s_{1}} \\
     & = \Pb{\mbf{S}_{n+1} = s_{n+1}}{\mbf{X}_n = x_n, \ldots, \mbf{X}_{1} = x_{1}} \\
     & = \Pb{\mbf{S}_{n+1} = s_{n+1}}{\mbf{S}_n = s_n} \tag{by definition of $\mbf{S}_n$}.
 \end{align*}
Finally, for $n=\cxtsize$, we denote $s_i = (x_{1}, \ldots, x_i)$ for $i \leq \cxtsize$ and $s_{\cxtsize+1} = (x_{2}, \ldots, x_{\cxtsize+1})$. We have
 \begin{align*}
     \Pb{\mbf{S}_{\cxtsize+1} = s_{\cxtsize+1}}{\mbf{S}_n=s_n, \ldots, \mbf{S}_{2} = s_{2}} \\
     &\qquad= \Pb{\mbf{S}_{\cxtsize+1} = s_{\cxtsize+1}}{\mbf{X}_{\cxtsize} = x_{\cxtsize}, \ldots, \mbf{X}_{1} = x_{1}} \\
     & \qquad= \Pb{\mbf{S}_{\cxtsize+1} = s_{\cxtsize+1}}{\mbf{S}_{\cxtsize} = s_{\cxtsize}} \tag{by definition of $\mbf{S}_{\cxtsize}$}.
 \end{align*}
This establishes the Markov property for $\mathbf{S}$.
\end{proof}

\subsubsection{Concentration Inequalities for Markov Chains}
\label{app:thm_icl_paulin}
We first state a concentration inequality for time-homogeneous Markov chains that will be used to obtain our final bound.

\begin{boxprop}[McDiarmid's inequality for time-homogeneous Markov chains]
    \label{prop:mcdiarmid_ineq_mc}
    Let $S \coloneqq \MCseq{S}{N}$ be a Markov chain with value in a discrete, finite state space $\polish$ and mixing time $\tmix{\varepsilon}$. Let
    $t_\mrm{min} \coloneqq \inf_{0 \leq \varepsilon < 1} \tmix{\frac{\varepsilon}{2}} \cdot \mleft( \frac{2 - \varepsilon}{1-\varepsilon}\mright)^2$.
    If $f \colon \polish \to \RR$ is such that there exists $\mbf{c} \in \RR^N$ satisfying
    \begin{equation*}
        \forall \mbf{x}, \mbf{y} \in \polish, \quad f(\mbf{x}) - f(\mbf{y}) \leq \sum_{i=1}^N \mbf{c}_i \bbm{1}_{\{\mbf{x}_i \neq \mbf{y}_i\}},
    \end{equation*}
    then we have for any $u \geq 0$,
    \begin{equation*}
        \PP \mleft( \lvert f(S) - \EE_S \mleft[ f(S)\mright] \rvert \geq u \mright) \leq 2\exp{\mleft( \frac{-2u^2}{\lVert \mbf{c}\rVert_2^2 \cdot t_\mrm{min}}\mright)}.
    \end{equation*}
\end{boxprop}

\begin{proof}
    We recall that Corollary 2.11 of~\citet{paulin2015ineqMC} ensures that for such a function $f$, we have
    \begin{equation}
        \label{eq:ineq_mc_icl}
        \PP \mleft( \lvert f(S) - \EE \mleft[ f(S)\mright] \rvert \geq u \mright) \leq 2\exp{\mleft( \frac{-2u^2}{\lVert \mbf{c}\rVert_2^2 \cdot \tau_\mrm{min}}\mright)},
    \end{equation}
    where $\tau_\mrm{min}$ is defined as
    \begin{equation*}
        \tau_\mrm{min} \coloneqq \inf_{0 \leq \varepsilon < 1} \tau(\varepsilon)  \mleft( \frac{2 - \varepsilon}{1-\varepsilon}\mright)^2,
    \end{equation*}
    with $\tau(\varepsilon)$ being the mixing time of a Markov chain \emph{without assuming time homogeneity} (see~\citet[Definition 1.4]{paulin2015ineqMC}). As in our case, we assume the time homogeneity, this inequality in \cref{eq:ineq_mc_icl} has to be adapted.  Following Remark 1.5 of~\citet{paulin2015ineqMC}, we notice that
    \begin{equation*}
        \forall \varepsilon \in [0, 1], \quad \tau(2\varepsilon) \leq \tmix{\varepsilon} \leq \tau(\varepsilon).
    \end{equation*}
    Let $ 0 \leq \varepsilon < 1$. Using the fact that $\mleft(\frac{2 - \varepsilon}{1-\varepsilon}\mright)^2 > 0$, the previous inequality ensures
    \begin{align*}
        \tau(\varepsilon) \leq \tmix{\frac{\varepsilon}{2}} & \iff \tau(\varepsilon)  \mleft( \frac{2 - \varepsilon}{1-\varepsilon}\mright)^2 \leq \tmix{\frac{\varepsilon}{2}}  \mleft( \frac{2 - \varepsilon}{1-\varepsilon}\mright)^2.
    \end{align*}
    Taking the infimum on the left-hand side leads to
    \begin{equation*}
        \tau_\mrm{min} = \inf_{0 \leq \varepsilon < 1} \tau(\varepsilon)  \mleft( \frac{2 - \varepsilon}{1-\varepsilon}\mright)^2 \leq \tmix{\frac{\varepsilon}{2}}  \mleft( \frac{2 - \varepsilon}{1-\varepsilon}\mright)^2.
    \end{equation*}
    As we took $\varepsilon$ arbitrary in $[0, 1)$, we can take the infimum on the right-hand side, which leads to
    \begin{equation*}
        \tau_\mrm{min} \leq t_\mrm{min}.
    \end{equation*}
    As the function $x \to \exp{\mleft( \frac{-2u^2}{\lVert \mbf{c}\rVert_2^2  x}\mright)}$ is decreasing, we finally obtain
    \begin{equation}
        \label{eq:bound_ineq_icl}
        \exp{\mleft( \frac{-2u^2}{\lVert \mbf{c}\rVert_2^2  \tau_\mrm{min}}\mright)} \leq \exp{\mleft( \frac{-2u^2}{\lVert \mbf{c}\rVert_2^2  t_\mrm{min}}\mright)}.
    \end{equation}
    Combining \cref{eq:ineq_mc_icl,eq:bound_ineq_icl} concludes the proof.
\end{proof}
Similarly to \cref{thm:pre_training_risk_bound_llm}, we want to apply \cref{prop:mcdiarmid_ineq_mc} to a function $f$ that consists of sums of total variation. We investigate in the next section how to find the bounding vector $\mbf{c}$ to apply \cref{prop:mcdiarmid_ineq_mc}.

\subsubsection{Finding the Bounding Vector}
\label{app:thm_icl_bound}
We want to apply the same arguments as in the proof of \cref{thm:pre_training_risk_bound_llm} to find the bounding vector $\mbf{c}$. The only difference in terms of setting is the definition of the probability of the next token. Indeed, in our case, we apply an extraction matrix $\mbf{M} \in \RR^{\mcsize \times \vocabsize}$ to recover the $\mcsize$ states of the input Markov chain. We first prove the following technical lemma.

\begin{boxlem}
    \label{lem:contraction_ineq}
    Let $\mcsize \leq \vocabsize$ and consider a subset of $\mcsize$ \emph{distinct} elements of $[\vocabsize]$ that writes $\II_{\mcsize} = \subspace$. We denote by $\mbf{M} \in \RR^{\mcsize \times \vocabsize}$ the matrix with rows $\mbf{M}_j = \mbf{e}_{i_j}^\top$, where $\mbf{e}_{i_j} \in \RR^{T}$ has value $1$ at entry $i_j \in \II$ and $0$ elsewhere. For any vector $\mbf{u} \in \RR^{\vocabsize}$, we have
    \begin{equation*}
        \lVert \mbf{M} \mbf{u}\rVert_1 \leq \lVert\mbf{u} \rVert_1.
    \end{equation*}
\end{boxlem}
\begin{proof}
    By definition of the $\ell_1$-norm, we have
    \[
        \lVert \mbf{M} \mbf{u}\rVert_1  = \sum_{k=1}^{\mcsize}  \lvert \sum_{l=1}^{\vocabsize} \mbf{M}_{kl}\mbf{u}_l\rvert
        \leq \sum_{k=1}^{\mcsize}  \sum_{l=1}^{\vocabsize} \lvert \mbf{M}_{kl}\mbf{u}_l\rvert
        \leq \sum_{l=1}^{\vocabsize} \lvert \mbf{u}_l\rvert \sum_{k=1}^{\mcsize} \lvert\mbf{M}_{kl}\rvert.
    \]
    Moreover, each column of $\mbf{M}$ contains at most one non-zero entry (with value $1$). Otherwise, it means that two $\mbf{e}_{i_j}$ are identical (as they only have one non-zero entry with value $1$, having it at the same position ensures their equality) which contradicts the fact that the $i_j$ where taken distinct. Hence, for all $l$, we have $\sum_{k=1}^{\mcsize} \lvert\mbf{M}_{kl}\rvert \leq 1$, which concludes the proof.
\end{proof}
We now prove a lemma analogous to \cref{lem:ineq_last_layer_output}.
\begin{boxlem}
    \label{lem:ineq_last_layer_output_icl}
    Let $\mbf{S} \in \RR^{\embdim \times \inputsize}$ denote the entry of the LLM $\model_{\params}$ and $\mbf{S}^{(L)}$ denote the output of the last layer before the softmax. Let $\mcsize \leq \vocabsize$ and consider a subset of $\mcsize$ \emph{distinct} elements of $[\vocabsize]$ that writes $\II_{\mcsize} = \subspace$. We denote by $\mbf{M} \in \RR^{\mcsize \times \vocabsize}$ the matrix with rows $\mbf{M}_j = \mbf{e}_{i_j}^\top$, where $\mbf{e}_{i_j} \in \RR^{\vocabsize}$ has value $1$ at entry $i_j \in \II$ and $0$ elsewhere. Then, the following inequality holds
    \begin{equation*}
        \frac{1}{\inputsize\tau}\lVert \mbf{M}\mbf{W}_U \mbf{S}^{(L)} \bbm{1}_\inputsize \rVert_1 \leq \frac{1}{\tau} \lVert \mbf{W}_U^\top \rVert_{2, 1}.
    \end{equation*}
\end{boxlem}
\begin{proof}
    Applying \cref{lem:contraction_ineq} with the matrix $\mbf{M} \in \RR^{\mcsize}$ and the vector $\frac{1}{\inputsize\tau}\mbf{W}_U \mbf{X}^{(L)} \bbm{1}_{\inputsize} \in \RR^{\vocabsize}$ leads to
    \begin{equation*}
        \frac{1}{\inputsize\tau}\lVert\mbf{M}\mbf{W}_U \mbf{S}^{(L)} \bbm{1}_{\inputsize} \rVert_1 \leq  \frac{1}{\inputsize\tau}\lVert \mbf{W}_U \mbf{X}^{(L)} \bbm{1}_{\inputsize} \rVert_1.
    \end{equation*}
    Applying \cref{lem:ineq_last_layer_output} concludes the proof.
\end{proof}

The previous lemma can be used to show that the logarithm of the ratio between the true probability of the next token and the one estimated by the LLM $\model_{\params}$ is upper bounded as a function of the number of states of the Markov chain $\mcsize$, the temperature $\tau$, the upper-bound on $\mbf{W}_U$ and some constant related to the ambiguity of language (see \cref{eq:ambiguity_language}).

\begin{boxprop}[Upper-bound on the logarithm]
    \label{prop:ub_ratio_log_icl}
    Consider an LLM $\model_{\params} \in \funcspace$ and an input Markov chain $X = \MCseq{X}{N_{\ICL}}$ with $d$ states. We recall that $B_U$ is the upper bound on the norm of $\mbf{W}_U$ in the definition of parameter space $\paramspace$, $\tau$ is the softmax temperature, and $\pmin$ is the constant related to the minimal transition probability between states. We have
    \begin{equation*}
        \forall n \in [N], \quad \left|\log{\mleft ( \frac{\Pb{\mbf{X}_{n+1}}{\mbf{S}_n}}{\Pbemp{\params}{\mbf{X}_{n+1}}{\mbf{S}_n}}\mright)}\right| \leq \cst = \max\{\log{(\mcsize)} + \frac{2B_U}{\tau}, \log{\mleft(\frac{1}{\pmin}\mright)}\}.
    \end{equation*}
\end{boxprop}
\begin{proof}
    The main idea of the proof is to bound the probability ratio and use the non-decreasing monotonicity of the $\log$. Let $n \in [N]$. The model $\model_{\params}$ receives as input sequences of tokens $\mbf{S}_n$ of size $n \leq \cxtsize$. We first lower-bound each term of the probability ratio. By definition of $\pmin$, we have
    \begin{equation}
        \label{eq:ineq_pb_icl}
        \Pb{\mbf{X}_{n+1}}{\mbf{S}_n} = \Pb{\mbf{X}_{n+1}}{\mbf{X}_n} \geq \pmin > 0,
    \end{equation}
    where we used the Markov property for the first equality.
    We want to obtain a similar inequality for $\Pbemp{\params}{\mbf{X}_{n+1}}{\mbf{S}_n}$. As the parameters $\params$ of the LLM are in $\paramspace$, we know that $\lVert \mbf{W}_U^\top \rVert_{2, 1} \leq B_U$. \cref{lem:ineq_last_layer_output_icl} ensures that
    \begin{equation*}
        \lVert \frac{1}{\inputsize\tau}\mbf{M}\mbf{W}_U \mbf{S}^{(L)} \bbm{1}_T \rVert_1 \leq \frac{1}{\tau} \lVert \mbf{W}_U^\top \rVert_{2, 1} \leq \frac{B_U}{\tau}.
    \end{equation*}
    We can then apply \cref{lem:ineq_softmax} with $c_1= \frac{B_U}{\tau}$ and given that $\frac{1}{T\tau}\mbf{M}\mbf{W}_U \mbf{S}^{(L)} \bbm{1}_T \in \RR^{\mcsize}$, it leads to
    \begin{equation*}
        \Pbemp{\params}{\cdot}{\mbf{S}_n} = \softmax{\frac{1}{\inputsize\tau}\mbf{M}\mbf{W}_U \mbf{S}^{(L)} \bbm{1}_{\inputsize}} \geq \frac{1}{\mcsize \exp{(2B_U/\tau)}},
    \end{equation*}
    where the inequality holds for each component of $\Pbemp{\params}{\cdot}{\mbf{S}_n}$. This is in particular the case $\Pbemp{\params}{\mbf{X}_{n+1}}{\mbf{S}_n}$ which is the entry we are interested in, i.e., we have
    \begin{equation}
        \label{eq:ineq_pbemp_icl}
        \Pbemp{\params}{\mbf{X}_{n+1}}{\mbf{S}_n} \geq \frac{1}{\mcsize \exp{(2B_U/\tau)}}.
    \end{equation}
    Going back to the ratio of probability, consider the situation where we have
    \begin{equation*}
        \frac{\Pb{\mbf{X}_{n+1}}{\mbf{S}_n}}{\Pbemp{\params}{\mbf{X}_{n+1}}{\mbf{S}_n}} \geq 1.
    \end{equation*}
    Then, using Eq.~\eqref{eq:ineq_pbemp_icl}, we have
    \begin{equation*}
        1 \leq \frac{\Pb{\mbf{X}_{n+1}}{\mbf{S}_n}}{\Pbemp{\params}{\mbf{X}_{n+1}}{\mbf{S}_n}} \leq \frac{1}{\Pbemp{\params}{\mbf{X}_{n+1}}{\mbf{S}_n}} \leq \mcsize \exp{(2B_U/\tau)},
    \end{equation*}
    which implies, as the $\log$ is non-decreasing monotonically,
    \begin{equation}
        \label{eq:ineq_log_rat_pos_icl}
        0 \leq \log{\mleft( \frac{\Pb{\mbf{X}_{n+1}}{\mbf{S}_n}}{\Pbemp{\params}{\mbf{X}_{n+1}}{\mbf{S}_n}} \mright)} \leq \log{(\mcsize \exp{(2B_U/\tau)})} = \log{(\mcsize)} + \frac{2B_U}{\tau}.
    \end{equation}
    Similarly, consider the case where we have
    \begin{equation*}
        \frac{\Pb{\mbf{X}_{n+1}}{\mbf{S}_n}}{\Pbemp{\params}{\mbf{X}_{n+1}}{\mbf{S}_n}} \leq 1.
    \end{equation*}
    Then, we have
    \begin{equation*}
        \frac{\Pbemp{\params}{\mbf{X}_{n+1}}{\mbf{S}_n}}{\Pb{\mbf{X}_{n+1}}{\mbf{S}_n}} \geq 1,
    \end{equation*}
    and similarly to above, we can use Eq.~\eqref{eq:ineq_pb_icl} to obtain
    \begin{equation*}
        1 \leq \frac{\Pbemp{\params}{\mbf{X}_{n+1}}{\mbf{S}_n}}{\Pb{\mbf{X}_{n+1}}{\mbf{S}_n}} \leq \frac{1}{\Pb{\mbf{X}_{n+1}}{\mbf{S}_n}} \leq \frac{1}{\pmin}.
    \end{equation*}
    This implies
    \begin{equation*}
        0 \leq \log{\mleft( \frac{\Pbemp{\params}{\mbf{X}_{n+1}}{\mbf{S}_n}}{\Pb{\mbf{X}_{n+1}}{\mbf{S}_n}} \mright)} \leq \log{\mleft(\frac{1}{\pmin}\mright)},
    \end{equation*}
    which also rewrites
    \begin{equation}
        \label{eq:ineq_log_rat_neg_icl}
        0 \leq - \log{\mleft( \frac{\Pb{\mbf{X}_{n+1}}{\mbf{S}_n}}{\Pbemp{\params}{\mbf{X}_{n+1}}{\mbf{S}_n}} \mright)} \leq \log{\mleft(\frac{1}{\pmin}\mright)}.
    \end{equation}
    By definition of the absolute value, combining Eq.~\eqref{eq:ineq_log_rat_pos_icl} and Eq.~\eqref{eq:ineq_log_rat_neg_icl} leads to
    \begin{equation*}
        \left | \log{\mleft( \frac{\Pb{\mbf{X}_{n+1}}{\mbf{S}_n}}{\Pbemp{\params}{\mbf{X}_{n+1}}{\mbf{S}_n}} \mright)} \right | \leq \max\{\log{(\mcsize)} + \frac{2B_U}{\tau}, \log{\mleft(\frac{1}{\pmin}\mright)}\}.
    \end{equation*}
    This concludes the proof.
\end{proof}

We are now ready to upper-bound the total variation.
\begin{boxcor}[Upper-bound on the total variation]
    \label{cor:ub_tv_icl}
    Consider an LLM $\model_{\params} \in \funcspace$ and an input Markov chain $X = \MCseq{X}{N_{\ICL}}$ with $d$ states. We recall that $B_U$ is the upper bound on the norm of $\mbf{W}_U$ in the definition of parameter space $\paramspace$, $\tau$ is the softmax temperature, and $\pmin$ is the constant related to the minimal transition probability between states. We have
    \begin{equation}
        \label{eq:ub_tv}
        \forall n \in [N], \quad \TV{\Pb{\cdot}{\mbf{S}_n}}{\Pbemp{\params}{\cdot}{\mbf{S}_n}} \leq \sqrt{2\max\{\log{(\mcsize)} + \frac{2B_U}{\tau}, \log{\mleft(\frac{1}{\pmin}\mright)}\}} \coloneqq c_4.
    \end{equation}
\end{boxcor}
\begin{proof}
    Using \cref{prop:ub_ratio_log_icl}, we can directly apply \cref{lem:ineq_tv_hellinger} with $B = \max\{\log{(\mcsize)} + \frac{2B_U}{\tau}, \log{\mleft(\frac{1}{\pmin}\mright)}\}$ for any $n \in [N]$. It leads to
    \begin{equation*}
        \forall n \in [N], \quad \TV{\Pb{\cdot}{\mbf{S}_n}}{\Pbemp{\params}{\cdot}{\mbf{S}_n}} \leq \sqrt{2 \max\{\log{(\mcsize)} + \frac{2B_U}{\tau}, \log{\mleft(\frac{1}{\pmin}\mright)}\}}.
    \end{equation*}
    This concludes the proof.
\end{proof}

\subsubsection{Concluding the Proof}
\label{app:thm_icl_conclude}
We are now ready to state our main result.

\begin{boxthm}[Restatement of \cref{thm:risk_bound_llm}]
    \label{thm:restate_icl_risk_bound_llm}
    Consider an LLM $\model_{\params} \in \funcspace$. We provide as input of $\model_{\params}$ a $\mcsize-$state Markov chain $X = \MCseq{X}{N_{\ICL}}$. The sequence of subsequences of the first $n$ terms is denoted by $S = \MCseq{S}{N_{\ICL}}$. $S$ is also a Markov chain, and we denote by $t_{\mrm{mix}}(\varepsilon)$ its mixing time. Let $t_\mrm{min} \coloneqq \inf_{0 \leq \varepsilon < 1} \tmix{\frac{\varepsilon}{2}} \cdot \mleft( \frac{2 - \varepsilon}{1-\varepsilon}\mright)^2$. Let $\delta > 0$. Then, with probability at least $1-\delta$,
    \begin{equation*}
        \riskicl{\params} \leq \inf_{\mcparams \in \mcparamspace} \{ \smallriskicl{\mcparams} + K(\mcparams, \params)\} + \bar{B}\sqrt{\frac{t_{\mrm{min}}}{N_{\ICL}}}\sqrt{\log{\mleft(\frac{2}{\delta}\mright)}},
    \end{equation*}
    where $\bar{B}$ is a constant depending on the parameters of the problem. More precisely,
    \begin{equation*}
        \bar{B} = 2\sqrt{\max\{\log{(\mcsize)} + \frac{2B_U}{\tau}, \log{\mleft(\frac{1}{\pmin}\mright)}\}}.
    \end{equation*}
\end{boxthm}
\begin{proof}
    Let $\mcparams \in \mcparamspace$. We first benefit from the metric properties of the total variation to decompose the risk.

    \begin{align}
        \label{eq:decompotision}
        \riskicl{\params} & = \frac{1}{N_{\ICL}}\sum_{n=1}^{N_{\ICL}} \Exp{\mbf{S}_n}{\TV{\Pb{\cdot}{\mbf{S}_n}}{\Pbemp{\params}{\cdot}{\mbf{S}_n}}} \notag                                                                                                                             \\
                          & \leq \frac{1}{N_{\ICL}}\sum_{n=1}^{N_{\ICL}} \Exp{\mbf{S}_n}{\TV{\Pb{\cdot}{\mbf{S}_n}}{\Pbemp{\mcparams}{\cdot}{\mbf{S}_n}} + \TV{\Pbemp{\mcparams}{\cdot}{\mbf{S}_n}}{\Pbemp{\params}{\cdot}{\mbf{S}_n}}}                                          \notag \\
                          & \leq \frac{1}{N_{\ICL}}\sum_{n=1}^{N_{\ICL}} \Exp{\mbf{S}_n}{\TV{\Pb{\cdot}{\mbf{S}_n}}{\Pbemp{\mcparams}{\cdot}{\mbf{S}_n}}} \notag                                                                                                                        \\
                          & \qquad \qquad + \frac{1}{N_{\ICL}}\sum_{n=1}^{N_{\ICL}} \Exp{\mbf{S}_n}{\TV{\Pbemp{\mcparams}{\cdot}{\mbf{S}_n}}{\Pbemp{\params}{\cdot}{\mbf{S}_n}}} \notag                                                                                                 \\
                          & \leq \riskicl{\mcparams} + K(\mcparams, \params).
    \end{align}

    By definition of the risk, we have
    \[
        \smallriskicl{\mcparams}  = \frac{1}{N_{\ICL}} \sum_{n=1}^{N_{\ICL}} \underbrace{\TV{\Pb{\cdot}{\mbf{S}_n}}{\Pbemp{\mcparams}{\cdot}{\mbf{S}_n}}}_{=g_n\mleft(\mbf{S}_n\mright)} = \frac{1}{N_{\ICL}} \sum_{n=1}^{N_{\train}} g_n\mleft(\mbf{S}_n\mright)                                                                                = f\MCseq{S}{N_{\ICL}}                                                                                                                                 = f\mleft(S\mright).
    \]
    Using \cref{cor:ub_tv_icl}, we know that
    \begin{equation*}
        \lvert g_n\mleft(\mbf{S}_n\mright) \rvert \leq \sqrt{2\max\{\log{(\mcsize)} + \frac{2B_U}{\tau}, \log{\mleft(\frac{1}{\pmin}\mright)}\}} \coloneqq c_4.
    \end{equation*}
    Similarly to \cref{thm:pre_training_risk_bound_llm}, and using the fact that $S = \MCseq{S}{N_{\ICL}}$ is a Markov chain, we can show that choosing $\mbf{c} \in \RR^{N_{\ICL}}$ with all entries equal to $\frac{2c_4}{N_{\ICL}}$ ensures that $f$ verifies the condition in \cref{prop:mcdiarmid_ineq_dependent_rv}, i.e.,
    \begin{equation*}
        \forall S, \Sigma, \quad f(S) - f(\Sigma) \leq \sum_{n=1}^{N_{\ICL}} \mbf{c}_n \bbm{1}_{\{\mbf{S}_n \neq \bm{\Sigma}_n\}}.
    \end{equation*}
    Putting everything together, we can apply \cref{prop:mcdiarmid_ineq_mc} which leads to
    \begin{equation}
        \label{eq:thm_pretraining_paulin_abs2}
        \forall u \geq 0, \quad \PP \mleft( \lvert f(S) - \EE_S \mleft[ f(S)\mright] \rvert \geq u \mright) \leq 2\exp{\mleft( \frac{-2u^2}{t_{\mrm{min}} \lVert \mbf{c}\rVert_2^2} \mright)}.
    \end{equation}
    Let $u \geq 0$. We have the following events ordering
    \begin{align*}
        \mleft( \EE_S \mleft[ f(S)\mright] - f(S) \geq u \mright) & \subseteq \mleft( \EE_S \mleft[ f(S)\mright] - f(S) \geq u \mright) \cup \mleft(f(S) - \EE_S \mleft[ f(S)\mright] \geq u \mright) \\
                                                                  & = \mleft(\left | f(S) - \EE_S \mleft[ f(S)\mright] \right | \geq u\mright).
    \end{align*}
    Hence, as $u$ was taken arbitrary and using \cref{eq:thm_pretraining_paulin_abs2}, we have
    \begin{equation*}
        \forall u \geq 0, \quad \PP \mleft( \EE_S \mleft[ f(S)\mright] - f(S) \geq u \mright) \leq 2\exp{\mleft( \frac{-2u^2}{t_{\mrm{min}} \lVert \mbf{c}\rVert_2^2} \mright)}.
    \end{equation*}
    We recall that by definition
    \begin{equation*}
        f(S) = \smallriskicl{\mcparams} \text{ and } \riskicl{\mcparams} = \EE_S \mleft[ \smallriskicl{\mcparams}\mright].
    \end{equation*}
    Moreover, the inequality on the probability holds for any $u \geq 0$, we can choose $u$ such that
    \begin{align*}
        \delta = 2\exp{\mleft( \frac{-2u^2}{t_{\mrm{min}} \mbf{c}\rVert_2^2} \mright)} & \iff \frac{-2u^2}{t_{\mrm{min}} \lVert \mbf{c}\rVert_2^2} = \log{\mleft( \frac{\delta}{2}\mright)} \iff u^2 = \frac{1}{2}t_{\mrm{min}} \lVert \mbf{c}\rVert_2^2 \log{\mleft( \frac{2}{\delta}\mright)} \\
        & \iff u = \frac{1}{\sqrt{2}}\sqrt{t_{\mrm{min}}} \lVert \mbf{c}\rVert_2 \sqrt{\log{\mleft( \frac{2}{\delta}\mright)}}.
    \end{align*}
    Using the fact that
    \[
        \lVert \mbf{c}\rVert_2 = \sqrt{\sum_{n=1}^{N_\ICL} \mbf{c}_n^2} = \sqrt{\sum_{n=1}^{N_\ICL} \mleft( \frac{2c_4}{N_{\ICL}}\mright)^2}= \sqrt{\sum_{n=1}^{N_\ICL} \frac{4c_4^2}{N_{\ICL}^2}}= \sqrt{\frac{4c_4^2}{N_{\ICL}}}= \frac{2c_4}{\sqrt{N_{\ICL}}}.\]
    Using the fact that $c_4 = \sqrt{2\max\{\log{(\mcsize)} + \frac{2B_U}{\tau}, \log{\mleft(\frac{1}{\pmin}\mright)}\}}$ (\cref{cor:ub_tv_icl}), we obtain
    \begin{align*}
        u & = \frac{1}{\sqrt{2}} \frac{2c_4}{\sqrt{N_{\ICL}}} \sqrt{t_{\mrm{min}}} \sqrt{\log{\mleft( \frac{2}{\delta}\mright)}}                                                                      = \frac{\sqrt{2}c_4}{\sqrt{N_{\ICL}}} \sqrt{t_{\mrm{min}}} \sqrt{\log{\mleft( \frac{2}{\delta}\mright)}}                                                                                  \\
          & = \frac{2\sqrt{t_{\mrm{min}}} \sqrt{\max\{\log{(\mcsize)} + \frac{2B_U}{\tau}, \log{\mleft(\frac{1}{\pmin}\mright)}\}}}{\sqrt{N_{\train}}}  \sqrt{\log{\mleft( \frac{2}{\delta}\mright)}} \\
          & = \bar{B}\sqrt{\frac{t_{\mrm{min}}}{N_{\ICL}}}\sqrt{\log{\mleft( \frac{2}{\delta}\mright)}},
    \end{align*}
    where we define
    \begin{equation*}
        \bar{B} = 2 \sqrt{\max\{\log{(\mcsize)} + \frac{2B_U}{\tau}, \log{\mleft(\frac{1}{\pmin}\mright)}\}}.
    \end{equation*}
    Putting everything together, we have
    \begin{equation*}
        \PP \mleft( \riskicl{\mcparams} - \smallriskicl{\mcparams} \geq \bar{B}\sqrt{\frac{t_{\mrm{min}}}{N_{\ICL}}}\sqrt{\log{\mleft( \frac{2}{\delta}\mright)}} \mright) \leq \delta.
    \end{equation*}
    Taking the opposite event leads to the following inequality with probability at least $1-\delta$
    \begin{equation*}
        \riskicl{\mcparams} \leq \smallriskicl{\mcparams} + \bar{B}\frac{\sqrt{t_{\mrm{min}}}}{\sqrt{N_{\ICL}}}\sqrt{\log{\mleft( \frac{2}{\delta}\mright)}}.
    \end{equation*}
    Going back to the decomposition of the risk in \cref{eq:decompotision} and rearranging the terms, we obtain
    \begin{equation*}
        \riskicl{\params} \leq \smallriskicl{\mcparams} + K(\params, \mcparams) + \bar{B}\frac{\sqrt{t_{\mrm{min}}}}{\sqrt{N_{\ICL}}}\sqrt{\log{\mleft(\frac{2}{\delta}\mright)}}.
    \end{equation*}
    As the left-hand side and the bound function of $\bar{B}$ do not depend on $\mcparams$, we can put them both on the left side of the inequality and then take the infimum on $\mcparams$. Rearranging the terms to keep only $\smallriskicl{\params}$ on the left side of the inequality leads to
    \begin{equation*}
        \riskicl{\params} \leq \inf_{\mcparams \in \mcparamspace} \{ \smallriskicl{\mcparams} + K(\mcparams, \params)\} + \bar{B}\sqrt{\frac{t_{\mrm{min}}}{N_{\ICL}}}\sqrt{\log{\mleft(\frac{2}{\delta}\mright)}},
    \end{equation*}
    which concludes the proof.
\end{proof}

\end{document}